%% file: main.tex
\documentclass{article}
\usepackage{array}
\usepackage{multirow}
\usepackage{makecell}
\usepackage{amsmath}
\usepackage{relsize}
\usepackage{physics}
\usepackage{amsfonts}
\usepackage{xspace}
\usepackage[numbers]{natbib}
\usepackage{authblk}
\usepackage{tikz}

\usepackage[english]{babel}
\usepackage{booktabs}
\usepackage{nicematrix}
\usepackage{tabularx}
\usepackage{graphics}

\newcommand{\ind}[1]{{{\mathbb{I}_{\{#1\}}}}}

\usepackage[letterpaper,top=2cm,bottom=2cm,left=3cm,right=3cm,marginparwidth=1.75cm]{geometry}

\usepackage{amsmath}
\usepackage{graphicx}
\usepackage[colorlinks=true, allcolors=blue]{hyperref}
\usepackage{bbm}
\usepackage{tikz}

\newcommand*{\ditto}{--\texttt{"}-- }

\definecolor{plum}  {rgb}{.4,0,.4}
\definecolor{forest}  {rgb}{0,.6,0}
\definecolor{midnight}  {rgb}{0,0,.5}

\newcommand*{\bx}{\mathbf{x}}
\newcommand*{\bX}{\mathbf{X}}

\newcommand*{\bY}{\mathbf{Y}}
\newcommand*{\bz}{\mathbf{z}}
\newcommand*{\bZ}{\mathbf{Z}}

\newcommand*{\bU}{\mathbf{U}}
\newcommand*{\bV}{\mathbf{V}}

\newcommand*{\new}{\textcolor{black}}
\newcommand*{\newold}{\textcolor{black}}

\DeclareMathOperator {\pe}{PE}
\usepackage{bm}
\usepackage{subcaption}

\usepackage{graphicx}
\usepackage[stable]{footmisc}
\usepackage{setspace}
\usepackage[toc,page]{appendix}

\usepackage{titlesec}
\titleformat{\paragraph}[runin]
{\normalfont\bfseries}{}{1em}{}[:]
\titlespacing*{\paragraph}{0pt}{6pt}{3pt}

\usepackage[capitalise]{cleveref}
\crefformat{section}{\S#2#1#3}
\crefformat{appendix}{\S#2#1#3}

\newenvironment{squishlist}
{   \begin{list}{$\bullet$}
    { \setlength{\itemsep}{2pt}      \setlength{\parsep}{2pt}
      \setlength{\topsep}{0pt}       \setlength{\partopsep}{0pt}
      \setlength{\leftmargin}{1.5em} \setlength{\labelwidth}{1em}
      \setlength{\labelsep}{0.5em} } }
      {\end{list}}

\title{Beyond Ensemble Averages: Leveraging Climate Model Ensembles for Subseasonal Forecasting}
\author[1]{Elena Orlova}
\author[1]{Haokun Liu}
\author[2]{Raphael Rossellini}
\author[3]{Benjamin A. Cash}
\author[1,2]{Rebecca Willett}
\affil[1]{Department of Computer Science, University of Chicago, Chicago, IL, USA}
\affil[2]{Department of Statistics, University of Chicago, Chicago, IL, USA}
\affil[3]{Department of Atmospheric, Oceanic, and Earth Sciences, George Mason University, Fairfax, VA, USA}
\affil[ ]{\textit {\{eorlova\}@uchicago.edu}}
\date{}                   
\begin{document}
\maketitle

\begin{abstract}
Producing high-quality forecasts of key climate variables, such as temperature and precipitation, on subseasonal time scales has long been a gap in operational forecasting. This study explores an application of machine learning (ML) models as post-processing tools for subseasonal forecasting. Lagged numerical ensemble forecasts (i.e., an ensemble where the members have different initialization dates) and observational data, including relative humidity, pressure at sea level, and geopotential height, are incorporated into various ML methods to predict monthly average precipitation and two-meter temperature two weeks in advance for the continental United States. For regression, quantile regression, and tercile classification tasks, we consider using linear models, random forests, convolutional neural networks, and stacked models (a multi-model approach based on the prediction of the individual ML models). Unlike previous ML approaches that often use ensemble mean alone, we leverage information embedded in the ensemble forecasts to enhance prediction accuracy. Additionally, we investigate extreme event predictions that are crucial for planning and mitigation efforts. Considering ensemble members as a collection of spatial forecasts, we explore different approaches to using spatial information. Trade-offs between different approaches may be mitigated with model stacking. Our proposed models outperform standard baselines such as climatological forecasts and ensemble means. In addition, we investigate feature importance, trade-offs between using the full ensemble or only the ensemble mean, and different modes of accounting for spatial variability.
\end{abstract}

\let\clearpage\relax


\include{main_part_final}
\bibliography{references_new}



\appendix
\include{appendix_part_final}

\newpage

\end{document}

%% file: main_part_final.tex
\section{Introduction}
High-quality forecasts of key climate variables such as temperature and precipitation on subseasonal time scales, defined here as the time range between two weeks and two months, have long been a gap in operational forecasting \citep{nas_subseasonal}.  Advances in weather forecasting on time scales of days to about a week \citep{lorenc1986analysis, boar16,nati10,simmholl2002} or seasonal forecasts on time scales of two to nine months \citep{barnston2012skill} do not necessarily translate to the challenging subseasonal regime.  Addressing the crucial need for forecasts on the seasonal-to-subseasonal (S2S) timescale, collaborative initiatives led by the World Weather Research Programme and the World Climate Research Programme aim to advance S2S forecasting by focusing on mesoscale–planetary-scale interactions, high-resolution simulations, data assimilation methods, and tailored socioeconomic support 
\citep{brunet2010collaboration}. Skillful forecasts on subseasonal time scales would have immense value in agriculture, insurance, and economics \citep{white2022advances, mouatadid2023adaptive}. The importance of improved subseasonal predictions is also detailed by \citet{nas_subseasonal} and \citet{nati10}.

The National Centers for Environmental Prediction (NCEP), part of the National Oceanic and Atmospheric Administration (NOAA), currently issues a ``week 3-4 outlook" for the contiguous United States (CONUS).\footnote{https://www.cpc.ncep.noaa.gov/products/predictions/WK34/}  
The NCEP outlooks are constructed using a combination of dynamical and statistical forecasts, with statistical forecasts based largely on how conditions in the past have varied (linearly) with indices of the El Ni\~no-Southern Oscillation (ENSO), Madden-Julian Oscillation (MJO), and global warming (i.e., the 30-year trend). There exists great potential to advance subseasonal forecasting (SSF) using machine learning (ML) techniques. \citet{haupt2021towards} provides an overview of \new{using} ML methods \new{for} post-processing of numerical weather predictions. \citet{vannitsem2021statistical} highlight the crucial role of statistical post-processing techniques, including ML methods, in national meteorological services. They discuss theoretical developments and operational applications, current challenges, and potential future directions, \new{particularly focusing} on translating research findings into operational practices.
A real-time forecasting competition called the Subseasonal Climate Forecast Rodeo \citep{rodeo}, sponsored by the Bureau of Reclamation in partnership with NOAA, USGS, and the U.S. Army Corps of Engineers, illustrated that teams using ML techniques can outperform forecasts from NOAA's operational seasonal forecast system. 

Here, we present work focused on developing ML-based forecasts that leverage lagged ensembles (\new{i.e., }an ensemble whose members are initialized from a succession of different start dates) of forecasts produced by NCEP in addition to observed data and other features. Previous studies, including successful methods in the Rodeo competition (e.g., \citet{hwang2019improving}), incorporate the ensemble mean as a feature in their ML systems but do not use any other information about the ensemble. In other words, variations among the ensemble members are not reflected in the training data or incorporated into the learned model. 
In contrast, this paper {\em demonstrates that the full ensemble contains important information for subseasonal forecasting outside the ensemble mean.} 
Specifically, we consider the test case of predicting monthly  2-meter temperatures and precipitation two weeks in advance over 3000 locations over the continental United States using physics-based predictions, such as NCEP-CFSv2 hindcasts \citep{kirtman2014north,saha2014ncep}, using an ensemble of 24 distinct forecasts. We repeat this experiment for the Global Modeling and Assimilation Office from the National Aeronautics and Space Administration (NASA-GMAO) ensemble, which has 11 ensemble members \citep{nasagmao}.

In this context, this paper makes the following contributions:
\begin{itemize}
    \item We train a variety of ML models (including neural networks, random forests, linear regression, and model stacking) that input all ensemble member predictions as features in addition to observations of geopotential heights, relative humidity, precipitation, and temperature from past months to produce new forecasts with higher accuracy than the ensemble mean; forecast accuracy is measured with a variety of metrics (Section \ref{sec:exp_res}).  
    These models are considered in the context of regression, quantile regression, and tercile classification. Systematic experiments are used to characterize the influence of individual ensemble members on predictive skill (Section \ref{sec:when_to_trust}).
    \item The collection of ML models employed allows us to consider different modes of accounting for spatial variability. ML models can account for spatial correlations among both features and targets; for example, when predicting Chicago precipitation, our models can leverage not only information about Chicago but also about neighboring regions. Specifically, we consider the following learning frameworks: (a) training a predictive model for each spatial location independently; (b) training a predictive model that inputs the spatial location as a feature and hence can be applied to any single spatial location; (c) training a predictive model for the full spatial map of temperature or precipitation -- i.e., predicting an outcome for all spatial locations simultaneously.
    ML models present various ways to account for spatial variability, each with distinct advantages and disadvantages. Our application of model stacking (an ML technique where multiple models are combined, with their predictions used as input features for another model that produces the final prediction) allows our final learned model to exploit the advantages of each method.
    \item We conduct a series of experiments to help explain the learned model and which features the model uses most to make its predictions. We systematically explore the impact of using lagged observational data in addition to ensemble forecasts and positional encoding to account for spatial variations (Section \ref{sec:variable_emportance}).
    \item The ensemble of forecasts from a physics-based model (e.g., NCEP-CFSv2 or NASA-GMAO) contain information salient to precipitation and temperature forecasting besides their mean, and ML models that leverage the full ensemble generally outperform methods that rely on the ensemble mean alone (Section \ref{sec:full_vs_avg}).
    \item Finally, we emphasize that the final validation of our approach was conducted on data from 2011 to 2020 that was not used during any of the training, model development, parameter tuning, or model selection steps. We only conducted our final assessment of the predictive skill for 2011 to 2020 after we had completed all other aspects of this manuscript. Because of this, our final empirical results accurately reflect the anticipated performance of our methods on new data.
\end{itemize}

This paper is organized as follows: Section \ref{sec:related_work} discusses related work, Section \ref{sec:data_desc} introduces data used in the experiments,  Section \ref{sec:forecat_tasks} describes forecasting problems, while baselines and learning-based methods are described in Section \ref{sec:baselines}, experimental setup and evaluation metrics are given in Section \ref{section:preprocessing}. Finally, we present our results in Section \ref{sec:exp_res} and discuss them in Section \ref{section:discussion}. Conclusions and directions for future work are given in Section \ref{sec:conclusions}.

\section{Related work}\label{sec:related_work}

While statistical models were common for weather prediction in the early days of weather forecasting \citep{nebeker1995calculating}, forecasts using physics-based dynamic system models have been carried out since the 1980s and have been the dominant forecasting method in climate prediction centers since the 1990s \citep{barnston2012skill}. Many physics-based forecast models are used both in academic research and operationally. Such systems often produce ensembles of forecasts -- e.g., the result of running a physics-based simulation multiple times with different initial conditions or parameters, and are a mainstay of operational forecast centers around the globe. 

Recently, skillful ML approaches have been developed for short-range weather prediction \citep{chen2023machine, nagaraj2023univariate, frnda2022ecmwf, herman2018dendrology, ghaderi2017deep, grover2015deep, radhika2009atmospheric, cofino2002bayesian} and longer-term weather forecasting \citep{lam2023learning, yang2023improving, chen2023machine, hewage2021deep, cohen2019s2s, totz2017winter, iglesias2015examination, badr2014application}. However, forecasting on the subseasonal timescale, with  2-8 week outlooks, has been considered a far more difficult task than seasonal forecasting due to its complex dependence on both local weather and global climate variables \citep{vitart2012subseasonal, min2020recent}. Seasonal prediction also benefits from targeting a much larger averaging period. 

Some ML algorithms for subseasonal forecasting use purely observational data (i.e., not using any physics-based ensemble forecasts). \citet{he2020sub} focuses on analyzing different ML methods, including Gradient Boosting trees and Deep Learning (DL) for SSF. They  propose a careful construction of feature representations of observational data and show that ML methods are able to outperform a climatology baseline, i.e., predictions corresponding to the 30-year
mean at a given location and time. This conclusion is based on comparing the relative $R^2$ scores for the ML approaches and climatology. \citet{srinivasan2021subseasonal} proposes a Bayesian regression model that exploits spatial smoothness in the data.

Other works use the ensemble mean as a feature in their ML models. For example, in the subseasonal forecasting Rodeo \citep{rodeo}, a prediction challenge for temperature and precipitation at weeks 3-4 and 5-6 in the western U.S.\ sponsored by NOAA and the U.S.\ Bureau of Reclamation, simple yet thoughtful statistical models consistently outperform NOAA's dynamical systems forecasts. In particular, the winning approach uses a stacked model from two nonlinear regression models, a selection of climate variables such as temperature, precipitation, sea surface temperature, sea ice concentration, and a collection of physics-based forecast models including the ensemble mean from various modeling centers in \new{the North American Multi-Model Ensemble} (NMME). From the local linear regression with multitask feature selection model analysis, the ensemble mean is the first- or second-most important feature for forecasting, especially for precipitation. \citet{he2021learning} perform a comparison of modern ML models that use data from the Subseasonal Experiment (SubX) project for SSF in the western contiguous United States. The experiments show that incorporating the ensemble mean as an input feature to ML models leads to a significant improvement in forecasting performance, but that work does not explore the potential value of individual ensemble members aside from the ensemble mean. \citet{GrEnsShort} note that physics-based ensembles are computationally demanding to produce and propose an ML method that can input a subset of ensemble forecasts and generate an estimate of the full ensemble; they observe that the output ensemble estimate has more prediction skill than the original ensemble. \citet{LokenRF} analyze the forecast skill of random forests leveraging the ensemble members for next-day severe weather prediction compared to only using the ensemble mean. However, their results only cover forecasts with a lead time of up to 48 hours, so it is unclear if their methods would have succeeded in the tougher subseasonal forecasting setting.  

This paper complements the prior work above by developing powerful learning-based approaches that incorporate both physics-based forecast models and observational data to improve SSF over CONUS. 

\section{Data}\label{sec:data_desc} 

Table \ref{table:data_desc} describes variables used in the experiments. Climatological means of precipitation and temperature are calculated using 1971-2000  NOAA data \citep{noaa_climate}. There are many ensembles of physics-based predictions produced by forecasting systems. \new{NMME} is a collection of physics-based forecast models from various modeling centers in North America, including NOAA/NCEP, NOAA/Geophysical Fluid Dynamics Laboratory (GFDL),  International Research Institute for Climate and Society (IRI), National Center for Atmospheric Research (NCAR), NASA, and Canadian Meteorological Centre \citep{kirtman2014north}. NMME  provides forecasts from multiple global forecast models from North American modeling centers \citep{kirtman2014north}. The NMME project has two predictive periods: hindcast and forecast. A hindcast period refers to when a dynamic model re-forecasts historical events, which can help climate scientists develop and test new models to improve forecasting and \new{to} evaluate model biases. In contrast, a forecast period has real-time predictions generated from dynamic models.  

\begin{table}[ht!]
\caption{\textbf{Description of climate variables and their data sources.} Our target climate variables for sub-seasonal forecasting are precipitation and 2-meter temperature. We use NOAA data to calculate the climatology from 1971 to 2000. We also perform linear spatial interpolation on the historical values to get values with the same \new{resolution and }support as target climate variables. 
}
\label{table:data_desc}
\centering
\small
\scalebox{0.92}{
\begin{tabular}{ c c  c  c  c  c  c  c  c} 
 \hline
 Type & Variable & Description & Unit  & Spatial Coverage & Time Range & Data Source \\  
 \hline 
 \multirow{2}{*}{\rotatebox{90}{Feature variable}} & tmp2m & \makecell{Daily average \\ temperature at 2 meters} & $^\circ C$ & \makecell{US mainland \\ $0.5^\circ \times 0.5^\circ$ grid} & \makecell{1985 to 2020} & \makecell{CPC Global \\ Daily Temperature \\ \citep{fan2008global}}  \\  \cline{2-7}
 & precip & \makecell{Daily average \\ precipitation} & mm & \makecell{US mainland \\ $0.5^\circ \times 0.5^\circ$ grid} & \makecell{1985 to 2020} & \makecell{CPC Global \\ Daily Precipitation \citep{xie2010cpc} } \\  \cline{2-7}
 & SSTs & \makecell{Daily sea \\ surface temperature} & $^\circ C$ & \makecell{Ocean only \\ $0.25^\circ \times 0.25^\circ$ grid} & \makecell{1985 to 2020} & \makecell{Optimum Interpolation \\ SSTs High Resolution \\ (OISST) \citep{reynolds2007daily} } \\  \cline{2-7}
 & rhum & \makecell{Daily relative humidity \\ near the surface} & Pa & \multirow{10}{*}{} & \multirow{10}{*}{}  \\  \cline{2-4} 
& slp & \makecell{Daily pressure  \\ at sea level } & $\%$ & \makecell{US mainland \\ and North Pacific \\ \& Atlantic Ocean \\$0.5^\circ \times 0.5^\circ$ grid}  & \makecell{1985 to 2020} & \makecell{Atmospheric Research \\ Reanalysis Dataset \\ \citep{kalnay1996ncep}}  \\  \cline{2-4}
 & hgt500 & \makecell{Daily geopotential \\ height at 500mb} & m & \multirow{10}{*}{} & \multirow{10}{*}{}  \\  
\hline 
\multirow{2}{*}{\rotatebox[origin=c]{90}{Climatology}} & \rule{0pt}{4ex}     tmp2m & \makecell{Daily average \\ temperature at 2 meters} & K & \makecell{Globally \\ $1^\circ \times 1^\circ$ grid} & \makecell{1971 to 2000} & NOAA \citep{noaa_climate}  \\ \cline{2-7}
 & \rule{0pt}{5ex} precip & \makecell{Daily average \\ precipitation} & mm & \makecell{Globally \\ $1^\circ \times 1^\circ$ grid} & \makecell{1971 to 2000} & NOAA \citep{noaa_climate}  \\  
 \hline 
\end{tabular}}
\end{table}

In this manuscript, we use ensemble forecasts from the NMME's NCEP-Climate Forecast System version 2 (CFSv2, \citet{kirtman2014north,saha2014ncep}), which has $K = 24$  ensemble members at a $1^\circ \times 1^\circ$ resolution over a 2-week lead time. NCEP-CFSv2 is the operational prediction model currently used by the U.S. Climate Prediction Center.
The NCEP-CFSv2 model has two different products available in the NMME archive: we use its hindcasts from 1982 to 2010 for training and validation of our models, and we use its forecasts from April 2011 to December 2020 for the final evaluation of our models. 

In order to ensure our results are not unique to a single forecasting model, we also analyze output from the NASA-Global Modeling and Assimilation (GMAO) from the Goddard Earth Observing System model version 5 (GEOS, \citet{nasagmao}), which has $K = 11$  ensemble members at a $1^\circ \times 1^\circ$ resolution over a 2-week lead time. Similarly, we use its hindcasts from 1981 to 2010 for training and validation of our models, and we use its forecasts from January 2011 to January 2018 for final evaluation. The test periods of NCEP-CFSv2 and NASA-GMAO data differ due to  data availability. Note that the identical version of each model is used to generate the test, train, and validation data.

Different ensemble members correspond to different initial conditions of the underlying physical model. The NCEP-CFSv2 forecasts are initialized in the following way: four initializations at times 0000, 0600, 1200, and 1800 UTC every fifth day, starting one month prior to the lead time of two weeks 
(Table B1 in \citet{saha2014ncep}). 
NASA-GMAO is a fully coupled atmosphere–ocean–land–sea ice model, with five forecasts initialized every five days. While additional members are generated through perturbation methods closest to the beginning of each month \footnote{https://gmao.gsfc.nasa.gov/products/climateforecasts/GEOS5/DESC/init.php}, not all members are initialized on different dates, meaning that the ensemble is not strictly lagged. However, NASA-GMAO members are not interchangeable, as each is created using a distinct method.

All data are interpolated to lie on the same $1^\circ \times 1^\circ$ grid, resulting in $L=3,274$ U.S.\ locations. Climate variables available daily (such as pressure at sea level or precipitation) are converted to monthly average values. When data are available as monthly averages only, we ensure
that our forecast for time $t+\delta_t$ does not use any information from the interval $(t,t+\delta_t)$.

\section{Forecast tasks}\label{sec:forecat_tasks}

The learning task can be formulated as learning a \newold{model}
$f_\theta: \mathbf{X} \to \mathbf{y}$
with parameters $\boldsymbol{\theta}$. 
This \newold{model} $f_\theta$ can be a linear regression (where $\boldsymbol{\theta}$ is a set of regression weights), the mean of ensemble members (no $\boldsymbol{\theta}$ needs to be learned), a random forest (where $\boldsymbol{\theta}$ parameterizes the set of trees in the forest), a convolutional neural network (where $\boldsymbol{\theta}$ \new{is the collection of} neural network weights), or other learned models. 
We consider three forecasting tasks: regression, tercile classification, and quantile regression. 

\paragraph{Regression} The goal of regression is to predict monthly average values of precipitation and 2-meter temperature two weeks in the future. 
These models are generally trained using the squared error  loss function: 
\begin{equation}
    \ell_{\rm sq-err}(\theta) = \mathbb{E}[(y-f_\theta(\bx))^2].
\end{equation}

\paragraph{Tercile classification} 
The goal of tercile classification is to predict whether the precipitation or 2-meter temperature will be ``high'' (above the $66$th percentile, denoted $q=1$), ``medium'' (between the $33$rd and the $66$th percentiles, denoted $q=0$), or ``low'' (below the $33$rd percentile, denoted $q=-1$). 
We compute these percentile values using the 1971-2000 climatology (see Section \ref{sec:data_desc}
for details), and these percentiles are computed for each calendar month $m$ and location $l$ pair.
These models are generally trained using the cross-entropy loss function:

\begin{equation}
\ell_{\rm CE}(\theta) = \mathbb{E}\left[\sum_{q = -1}^1 -\ind{y=q} \log (f_\theta(\bx))_q\right],
\end{equation}
where
$\ind{A} := {\begin{cases} 
1, & \text{if } A \text{ true}\\
0, & \text{if } A \text{ false}
\end{cases}}$ is the indicator function and $(f_\theta(x))_q$ is the predicted probability that the target $y$ corresponding to feature vector $x$ will be in tercile $q$.
\paragraph{Quantile regression} 
For a given percentile $\alpha$, the goal of quantile regression is to predict
the value $z$ so that, conditioned on features $x$, the target $y$ satisfies $y \le z$ with probability $\alpha$. When we set $\alpha$ to a value close to one, such as $\alpha = 0.9$, this value $z$ indicates what we can expect in ``extreme outcomes'', not just on average. 
These models are generally trained using the pinball loss function:

\begin{equation}
    \ell_{\rm quantile}(\theta) = \mathbb{E}[\rho_\alpha (y - f_\theta(\bx))]\\
\label{eq:pinball}
\end{equation}
where
\begin{equation}
\label{qrloss}
    \rho_\alpha (z) := z(\alpha - \ind{(z<0)}) = 
    \begin{cases}
        \alpha |z| & \text{if } z\geq 0\\
        (1-\alpha) |z| & \text{if } z <0
    \end{cases}.
\end{equation}

\section{Prediction methods}\label{sec:baselines}

Our goal is to predict either the monthly average precipitation or the monthly average 2-meter temperature two weeks in advance (for example, we predict the average monthly precipitation for February on January 15).
This section describes the notation used for features and targets, baselines and learning methods, and how spatial features are accounted for.

\subsection{Notation}
We let $T$ denote the \new{number of time steps used in our analysis}, and $L$ denote the number of spatial locations.
We define the following variables:
\begin{itemize}
    \item $u_{t,l}^{(k)}$ 
    is the $k$-th ensemble member at time $t$ and location $l$, where $k=1, \dots, K$, $t=1, \dots, T$,   $l=1, \dots, L$. Every ensemble member represents the output of a given physics-based model forecast from different initial states.  
    \item $v_{t,l}^{(p)}$ is the $p$-th observational variable, such as precipitation or temperature, geopotential height at 500mb, relative humidity near the surface, pressure at sea level and sea surface temperature,
    at time $t$ and location $l$, with $p= 1,..., P$.
    \item  $\bz_l^{(1)}, \bz_l^{(2)}$ represent information about longitude and latitude of location $l$, respectively; each is a vector of length $d$. More details about this representation, \new{called positional encoding (PE),} can be found in Section \ref{sec:PE}. 
    \item$\mathbf{x}_{t, l}:=[u_{t, l}^{(1)}, ..., u_{t, l}^{(K)}, v_{t, l}^{(1)} ..., v_{t, l}^{(P)}, \bz_l^{(1)}, \bz_l^{(2)}]$ is a set of features at time $t$ and location $l$.
    \item $y_{t, l}$ is the target
    -- the ground truth monthly average precipitation or 2-meter temperature at the target forecast time $t+\delta_t$ 
    at location $l$, \new{where $\delta_t = 14$ days is our forecast horizon}. 
    For simplicity, we use a subscript $t$ for $y_{t, l}$ instead of $t+\delta_t$ to match with the input features notation. The same holds for \new{our ensuing definitions}. 
    \item $\hat{y}_{t, l}$ is the output of a forecast model for a given task at target forecast time $t+\delta_t$ 
        and location $l$. 
    \item $s_{m, l}$ -- a 30-year mean (climatology) of an observed climate variable, such as precipitation or temperature, at a month $m = 1,\ldots,12$ and location $l$.
    \item $\hat{s}_{m, l} $ -- a
    30-year climatology of a \textit{predicted} climate variable, such as precipitation or temperature, at a month $m = 1,\ldots,12$ and location $l$. For each location $l$ and each month $m$, it is calculated as a mean of ensemble member predictions over the training period, as defined formally in \cref{eq:model_climatology}.
    \item ${y}_{t, l}^\text{anomaly}$ and $\hat{y}_{t, l}^\text{anomaly}$ are anomaly predictions and a true anomaly, at a month $m = 1,\ldots,12$ and location $l$. They are used during evaluation. We define anomalies as 
    \begin{align}\label{eq:anomalies}
        y_{t, l}^{\text{anomaly}} &= y_{t, l} - s_{m(t), l},\\
        \hat y_{t, l}^{\text{anomaly}} &= \hat y_{t, l} - s_{m(t), l}.
    \end{align}
    For the special case of $\hat y$ corresponding to the ensemble mean, the ensemble members may exhibit bias, in which case we also consider
        $\hat y_{t, l}^{\text{anomaly}} = \hat y_{t, l} - \hat s_{m(t), l},$
    where $\hat s_{m(t), l}$ is evaluated on the model's (ensemble mean) predictions:
    \begin{equation}
        \hat s_{m,l} := \frac{1}{T}\sum_{t=1}^{T} \hat y_{t,l} \ind{m=m(t)}, ~ l = 1, \dots, L.
        \label{eq:model_climatology}
    \end{equation}
   The model climatology $\hat s_{m,l}$ is computed using the training data. Note that we do not subtract the climatology from the input features and target variables, i.e., precipitation and temperature, when training our ML models. We subtract climatology from the model outputs only when evaluating their performance, as including climatology in the inputs to our ML models \new{during training} improves performance. Section \ref{sec:evaluation_metrics} provides more details on the model evaluation.
\end{itemize}

In our analyses, the number of locations is $L=3274$, there are $K=24$ NCEP-CFSv2 ensemble members or $K=11$ NASA-GMAO ensemble members. The ensemble members are used as input features to the learning-based methods as they are, we do not perform any feature extraction from them. The number of observational variables is usually $P=17$. The details on these variables can be found in Section \ref{sec:data_desc} and Section \ref{sec:input_ftrs}.

The target variable $y$ is observed from 1985 to 2020. Data from January 1985 to September 2005 are used for training (249 time steps), and data from October 2005 to December 2010 are used for validation and model selection (63 time steps). Data from 2011 to 2020 (or from 2011 to 2018 in the case of NASA-GMAO data) are used to test our methods after all model development, selection, and parameter tuning are completed.

\subsection{Baselines}
\paragraph{Climatology} 
It is the fundamental benchmark for weather and climate predictability. In particular, for a given time $t$, let $m(t) := (t\mod 12)$ correspond to the calendar month corresponding to $t$; then we compute the 30-year climatology of the target variable for a given location and time via

\begin{equation}
    \hat y_{t,l}^{\rm hist} = 
    s_{m(t),l}, ~ t = 1, \dots, T, ~l = 1, \dots, L.
    \label{eq:climatology}
\end{equation}

\paragraph{Ensemble mean} This is the mean of all ensemble members for each location $l$ at each time step $t$: 
\begin{equation}
    \hat y_{t,l}^{\rm ens~mean} := \frac{1}{K} \sum_{k = 1}^K u_{t, l}^{(k)}, ~ t = 1, \dots, T, ~l = 1, \dots, L.
\end{equation}

\paragraph{Linear regression}
Finally, we consider, as a baseline, a \textit{linear regression} model applied to input features corresponding to ensemble member predictions: $\bx_{t,l} = [u^{(1)}_{t,l}, \dots, u^{(K)}_{t,l}]$.
Then, the model's output
 \begin{equation}\label{eq:LR}
\hat y_{t,l}^{\rm LR} :=   	\langle \boldsymbol{\theta}_l,\bx_{t,l} \rangle + \theta_l^0, 
\end{equation}
where $\theta_l$ are the trained coefficients for input features for each location $l$, and $\theta_l^0$ are the learned intercepts for each location $l$. Note that we train a different model for each spatial location, and the illustration for this model and its input's format is given in \Cref{fig:models_fig}(a).

\subsection{Learning-based methods}
\paragraph{Linear regression (LR)} In contrast to the linear regression baseline, here other climate variables are added to the input features:
$\bx_{t,l} = [u^{(1)}_{t,l}, ..., u^{(K)}_{t,l},v^{(1)}_{t,l},...,v^{(P)}_{t,l}] $. Then the model's output is defined with \cref{eq:LR}.
Because the feature vector is higher dimensional here than for the baseline, the learned $\theta_l$ is also higher dimensional. We train a different model for each spatial location. In our experiments with linear models, we do not include positional encoding $(\mathbf{z}_l^{(1)}, \mathbf{z}_l^{(2)})$ as input features, since they would be constants for \new{each} location's linear model. 

In the context of regression, we minimize the squared error loss. The linear quantile regressor (Linear QR) is a linear model trained to minimize the quantile loss
\begin{equation}
    \label{QRloss}
     \ell^{\rm QR} = \frac{1}{L} \sum_{l=1}^{L}\left[\frac{1}{T}\sum_{t=1}^{T} \rho_\tau\left(y_{t, l} - \hat{y}_{t, l,}\right)\right],
\end{equation}
where $\rho_\tau$ is defined in \cref{qrloss}.

\paragraph{Random forest} 
In the context of regression and tercile classification, we train a random forests that use ensemble predictions, the spatial location, and additional climate features to form the feature vector $\bx_{t,l} = [u^{(1)}_{t,l}, ..., u^{(K)}_{t,l},v^{(1)}_{t,l},...,v^{(P)}_{t,l}, \mathbf{z}_l^{(1)}, \mathbf{z}_l^{(2)}]$ for all location $l$ and time $t$ pairs. One random forest is trained to make predictions for any spatial location. The illustration for RF and its input's format is given in \Cref{fig:models_fig}(b): we train one RF model for all locations, and the spatial information is encoded as input features via PE vectors $\bz_l^{(1)}, \bz_l^{(2)}$.

In the context of quantile regression, we train a random forest quantile regressor (RFQR, \citet{RFQR:2006}), which grows trees the same way as the original random forest while storing all training samples. To make a prediction for a test point, the RFQR computes a weight for each training sample that corresponds to the number of leaves (across all trees in the forest) that contain the test sample and the training sample. The RFQR prediction is then a quantile of the weighted training samples across all leaves that contain the test sample. We show a figure representation of the RFQR in Section \ref{RFQR}.
With this formulation, training a single RFQR for all locations is computationally demanding, so we train individual RFQRs for every location.

Random forests are often referred as the best off-the-shelf classifiers \citep{hastie2009elements} even using the default hyperparameters \citep{biau2016random}. Our cross-validation (CV) and grid search experiments show that the RFs hyperparameters have little impact on the accuracy, and thus, we use the default parameters for RFs from the Scikit-learn library \citep{scikit-learn}.

\paragraph{Convolutional neural network} 
To produce a forecast map for the U.S., we adapted a U-Net architecture \citep{Ronneberger2015UNetCN}, which has an encoder-decoder structure with convolutional layer blocks. The U-Net maps a stack of images to an output image; in our context, we treat each spatial map of a climate variable or forecast as an image. Thus, the input to our U-Net is can be represented as a tensor composed of matrices:
$\bX_t = [\bU^{(1)}_t, ..., \bU^{(K)}_t, \bV^{(1)}_t, ..., \bV^{(P)}_t, \bZ^{(1)}, \bZ^{(2)}]$.

Note that here, we use capital letters because the input to our U-Net consists of 2-D spatial maps, which are represented as matrices instead of vectors. The model output is a spatial map of the predicted target. This process is illustrated in \Cref{fig:models_fig}(c).

For the U-Net, we modify an available PyTorch implementation \citep{Yakubovskiy:2019}. The training set consists of $249$ samples (images), which may be considered relatively limited for CNN training. To address this concern, we conduct bootstrapping experiments for the U-Net architecture, offering detailed insights into the impact of sample size on model performance. Further details are presented in Appendix \ref{sec:unet_bootstrap}. We use a 10-fold CV over our training data and grid search to select parameters such as learning rate, weight decay, batch size, and number of epochs. The Adam optimizer \citep{kingma2014adam} is used in all experiments. 
After selecting hyperparameters, we train the U-Net model with those parameters on the full training dataset. The validation set is used to perform feature importance analysis.
For regression, we train using squared error loss. 
In the context of quantile regression,  we initialize the weights with those learned on squared error loss and then train on the quantile loss \cref{QRloss}.

\paragraph{Nonlinear model stacking}
Model stacking can improve model performance by combining the outputs of several models (usually called base models) \citep{pavlyshenko2018using}. 
In our case, linear regression, random forests, and the U-Net are substantially different in architecture and computation, and we observe that they produce qualitatively different forecasts.
We stack the linear model, random forest, and U-Net forecasts using a nonparametric approach: 
\begin{equation}
    \hat y_{t, l} =  h(\hat y_{t, l}^{\text{LR}}, \hat y_{t, l}^{\text{RF}}, \hat y_{t, l}^{\text{UNET}}),
\end{equation}
where $h$ is a simple feed-forward neural network with a non-linear activation \new{and } $\hat y_{t, l}^{\text{LR}} , \hat y_{t, l}^{\text{RF}}, \hat y_{t, l}^{\text{UNET}}$ are the predictions of a linear model, random forest and the U-Net correspondingly and referred to as ``base models".
One stacking model is trained to make predictions for any spatial location. \Cref{fig:models_fig}(b) \new{with input features that are predictions from other ML models and no PE vectors} demonstrates the stacking model's framework. Model stacking can improve the forecast quality by combining predictions from three forecasting paradigms -- spatial independence, conditional spatial independence, and spatial dependence (Section \ref{sec:spatial}), and is analogous to the multi-model ensemble approach commonly used in weather and climate forecasting. The architecture details can be found in Appendix \ref{app:architecture}.

We apply the following procedure for model stacking: the base models are first trained on half of the training data, and predicted values on the second half are used to train the stacking model $h$. Then, we retrain the base models on all the training data and apply the trained stacked model to the outputs of the base models. The proposed procedure helps to avoid overfitting.

\subsection{Models of spatial variation}
\label{sec:spatial}
We consider three different forecasting paradigms. In the first, which we call the {\bf spatial independence} model, we ignore all spatial information and train a separate model for each spatial location. In the second, which we call the {\bf conditional spatial independence} model, we consider samples corresponding to different locations $l$ as independent conditioned on the spatial location as represented by features $(\mathbf{z}_l^{(1)}, \mathbf{z}_l^{(2)})$. In this setting, a training sample corresponds to $(\mathbf{x}_i,y_i) = (\mathbf{x}_{t,l},y_{t,l})$, where, with a small abuse of notation, we let $i$ index a $t,l$ pair. In this case, the number of training samples is $n=TL$. 
In the third paradigm, which we call the {\bf spatial dependence} model, we consider a single training sample as corresponding to full spatial information (across all $l$) for a single $t$; that is
$(\bX_i,\bY_i) = ([x_{t,l}]_{l=1,\ldots,L},[y_{t,l}]_{l = 1,\ldots,L}])$, where now $i$ indexes $t$ alone. Models developed under the spatial dependence model account for the spatial variations in the features and targets. For instance, a convolutional neural network might input ``heatmaps'' representing the collection of physics-based model forecasts across the continental U.S.\ and output a forecast heatmap predicting spatial variations in temperature or precipitation instead of treating each spatial location as an independent sample. 

\begin{figure}[ht]
\centerline{\includegraphics[width=33pc]{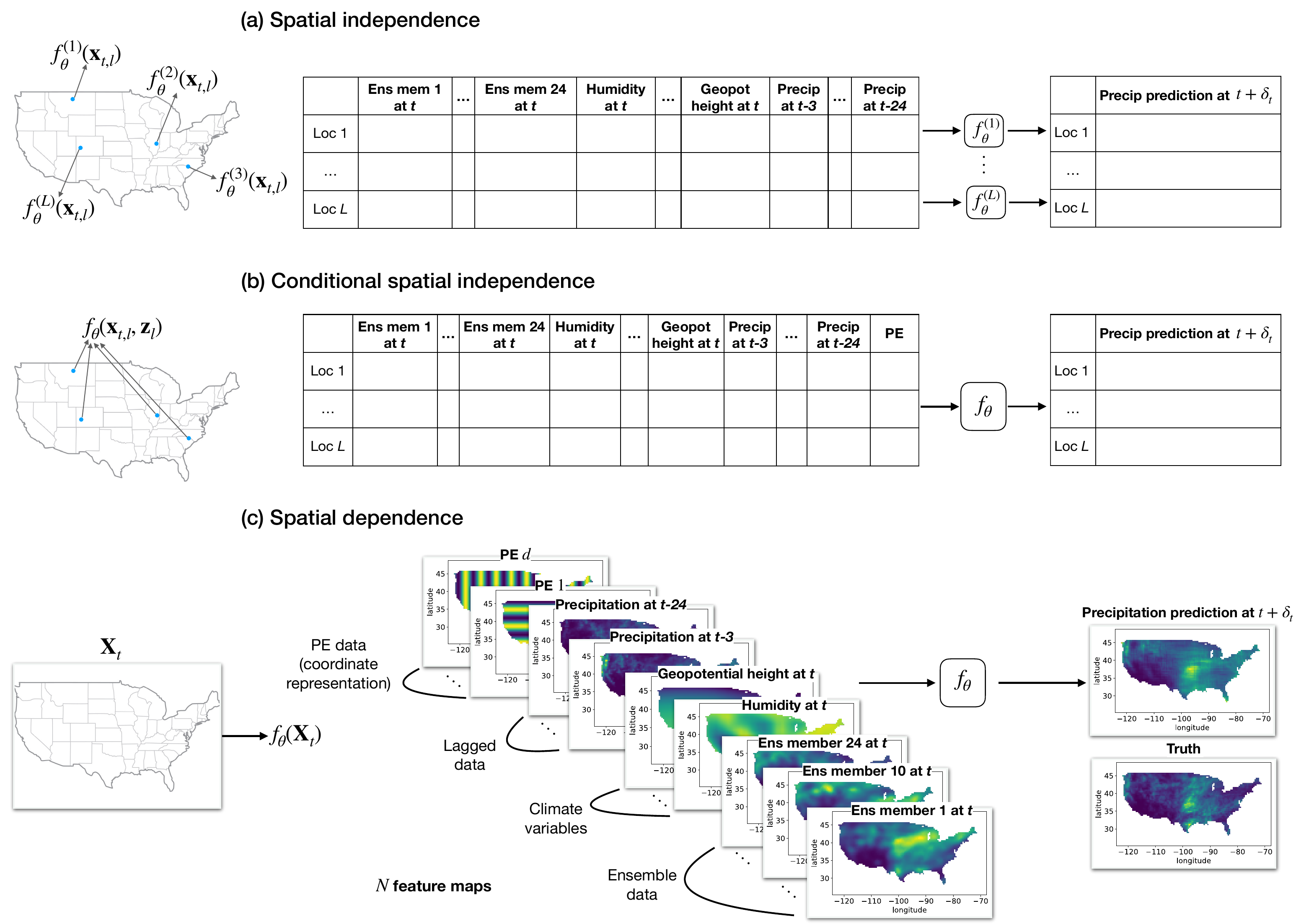}}
  \caption{\textbf{An illustration of different forecasting paradigms}: (a) spatial independence models with a model for each spatial location, no accounting for spatial information; (b) conditional spatial independence models with one model for all locations, might consider the spatial information; (c) spatial dependence models that account for the spatial information by design. We replace “precipitation” in the illustration with “temperature” for temperature prediction, but the overall structure remains the same. }\label{fig:models_fig}
\end{figure}

\Cref{fig:models_fig} shows general frameworks of these paradigms. All models combine information from all the different ensemble forecasts, and so in a broad sense, we can think of each prediction at a given time and location as a weighted sum of the ensemble forecasts across space, time, and ensemble members, where the weights are learned during the model training and may be data-dependent (i.e., nonlinear). From this perspective, we may think of different modeling paradigms as essentially placing different constraints on those weights:
\begin{itemize}
\item under spatial independence models, the weights may vary spatially but do not account for spatial correlations in the data;
\item under conditional spatial independence models, the interpretation depends on the model being trained -- linear models have the same weights on ensemble predictions regardless of spatial location, while nonlinear models (e.g., random forests) have weights that may depend on the spatial location;
\item under spatial dependence models, the weights vary spatially, depend on the spatial location, and account for spatial correlations among the ensemble forecasts and other climate variables. 
\end{itemize}

\section{Experimental setup}\label{section:preprocessing}

This section provides details on the experimental setup, including positional encoding, removing climatology, and evaluation metrics.
Data preprocessing details are presented in Appendix \ref{app:preprocessing}.

\subsection{Models' inputs details}\label{sec:input_ftrs}

Based on the available data, we use the following input features for our ML models:  
\begin{squishlist}
\setlength{\itemsep}{1pt}
    \item $K$ ensemble forecasts for the target month, 
    \item four climate variables: relative humidity, pressure, geopotential height, and temperature (if \new{the} target is precipitation) or precipitation (if \new{the} target is temperature) two months before the target month,
    \item the lagged target variable (the target variable two, three, four, twelve, and 24 months before the target date -- five additional features),
    \item SSTs that are represented via principal components (PCs),
    \item and, finally, the positional embeddings.
\end{squishlist} 
SSTs are usually represented as eight PCs, and the embedding vector size is usually $d=12$ as we describe Section \ref{sec:PE}. For example, using the NCEP-CFSv2 members, there are $\underbrace{24}_{K} + \underbrace{4 + 5 + 8}_{P} + \underbrace{12}_{d} \times 2 = 65$ input features for every time step and location. \new{\Cref{fig:models_fig} provides an illustration of these input features.}

\subsection{Positional encoding}\label{sec:PE}

Positional encoding \citep{vaswani2017attention} is a technique used in natural language processing (NLP) to inject positional information into data. In sequence-based tasks, such as language translation or text generation, the order of elements in the input sequence is important, but neural networks do not naturally capture this information. PE assigns unique encodings to each position in the sequence, which are then added to the original input before being processed by the model. This enables the model to consider the order and relative positions of elements, improving its ability to capture local and global context within a sequence and make accurate predictions \citep{devlin2018bert, petroni2019language, narayanan2016subgraph2vec}. This technique is helpful to represent the positional information outside the \new{original} NLP tasks \citep{gamboa2017deep, gehring2017convolutional, khan2022transformers}. Several of our models use the spatial location as an input feature. Rather than directly using latitudes and longitudes, we  use PE \citep{vaswani2017attention}:

\begin{align}
        z_l^{(1) } (i)=& {\pe}(l, 2i)=\sin(l/10000^{2i/d}), \\
        z_l^{(2) } (i)=& {\pe}(l, 2i+1)=\cos(l/10000^{2i/d}),
    \label{eq:pe}
\end{align}
where $l$ is a longitude or latitude value, $d=12$ is the dimensionality of the positional encoding, and $i \in \{ 1, \dots, d\}$ is the index of the positional encoding vector. For the U-Net model, PE vectors are transformed into images in the following way: we take every value in the vector and fill the image of the desired size with this value. So, there are $d$ images with the corresponding PE values. For the RF models, PE vectors can be used as they are.

\subsection{Evaluation metrics}
\label{sec:evaluation_metrics}

\paragraph{Regression metrics}

The forecast skill of our regression models is measured using the \textit{$R^2$ value}. For each location $l$ and ground-truth values $y_{t,l}$ and predictions $\hat{y}_{t,l}$ at this location, we compute
\begin{equation}
    R^2_l = 1  - \frac{ { \sum_{t = 1}^T} ( y_{t, l}^{\text{anomaly}} - \hat{y}_{t, l}^{\text{anomaly}})^2}{ {\sum_{t = 1}^T} (y_{t, l}^{\text{anomaly}} - \bar{y}_{l}^{\text{anomaly}})^2} ,
    \label{eg:r2_loc}
\end{equation}
where
\begin{equation*}
    \bar{y}_{l}^{\text{anomaly}} = \frac{1}{T}\sum_{t = 1}^T \hat{y}_{t, l}^\text{anomaly}.
\end{equation*}
Then, the average $R^2$ for all locations is calculated as 
\begin{equation}
   R^2 = \frac{1}{L}\sum_{l=1}^{L} R_l^2.
   \label{eq:r2_avg}
\end{equation}
In addition to the average $R^2$ on the test data, we also estimate the median $R^2$ score across all U.S.\ locations. 

We further report the \textit{mean squared error} (MSE) of our predictions  across all locations:
\begin{equation*}
{\rm MSE}_l := \frac{1}{T}\sum_{t=1}^{T} \left(y_{t, l} - \hat{y}_{t, l}\right)^2,
\end{equation*}
for $l = 1, \dots, L$, and 
\begin{equation}
   {\rm MSE} = \frac{1}{L}\sum_{l=1}^{L} {\rm MSE}_l.
   \label{eq:mse}
\end{equation}
We also report the standard error (SE), median, and 90th percentile of $\{{\rm MSE}_l\}_l$. We say the difference between the two models is significant if their MSE $\pm$ SE intervals do not overlap.  Note that the standard errors provided here should be used with caution since there are significant spatial correlations in the MSE values across locations, so we do not truly have $L$ independent samples from an asymptotically normal distribution.

\paragraph{Tercile classification metrics} We estimate the accuracy of our tercile classification predictions as the proportion of correctly classified samples out of all observations.

\paragraph{Quantile regression metrics}
For the quantile regression task, we report \textit{mean quantile loss} from \cref{QRloss} across all locations. 

\section{Experimental results}\label{sec:exp_res}

In this section, we report the predictive skill of different models applied to SSF over the continental U.S.\  using NCEP-CFSv2 ensemble members for regression and quantile regression. Precipitation forecasting is known to be more challenging compared to temperature forecasting \citep{knapp2011globally}. The results for the NASA-GMAO dataset are presented in Appendix \ref{sec:app_nasa_regr}. 
The skill of different models on the tercile classification task is presented in Appendix \ref{appendix:tercile} for both datasets. 
Recall that all methods are trained on data spanning January 1985–September 2005, with data spanning October 2005 - December 2010 used for
validation (i.e., model selection and hyperparameter tuning). Test data spanning 2011 to 2020 was \textbf{not} viewed at any point of the model development and training process and only used to evaluate the predictive skill of our trained models on previously unseen data; we refer to this period as the ``test period''.
As a navigation tool for the reader, \cref{table:all_tables_res} 
gives references to the presented results for different tasks.

\begin{table}[ht!] 
\caption{A table with references to the main results.}
\centering
\small
\begin{tabular}{c c c} 
 \hline
Task & Data & Reference  \\ 
 \hline
\multirow{1}{*}{Regression} 
 & precip NCEP-CFSv2 & \cref{table:precip_regression_ncep_test}; \Cref{fig:precip_reg_ncep_test}   \\  
 & tmp NCEP-CFSv2 & \cref{table:tmp_regression_ncep_test}; \Cref{fig:tmp_reg_ncep_test}  \\ \cline{2-3}
 & precip NASA-GMAO & \Cref{table:precip_regression_nasa_test}; \Cref{fig:precip_reg_nasa_test}  \\
 & tmp NASA-GMAO & \Cref{table:tmp_regression_nasa_test}; \Cref{fig:tmp_reg_nasa_test} \\ 
 \hline

\multirow{1}{*}{Quantile regression} 
& precip NCEP-CFSv2 & \cref{table:precip_qtr_ncep_test};
\Cref{fig:precip_qtr_loss_ncep_test} \\  
 & tmp NCEP-CFSv2 & \cref{table:tmp_qtr_ncep_test}; \Cref{fig:tmp_qtr_loss_ncep_test} \\ \cline{2-3}
 & precip NASA-GMAO & \Cref{table:precip_qtr_nasa_test}; \Cref{fig:precip_qtr_loss_nasa_test}  \\  
 & tmp NASA-GMAO & \Cref{table:tmp_qtr_nasa_test}; \Cref{fig:tmp_qtr_loss_nasa_test} \\ 

\hline
\multirow{1}{*}{Feature importance}
& precip NCEP-CFSv2 &  \cref{table:precip_regression_ncep_ftrs} \\ 
& tmp NCEP-CFSv2 &  \cref{table:tmp_regression_ncep_ftrs} \\ 
\hline

\multirow{1}{*}{Tercile classification} 
 & precip NCEP-CFSv2 & \Cref{table:precip_tercile_test}; \Cref{fig:tercile_precip_ncep} \\  
 & tmp NCEP-CFSv2 & \Cref{table:tmp_tercile_test}; \Cref{fig:tercile_precip_nasa}  \\ \cline{2-3}
 & precip NASA-GMAO & \Cref{table:precip_tercile_test}; \Cref{fig:tercile_precip_nasa} \\  
 & tmp NASA-GMAO & \Cref{table:tmp_tercile_test}; \Cref{fig:tercile_tmp_nasa} \\ 
\hline
\end{tabular}
\label{table:all_tables_res}
\end{table}

\subsection{Regression} \label{sec:regr_ncep}

\paragraph{Precipitation regression using NCEP-CFSv2 }\label{sec:precip_regr_ncep}
Precipitation regression results 
are presented in \cref{table:precip_regression_ncep_test}. 
While the individual ML approaches produce results generally similar to those of the baselines, the stacked ML model, in particular, outperforms the baseline models in almost all metrics. Note that the best $R^2$ value, associated with the stacked model, is still near zero; while this is a significant improvement over, for example, the ensemble mean, which has an $R^2$ value of -0.08, the low values for all methods indicate the difficulty of the forecasting problem.
It is important to note that $R^2$ measures the accuracy of a model relative to a baseline corresponding to the mean of the target \textit{over the test period} -- that is, relative to a model that could never be used in practice as a forecaster because it uses future observations. The best \textit{practical} analog to this would be the mean of the target \textit{over the train period} -- what we call the ``historical mean'' or climatology model. These two models are not the same, possibly because of the nonstationarity of the climate \citep{min2020recent}. Thus, even when our $R^2$ values are negative (i.e., we perform worse than the impractical mean of the target over the test period), we still perform much better than the practical climatology predictor.
The model stacking approach is applied to the models trained on all available features (i.e., ensemble members, PE, climate variables; linear regression is trained on all features except PE). We decide what models to include in the stacking approach based on their performance on validation data. The low 90th percentile error implies that our methods not only have high skill on average but also that there are relatively few locations with large errors.
While acknowledging the overall performance may not be exceptional, it is important to recognize the potential of machine learning methods in improving the quality of estimates relative to the standard baselines. 
To further evaluate the capabilities of the stacking approach, we \new{also }apply the \new{approach} to the baseline predictions, which include historical and ensemble means, as well as linear regression. The performance of the stacked baseline model exceeds that of any of the individual baseline models and is similar to the performance of the stacked ML approach in terms of the $R^2$ metric. However, the stacked ML approach outperforms it in all MSE-based metrics, indicating that the ML techniques can still provide additional skill even for as notoriously challenging a quantity as precipitation.

\begin{table}[ht!]
\caption{\textbf{Results for precipitation regression the using NCEP-CFSv2 ensemble, with errors reported over the test period.}  
LR refers to linear regression on all features, including ensemble members, lagged data, climate variables, and SSTs. \new{ML} model stacking is performed on models that are trained on all features. The \textbf{best} results are in bold. \newold{MSE is reported in squared mm.}}
\label{table:precip_regression_ncep_test}
\centering
\small
\begin{tabular}{c c  c  c  c  c  c c } 
\hline
Model & Features & \makecell{Mean \\ $R^2$ ($\uparrow$)} & \makecell{Median\\ $R^2$ ($\uparrow$)}& \makecell{Mean\\ Sq Err ($\downarrow $)} & \makecell{Median \\ MSE ($\downarrow$)} &
\makecell{90th prctl \\ MSE ($\downarrow$)} \\
\hline
& Climatology & -0.06 & -0.01 & 2.33 $\pm$ 0.04 & 1.59 & 
4.96\\ 
& Ens mean & -0.08 & 0.01 & 2.19 $\pm$  0.04& 1.55 & 
4.57 \\ 
\multirow{-3}{*}{Baseline}
& Linear Regr &  -0.11 & -0.07 & 2.26 $\pm$  0.04 & 1.54 & 
4.72 \\
& \newold{Baseline stacking} &  \newold{0.00} & \newold{0.04} & \newold{2.15 $\pm$  0.04} & \newold{1.44} & \newold{4.55} \\
 \hline 
 \multirow{1}{*}{LR} 
& All features & -0.33 & -0.25 & 2.71 $\pm$  0.05 & 1.91 & 
5.45 \\
\multirow{1}{*}{U-Net} 
& All features &  -0.10 & -0.01 & 2.18 $\pm$  0.03 & 1.44 & 
4.62 \\ 
 \multirow{1}{*}{RF}
& All features & -0.11 & -0.01 & 2.17 $\pm$  0.05 & 1.48 & 
4.45 \\ 
\multirow{1}{*}{Stacked} & LR, U-Net, RF outputs &   \bf 0.02 & \bf 0.04 & \bf 2.07 $\pm$  0.03 & \bf 1.42 & 
\bf 4.38\\ 
\hline
\end{tabular}
\end{table}

\begin{figure}[ht]
\centerline{\includegraphics[width=27pc]{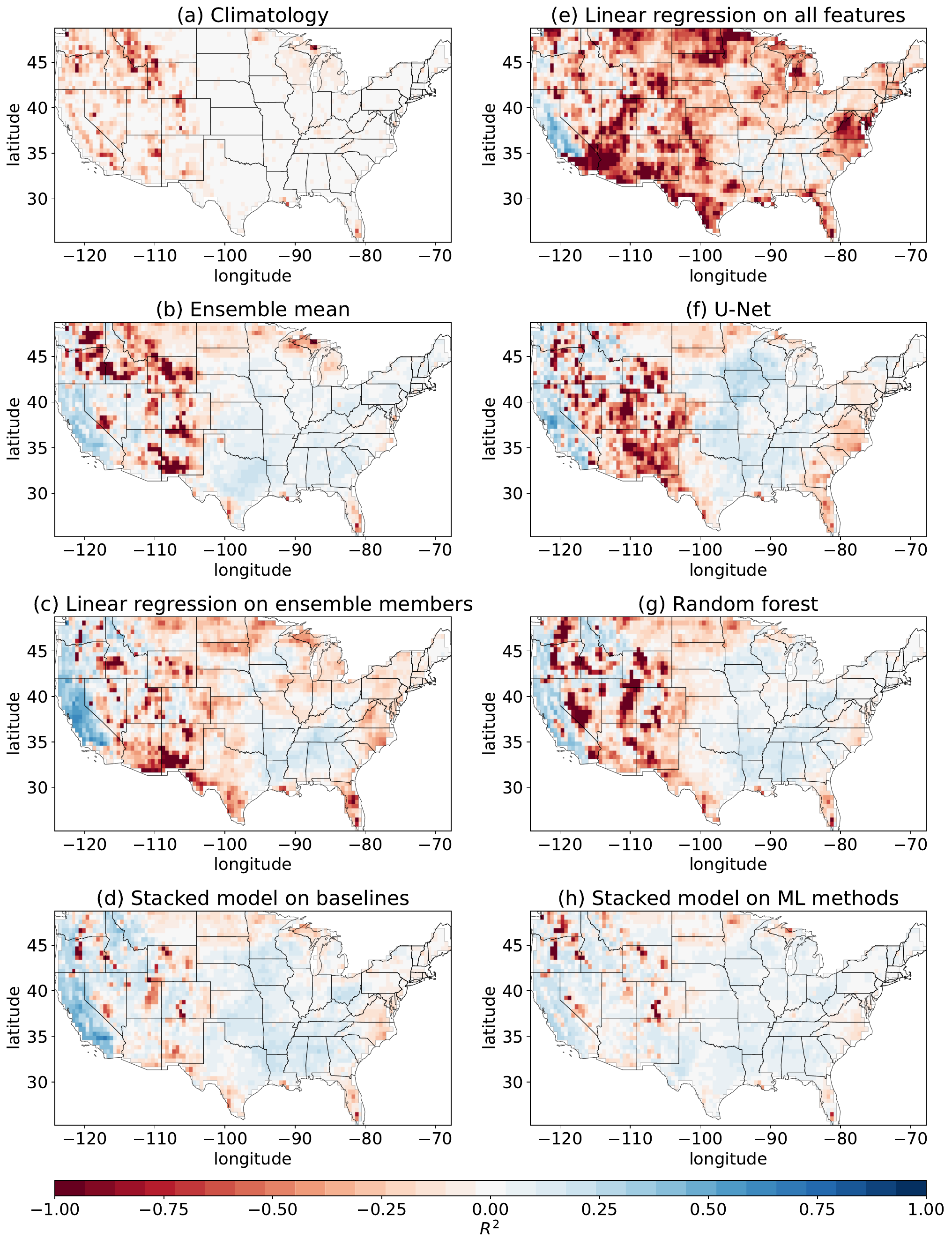}}
  \caption{\textbf{$R^2$ score heatmaps of baselines and learning-based methods for precipitation regression using  NCEP-CFSv2 ensemble members; errors are computed over the test period}. Positive values (blue) indicate better performance. \newold{See Section \ref{sec:precip_regr_ncep} for details.}}\label{fig:precip_reg_ncep_test}
\end{figure}

\Cref{fig:precip_reg_ncep_test} illustrates performance of key methods 
with $R^2$ heatmaps over the U.S. to highlight spatial variation in errors. \newold{The RF and U-Net $R^2$ fields are qualitatively similar, but they are still quite different in certain states such as Georgia, North Carolina, Virginia, Utah, and Colorado. The LR map is noticeably poor across most of the regions.}  The stacked \new{ML} model's heatmap reveals large regions where its predictive skill exceeds that of all other methods. Note that model stacking yields relatively accurate predictions even in regions where the three constituent models individually perform poorly (e.g., southwestern Arizona), highlighting the generalization abilities of our stacking approach. All methods tend to have higher accuracy on the Pacific Coast, in the Midwest, and in southern states such as Alabama and Missouri. \new{The stacking model heatmaps both} look similar. The stacking model applied to the baselines has better $R^2$ scores in California compared to the stacked ML methods. However, the stacked ML model reveals larger positive $R^2$ regions and fewer dark red spots, particularly evident in New Mexico, Minnesota, and Utah.

\paragraph{Temperature regression}\label{sec:tmp_regr_ncep}

\begin{figure}[ht]
\centerline{\includegraphics[width=27pc]{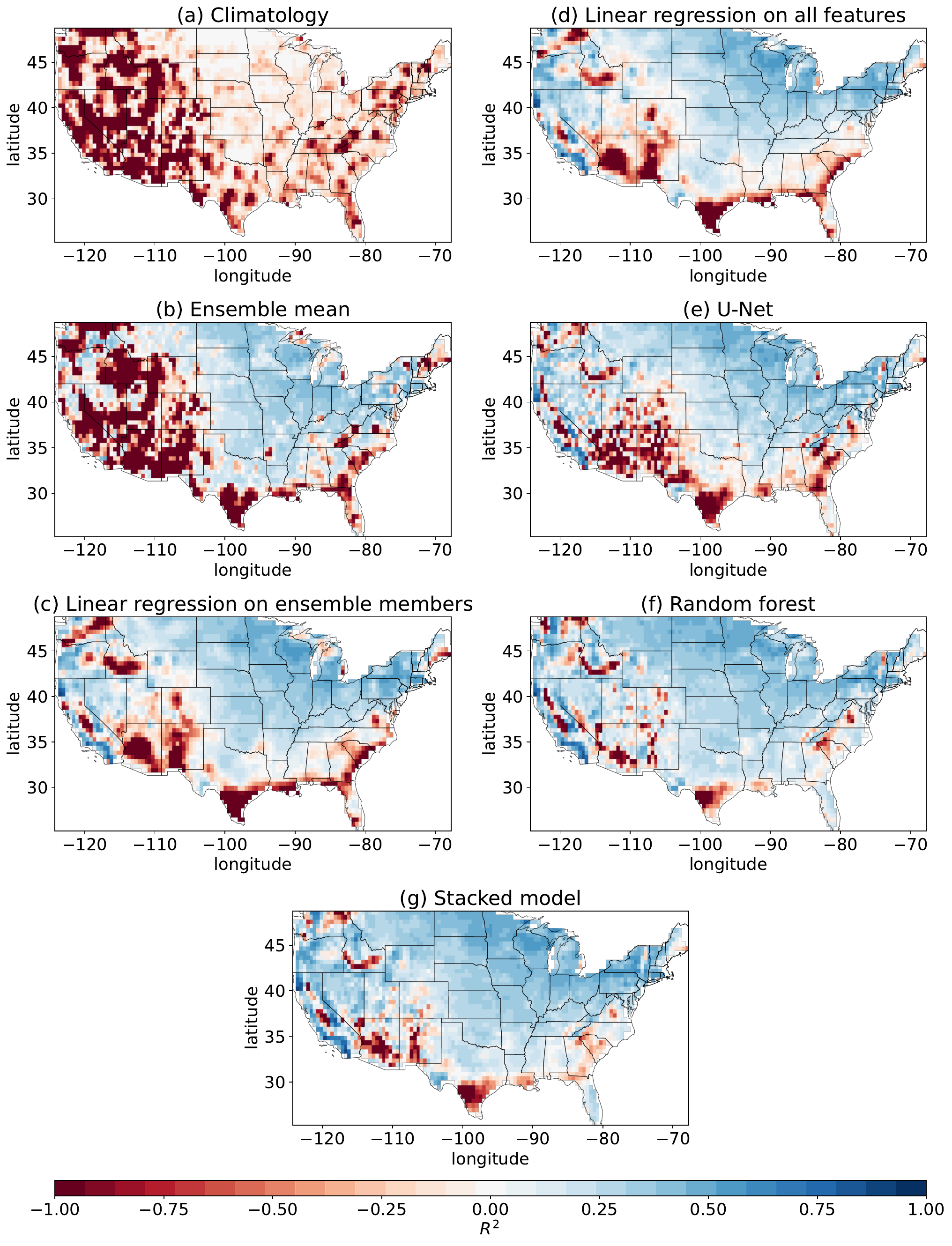}}
  \caption{\textbf{$R^2$ score heatmaps of baselines and learning-based methods for temperature regression using NCEP-CFSv2 ensemble members; errors are computed over the test period.} Positive values (blue) indicate better performance. See Section \ref{sec:tmp_regr_ncep} for details.}\label{fig:tmp_reg_ncep_test}
\end{figure}

\cref{table:tmp_regression_ncep_test} shows results for  2-meter temperature regression.
The learning-based models, especially the random forest and stacked model, significantly outperform the baseline models in terms of MSE and $R^2$ score. The random forest also outperforms linear regression and the U-Net. Note that LR, U-Net, and RF are trained without using SST information since SST features yielded worse performance over the validation period. \Cref{fig:tmp_reg_ncep_test} illustrates the performance of these methods with $R^2$ heatmaps over the U.S. As expected, the model stacking approach shows the best results across spatial locations. We notice that there are still regions such as the West, some regions in Texas, Florida, and Georgia where all models tend to achieve a negative $R^2$ score.

\begin{table}[ht!]
\caption{\textbf{Results for temperature regression using the NCEP-CFSv2 ensemble, with errors reported over the test period.} LR refers to linear regression on all features including ensemble members, lagged data, land
variables. Model stacking is performed on models that are \newold{trained} on all features except SSTs. The \textbf{best} results are in bold. \newold{MSE is reported in squared $ ^\circ C $.}} 
\label{table:tmp_regression_ncep_test}
\centering
\small
\begin{tabular}{c c  c  c  c  c  c  c  c } 
\hline
Model & Features & \makecell{Mean \\ $R^2$ ($\uparrow$)} & \makecell{Median\\ $R^2$ ($\uparrow$)}& \makecell{Mean\\ Sq Err ($\downarrow $)} & \makecell{Median \\ MSE ($\downarrow$)} & 
\makecell{90th prctl \\ MSE ($\downarrow$)} \\  
 \hline 
& Climatology &  -0.66 & -0.17 & 6.57 $\pm$ 0.11 & 5.04 & 
9.99 \\ 
 & Ens mean &  -0.47 & 0.08 & 5.51 $\pm$ 0.10 & 3.83 & 
9.16 \\ 
\multirow{-3}{*}{Baseline} 
& Linear Regr &  0.04 & 0.17 & 3.60 $\pm$ 0.03 & 3.25 & 
5.49 \\
 \hline 
 
\multirow{1}{*}{LR} 
& All features w/o SSTs  & 0.05 & 0.16 & 3.57 $\pm$ 0.02 & 3.33 & 
5.41 \\
\multirow{1}{*}{U-Net} 
& All features w/o SSTs  &   0.01 & 0.18 & 3.65 $\pm$ 0.02 &  3.38 & 
5.31 \\
 \multirow{1}{*}{RF}
& All features w/o SSTs  &  0.16 & 0.25 & 3.17 $\pm$ 0.02 & 2.99 & 
4.63 \\ 
\multirow{1}{*}{Stacked} & LR, U-Net, RF outputs &    \bf 0.18 & \bf 0.27 & \bf 3.11 $\pm$ 0.02 & \bf 2.93 & 
\bf 4.56\\ 
\hline
\end{tabular}
\end{table}

\subsection{Quantile regression}\label{sec:qregr_ncep}

We explore the use of quantile regression to predict values $z$ so that ``there's a 90\% chance that the average temperature will be below $z^\circ$ at your location next month'' -- or, equivalently, ``there is a 10\% chance that the average temperature will exceed $z^\circ$ at your location next month.'' In this sense, quantile regression focused on the $90$-th percentile 
predicts temperature and precipitation extremes, a task highly relevant to many stakeholders.
We train a linear regression model fitting the quantile loss (Linear QR), a random forest quantile regressor (RFQR), \citep{RFQR:2006}, a U-Net, and the stacked model. The Linear QR and the RFQR details are discussed in Section \ref{sec:baselines}.
The below experimental results show that temperature extremes can be predicted with high accuracy by the learning-based models (particularly our stacked model), in stark contrast to historical quantiles or ensemble quantiles in the case of temperature quantile regression.
The results for precipitation are less striking overall, though the learned models are significantly more predictive in some locations \new{on this} quantile regression task.

\paragraph{Quantile regression of precipitation}\label{sec:qregr_ncep_precip}

For each location,
the 90th percentile value is calculated based on the historical data. For the ensemble 90th percentile, we simply take the 90th percentile of the $K$ ensemble members. \cref{table:precip_qtr_ncep_test} summarizes results for precipitation quantile regression using the NCEP-CFSv2 ensemble. Our stacked model is able to significantly outperform all baselines. The performance illustration is given in Appendix \ref{sec:app_nasa_res}, \Cref{fig:precip_qtr_loss_ncep_test}.

\begin{table}[ht!]
\caption{
\textbf{Test results for precipitation quantile regression using NCEP-CFSv2 dataset, with target quantile = 0.9}. Linear QR refers to a linear quantile regressor. RFQR corresponds to a Random Forest Quantile Regressor. Model stacking is performed on models that are \newold{trained} on all features. The \textbf{best} results are in bold. \newold{Quantile loss is reported in mm.}}
\label{table:precip_qtr_ncep_test}
\centering
\small
\begin{tabular}{c c  c  c  c  c  c  c  c } 
\hline
Model & Features & \makecell{Mean \\ Qtr Loss ($\downarrow$)}  & \makecell{Median\\ Qtr Loss ($\downarrow$)}& \makecell{90th prctl \\Qtr Loss ($\downarrow$)} \\ %
 \hline 
& Historical 90th percentile & 0.304 $\pm$ 0.003& 0.278  & 0.504\\ 

& Ens 90th percentile & 0.311 $\pm$ 0.003& 0.275  & 0.488\\ 
\multirow{-3}{*}{Baseline}
& Linear QR ens only & 0.310 $\pm$ 0.003& 0.266  & 0.505\\ 
\hline 
 
\multirow{1}{*}{Linear QR} 
& All features &  0.287 $\pm$ 0.003& 0.248 &0.463 \\
 \multirow{1}{*}{U-Net} 
& All features &  0.312 $\pm$ 0.002& 0.281 &0.504 \\
 \multirow{1}{*}{RFQR}
& All features & \bf 0.282 $\pm$ 0.002& 0.257  & \bf 0.453 \\ 
 \multirow{1}{*}{Stacked} & U-Net, RFQR, LQR outputs &   \bf 0.282 $\pm$ 0.002& \bf 0.256  & 0.457 \\ 
\hline
\end{tabular}
\end{table}

\paragraph{Quantile regression of temperature}\label{sec:qregr_ncep_tmp}

\begin{figure}[ht]
  \centerline{\includegraphics[width=25pc]{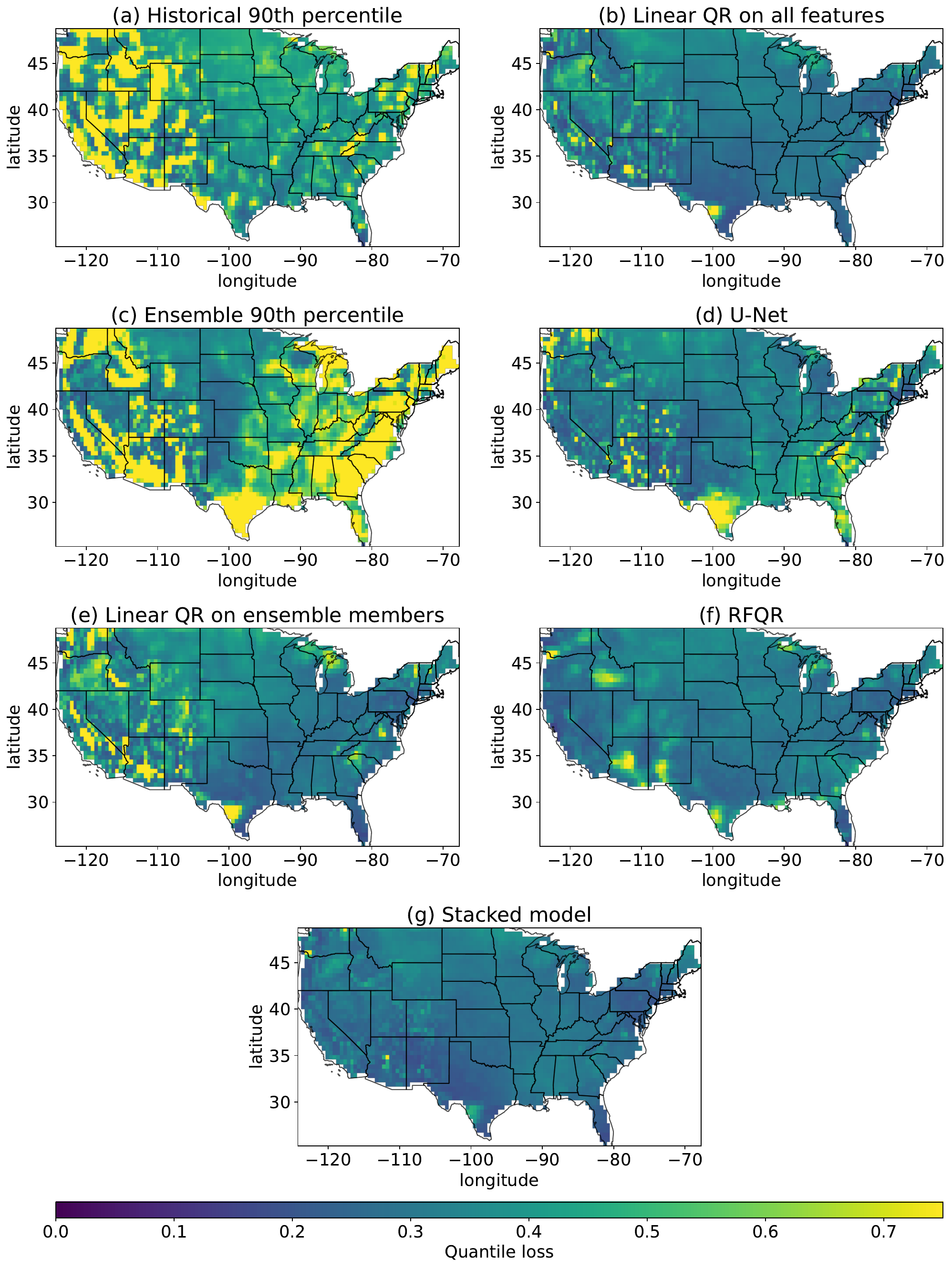}}
  \caption{\textbf{Test quantile loss heatmaps of baselines and learning-based methods for temperature quantile regression using NCEP-CFSv2 dataset}. Blue regions indicate smaller quantile loss. See Section \ref{sec:qregr_ncep_tmp} for details.}
  \label{fig:tmp_qtr_loss_ncep_test}
\end{figure}

\cref{table:tmp_qtr_ncep_test} summarizes results for temperature quantile regression using the NCEP-CFSv2 ensemble. Note that we do not include SST  features for temperature quantile regression in our learned models. We observe that all of our learned models are able to significantly outperform all baselines. In \Cref{fig:tmp_qtr_loss_ncep_test}, 
we show the heatmaps of quantile loss of baselines and our learned models. We observe that the learned models produce \new{predictions with varied quality}, and the stacked model can pick up useful information from them. For example, in Arizona and Texas, the Linear QR, U-Net, and RFQR show some errors but in different locations, and the stacked model can exploit the advantages of each model.

\begin{table}[ht!]
\caption{\textbf{Test results for temperature quantile regression using NCEP-CFSv2 dataset, with target quantile = 0.9}. Linear QR refers to a linear quantile regressor. RFQR corresponds to a Random Forest Quantile Regressor. Model stacking is performed on models that are \text{trained} on all features except for SSTs.
Learned models can predict highly likely temperature ranges accurately, meaning there are fewer unpredicted temperature spikes. The \textbf{best} results are in bold. Quantile loss is reported in $^{\circ}C$.}
\label{table:tmp_qtr_ncep_test}
\centering
\small
\begin{tabular}{c c  c  c  c  c  c  c  c } 
\hline
Model & Features & \makecell{Mean \\ Qtr Loss ($\downarrow$)} & \makecell{Median\\ Qtr Loss ($\downarrow$)}& \makecell{90th prctl \\Qtr Loss ($\downarrow$)} \\ %
 \hline 
& Historical 90th percentile & 0.589 $\pm$ 0.008& 0.435  & 0.980\\ 

& Ens 90th percentile & 0.642 $\pm$ 0.009& 0.468& 1.196\\
\multirow{-3}{*}{Baseline}
& Linear QR ens only & 0.336 $\pm$ 0.004& 0.286  & 0.488\\ 
 
\hline 
\multirow{1}{*}{Linear QR} 
& All features w/o SSTs  &  0.318 $\pm$ 0.002& 0.301 &0.407 \\ 
\multirow{1}{*}{U-Net} 
& All features w/o SSTs  &  0.363 $\pm$ 0.003& 0.329  &0.488 \\
 \multirow{1}{*}{RFQR}
& All features w/o SSTs  & 0.320 $\pm$ 0.002& 0.307  & 0.384 \\ 
 \multirow{1}{*}{Stacked} & U-Net, RFQR, LQR outputs &   \bf 0.287 $\pm$ 0.001& \bf 0.285  & \bf0.344\\ 
\hline
\end{tabular}
\end{table}

\section{Discussion} \label{section:discussion}
\subsection{The efficacy of machine learning for SSF} \label{sec:ens_members_analysis}

Several hypotheses might explain why ML \new{may be} a \new{promising} approach for SSF, and we probe those hypotheses in this section.

\paragraph{Using full ensemble vs.\ ensemble mean}\label{sec:full_vs_avg}
Past works use ensemble mean as an input feature to machine learning methods in addition to the climate variables \citep{rodeo, he2021learning}. 
Ensembles provide valuable information not only about expected climate behavior but also variance or uncertainty in multiple dimensions; methods that rely solely on ensemble mean lack information about this variance. \new{Ensemble members may have systematic errors, either in the mean or the variability, arising from different initial conditions of the corresponding dynamic model that are not readily apparent to users.}
The more recently initialized an ensemble member is, the better it usually performs. While taking the average of these ensemble members may cancel out the deficiencies of each individual member, it is also possible that details of each member's systematic errors may be directly discovered and corrected independently by a machine learning model. Therefore, using a single ensemble statistic, such as the ensemble mean, as a feature may not fully capitalize on the information provided by using all \new{members of the lagged ensemble} as features. 

\begin{figure}[!hbt]
\centerline{\includegraphics[width=33pc]{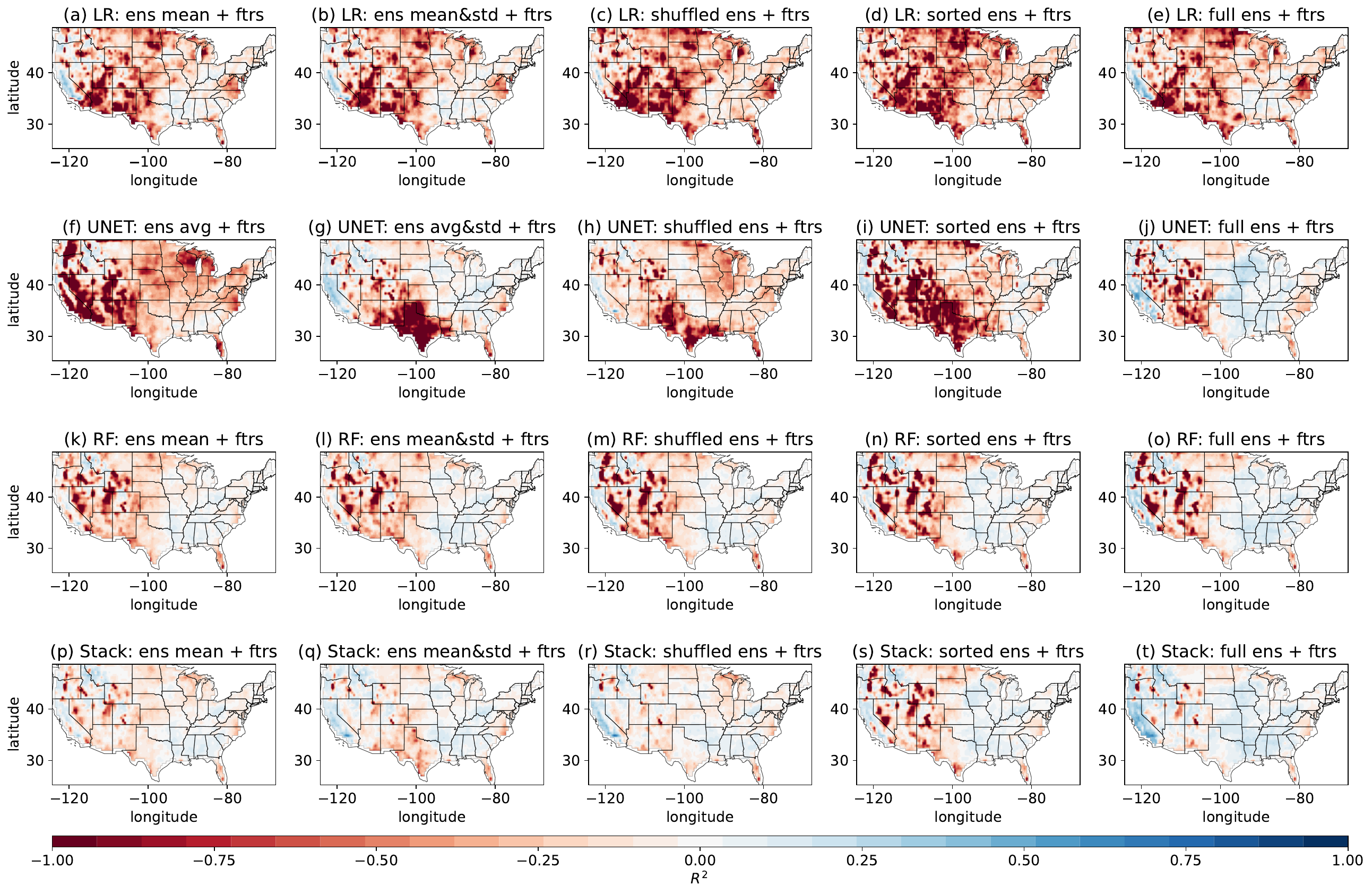}}
  \caption{\textbf{Precipitation regression test $R^2$ heatmaps of LR, U-Net, RF, and stacked model trained using ensemble mean only, using sorted and shuffled ensemble, or using the full ensemble.} The NCEP-CFSv2 ensemble is used. See Section \ref{sec:ens_members_analysis} for details.
  }\label{fig:ens_avg_test_precip_ncep}
\end{figure}

\begin{figure}[!hbt]
\centerline{\includegraphics[width=33pc]{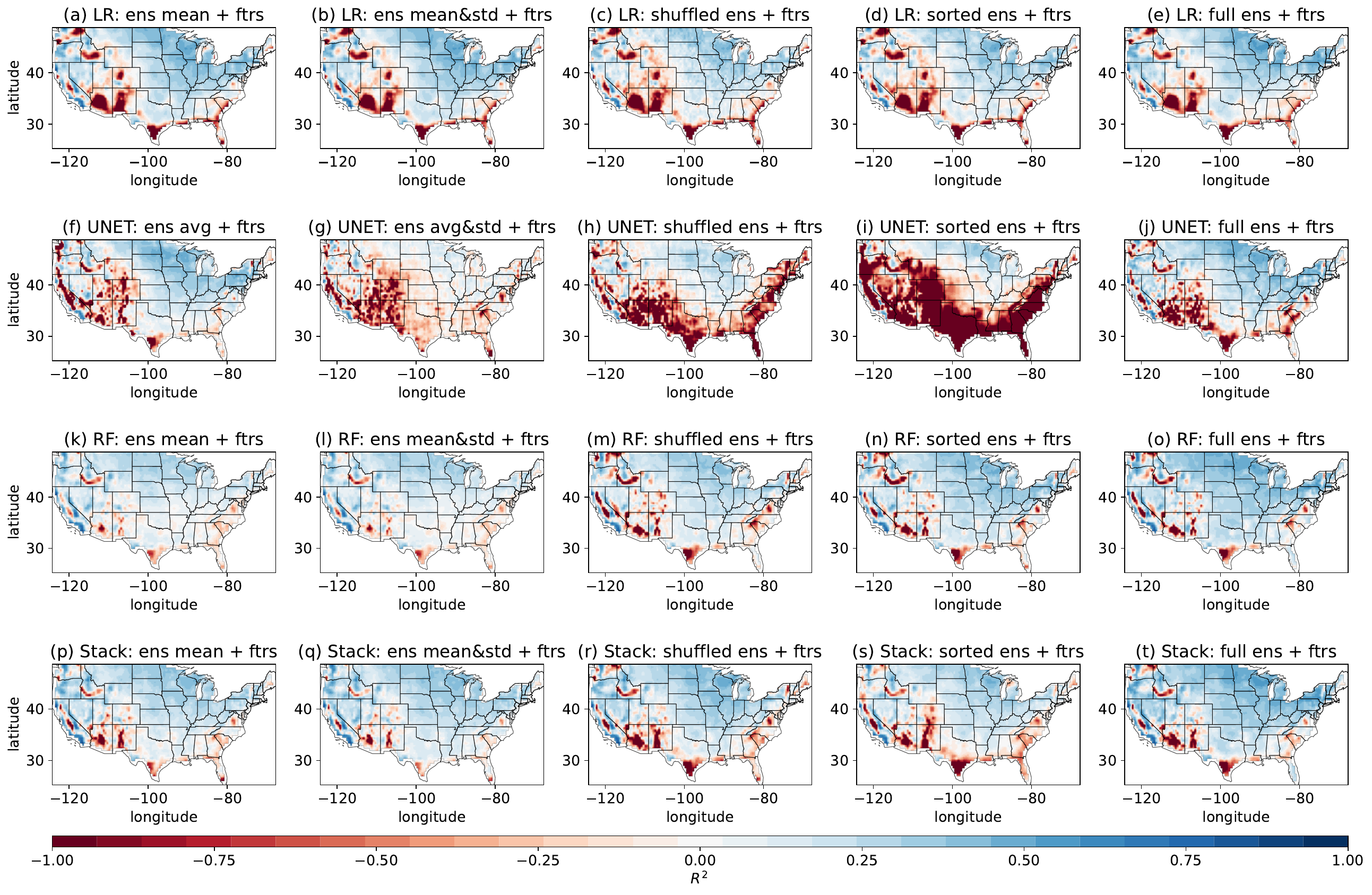}}
  \caption{\textbf{Temperature regression test $R^2$ heatmaps of LR, U-Net, RF and stacked model trained using ensemble mean only, using sorted and shuffled ensemble, or using the full ensemble.} The NCEP-CFSv2 ensemble is used. See Section \ref{sec:ens_members_analysis} for details.}\label{fig:ens_avg_test_tmp_ncep}
\end{figure}

In our experiments, we find that using all available ensemble members enhances the prediction quality of our approaches. As an illustration, we show the results of the LR, RF, U-Net, and the stacked model trained on all ensemble members, compared to the ML models trained on the ensemble mean only.
In addition to the full ensemble or the ensemble mean, we use other available features (as in our previous regression results). \cref{table:comparing_ens_avg_vs_all_test} and \Cref{table:comparing_ens_avg_vs_all_test_tmp}  demonstrate the precipitation and temperature forecasting results. For the linear regression, utilizing the ensemble mean with all other features produces the best test performance compared to using the full ensemble. Such behavior is not surprising for the LR since the full ensemble incorporates large variance across ensemble members, which may result in a worse linear fit. For the U-Net, RF, and stacked model, we observe significant performance improvements, in terms of having at least one standard error smaller MSE, when using the full ensemble instead of using the ensemble mean. When we compare the performance of learned models using only the ensemble mean to that of the learned models that use both the ensemble mean and the ensemble standard deviation for each spatial location, we find that the addition of the standard deviation feature does not provide enough information to significantly improve the performance of ML models, and in fact the U-Net that exhibits a performance degradation -- a potential a sign of overfitting. These observations are visually supported by Figure \ref{fig:ens_avg_test_precip_ncep} and \Cref{fig:ens_avg_test_tmp_ncep}, where the $R^2$ heatmaps of our methods (except the U-Net) utilizing ensemble mean and standard deviation closely resemble the performance of methods solely relying on ensemble mean. We conclude that the full ensemble contains important information for SSF aside from the ensemble mean, and our models can capitalize on this information for precipitation and temperature forecasting.

We can perform a statistical test to verify that the performance discrepancies between using the ensemble mean and using the full ensemble are statistically significant for the stacked model. As before, let $\hat y_{t,l}$ refer to the estimate under our usual stacked model (i.e., with all ensemble members). Let $\hat y_{t,l}^{\text{SEA}}$ refer to a stacked model with just the ensemble mean \new{as a feature, instead of all ensemble members}. We can employ a sign test framework \citep{delsole2014comparing, cash2019evaluation} to compare model performance under minimal distributional assumptions. Namely, we only make the following i.i.d. assumption over the time dimension: 
\[
\ind{|\hat y_{t,l}-y_{t,l}| < |\hat y_{t,l}^{\text{SEA}}-y_{t,l}|} \overset{\text{iid}}{\sim}\mathrm{Bernoulli}(p_{l})
\]
Intuitively, this corresponds to assuming it is a coin flip which model will perform better at each time point and location, and we would like to test whether each location's ``coin'' is fair or not. We can then formulate our null and alternative hypotheses for each location $l$ as follows:
\[
H_{0,l}: p_{l}=0.5,\;\; H_{0,l}: p_{l}>0.5
\]
Thus, our overall test for significance is for the global null hypothesis $H_0 = \cap_{l=1}^{3274} H_{0,l}$. We calculate a p-value for each $H_{0,l}$, and then we check whether any of these p-values is below a Bonferroni-corrected threshold of $0.05/3274=1.53\times10^{-5}$, where 3274 refers to the number of locations. In fact, the minimum p-values for this test with precipitation and regression alike are far below this threshold ($1.68\times10^{-10}$ and $4.42\times10^{-24}$, respectively). This allows us to reject the global null hypothesis for both temperature and precipitation, and we conclude that including the full ensemble in our stacked model significantly outperforms including just the ensemble mean. 

\begin{table}[ht!]
\caption{\textbf{Precipitation forecasting performance comparison of the LR, RF, U-Net, and stacked model trained using the ensemble mean, using the sorted ensemble members, or using the original ensemble, in addition to other features}. Scores on the test data are reported, and NCEP-CFSv2 data is used. The \textbf{best} results are in bold. MSE is reported in squared mm.}
\label{table:comparing_ens_avg_vs_all_test}
\centering
\small
\begin{tabular}{c c  c  c  c  c  c  c  c} 
 \hline
Model & Features & \makecell{Mean \\ $R^2$ ($\uparrow$)} & \makecell{Mean \\ Sq Err ($\downarrow$)}\\
 \hline
& Ensemble mean + all features &  \bf -0.28 & \textbf{2.59}$\pm$0.04 \\  
\multirow{-2}{*}{LR}
& Ensemble mean \& std + all features &  -0.29 & 2.61 $\pm$ 0.04 \\ 
 & Shuffled ensemble + all features  & -0.41 & 2.84$\pm$0.05 \\ 
 & Sorted ensemble + all features  & -0.43 & 2.87$\pm$0.05 \\ 
 & Full ensemble + all features & -0.33 & 2.71$\pm$0.05\\ 
 
 \hline
& Ensemble mean + all features & -0.45 & 2.76$\pm$0.04 \\  
\multirow{-2}{*}{\newold{U-Net}}
& Ensemble mean \& std + all features &  \newold{-0.25} & 2.65 $\pm$ 0.04 \\ 
& Shuffled ensemble + all features  & \newold{-0.27} & 2.77$\pm$0.05 \\ 
 & Sorted ensemble + all features  & -0.43 & 2.78$\pm$0.04 \\ 
 & Full ensemble + all features & \bf -0.1 & \textbf{2.18}$\pm$0.03\\ 
 
  \hline
& Ensemble mean + all features &  -0.16 & 2.36$\pm$0.04 \\  
\multirow{-2}{*}{\newold{RF}}
& Ensemble mean \& std + all features &  -0.15 & 2.32 $\pm$ 0.04 \\
& Shuffled ensemble + all features  & -0.16 & 2.29$\pm$0.04 \\ 
 & Sorted ensemble + all features  & -0.18 & 2.30$\pm$0.04 \\ 
 & Full ensemble + all features & \bf -0.11 & \textbf{2.17}$\pm$0.05\\ 
 
  \hline
& Ensemble mean + all features &  -0.08 & 2.28$\pm$0.04 \\  
\multirow{-2}{*}{Stacked}
& Ensemble mean \& std + all features &  -0.05 & 2.26 $\pm$ 0.04 \\
& Shuffled ensemble + all features  & -0.04 & 2.25$\pm$0.04 \\
 & Sorted ensemble + all features  & -0.11 & 2.24$\pm$0.04 \\ 
 & Full ensemble + all features &\bf 0.02 & \textbf{2.07}$\pm$0.03\\ 
 
\hline
\end{tabular}
\end{table}

\begin{table}[ht!]
\caption{\textbf{Temperature forecasting performance comparison of the LR, RF, U-Net, and stacked model trained using the ensemble mean, using the sorted ensemble members, or using the original ensemble, in addition to other features.} Scores on the test data are reported and NCEP-CFSv2 data is used. The \textbf{best} results are in bold. MSE is reported in squared $ ^\circ C $.}
\label{table:comparing_ens_avg_vs_all_test_tmp}
\centering
\small
\begin{tabular}{c c  c  c  c  c  c  c  c} 
 \hline
Model & Features & \makecell{Mean \\ $R^2$ ($\uparrow$)} & \makecell{Mean \\ Sq Err ($\downarrow$)}\\
 \hline
& Ensemble mean + all features &  \bf 0.06 & \textbf{3.55}$\pm$0.03 \\  
\multirow{-2}{*}{LR}
& Ensemble mean \& std + all features & 0.05 & \newold{3.59$\pm$0.03} \\
& Shuffled ensemble + all features  & 0.03 & \newold{3.95$\pm$0.03} \\  
 & Sorted ensemble + all features  & -0.02 & 3.87$\pm$0.03 \\ 
 & Full ensemble + all features & 0.05 & 3.57$\pm$0.02\\ 
 
 \hline
& Ensemble mean + all features & 0.00 & 3.77$\pm$0.03 \\  
\multirow{-2}{*}{U-Net}
& Ensemble mean \& std + all features &  0.19 & \newold{4.61$\pm$0.03} \\
& Shuffled ensemble + all features  & -0.29 & 4.75$\pm$0.03 \\ 
 & Sorted ensemble + all features  & -0.94 & 6.51$\pm$0.05\\ 
 & \newold{Full ensemble + all features} & \bf0.01 & \textbf{3.65}$\pm$0.02\\ 
 
  \hline
& Ensemble mean + all features &  0.10 & 3.57$\pm$0.02 \\  
\multirow{-2}{*}{RF}
& Ensemble mean \& std + all features &  0.10 &  3.56 $\pm$0.02 \\
& Shuffled ensemble + all features  & 0.05 & 3.65$\pm$0.02 \\
 & Sorted ensemble + all features  & 0.10 & 3.44$\pm$0.02 \\ 
 & \newold{Full ensemble + all features} & \bf 0.16 & \textbf{3.17}$\pm$0.02\\ 
 
  \hline
& Ensemble mean + all features &  0.11 & 3.43$\pm$0.02 \\  
\multirow{-2}{*}{Stacked}
 & Ensemble mean \& std + all features &  0.13 &  3.30 $\pm$0.02 \\
 & Shuffled ensemble + all features  & 0.08 & 3.47$\pm$0.02 \\
 & Sorted ensemble + all features  & 0.03 & 3.70$\pm$0.02 \\ 
 & Full ensemble + all features &\bf 0.18 & \textbf{3.11}$\pm$0.02\\ 
 
\hline
\end{tabular}
\end{table}

\paragraph{Sensitivity to ensemble formulation} \label{sec:when_to_trust}

We consider the hypothesis that there is a set of $k$ ensemble members that are always best. To test this hypothesis, we use a training period to identify which $k$ members perform best for each location, and then during the test period, compute the average of only these $k$  ensemble members. The performance of this approach depends on $k$, the number of ensemble members we allow to be designated ``good.'' \new{We have not found that} the performance for any $k$ exhibits a significant improvement over the ensemble mean. 

If the ensemble members have different levels of accuracy over various seasons, locations, and conditions, then a machine learning model may be learning when to ``trust" each member. We know that our ensemble members are lagged, meaning they are initialized at different times. We believe each ensemble member encapsulates valuable information derived from the underlying physical model during each initialization. To investigate the impact of ensemble member order, we perform the following experiment: we randomly permute ensemble members at every time step $t$ for all locations (preserving the spatial information) and apply our ML models to these shuffled ensembles. From \Cref{table:comparing_ens_avg_vs_all_test} and \Cref{table:comparing_ens_avg_vs_all_test_tmp}, this approach negatively affects the performance of the ML models compared to using the full ensemble with the original order. 
One possible explanation is that the learned models lose the ability to learn \new{which ensemble member to trust, as this information is tied to the initialization time of each ensemble member. Even though the spatial information remains intact after the shuffling, the models can no longer exploit dependencies associated with the original ensemble structure.}

\new{Additionally, we conduct an experiment designed to test whether it is important to keep track of which ensemble member made each prediction or whether it is the \textit{distribution} of predictions that is important. The modeling approach for the former would be to feed in ensemble member 1's forecast as the first feature, ensemble member 2's forecast as the second feature, etc. The modeling approach under the distributional hypothesis is to make the smallest prediction be the first feature, the second-smallest prediction be the second feature, and so on -- i.e., we sort the ensemble forecasts for each location separately. Note that this entails treating the ensemble members symmetrically: the model would give the same prediction if ensemble member 1 predicted $a$ and ensemble member 2 predicted $b$ or if ensemble member 1 predicted $b$ and ensemble member 2 predicted $a$.
In statistical parlance, this is passing in the order statistics of the forecasts as the features rather than their original ordering. Note that for NCEP-CFSv2, ensemble forecasts are originally ordered according to the time their initial conditions are set \citep{saha2014ncep}.}
According to \Cref{table:comparing_ens_avg_vs_all_test} and \Cref{table:comparing_ens_avg_vs_all_test_tmp}, using the sorted ensemble drastically degrades U-Net's performance, which is essentially because we sort the ensemble members for each location individually, and sorting the ensemble members individually for each location may \new{hamper the ability for the U-Net to learn spatial structure}.  In the case of precipitation regression with the stacking model from \Cref{table:comparing_ens_avg_vs_all_test}, the MSE of the sorted approach is 2.24, which is worse than the 2.07 MSE for using the original ordering. In the case of temperature forecasting from \Cref{table:comparing_ens_avg_vs_all_test_tmp}, the MSE of the sorted approach is 3.70, which is much worse than the 3.11 MSE for using the original ordering. The mean $R^2$ of the sorted approach is also lower compared to the original ordering. In both cases, the performance is better when we feed in the features in such a way that the machine learning model has an opportunity to learn aspects of each ensemble member, not merely their order statistics. Therefore, imposing a symmetric treatment of ensemble members degrades performance. 
\Cref{fig:ens_avg_test_precip_ncep} and \Cref{fig:ens_avg_test_tmp_ncep} shows the corresponding $R^2$ heatmaps of our models for precipitation and temperature regression tasks. 

\subsection{Using spatial data}\label{sec:disc_spatial_data}
There are several ways to incorporate information about location in our models. U-Net has access to spatial dependencies through its design. Specifically, our U-Net inputs the spatial location of each point in the map. Naively, we might represent each location using the latitude and longitude values. Alternatively, we may use positional encoding, which is known to be beneficial in many ML areas, not only in NLP (as we mention in Section \ref{sec:PE}). PE captures the order (or position) and allows one to learn the contextual relationships (local context -- relationships between nearby elements and global context dependencies across the entire sequence). We assume that the PE approach represents spatial information in a manner more accessible to our learned models. 

\begin{figure}[ht]

\centerline{\includegraphics[width=33pc]{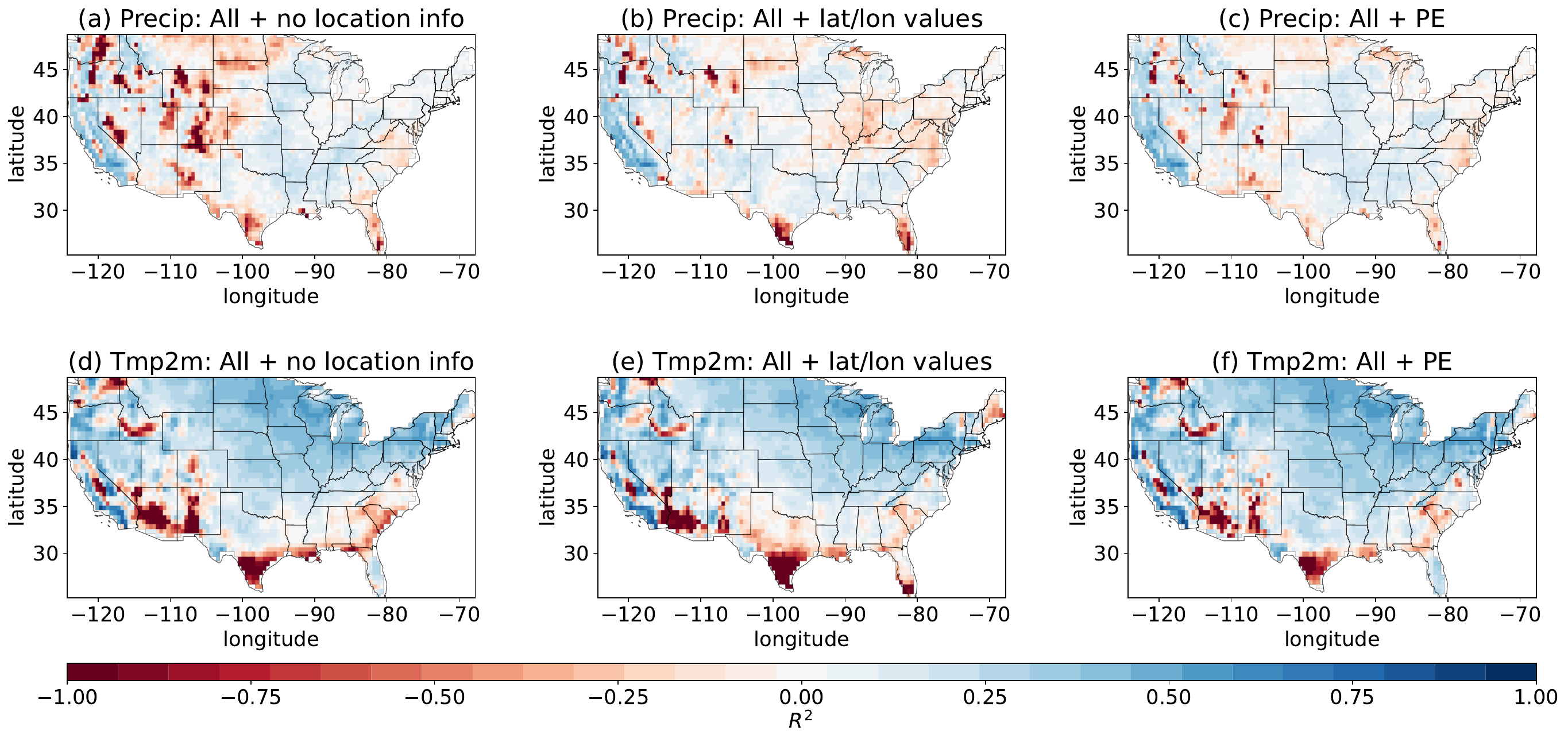}}
  \caption{\newold{\textbf{Test $R^2$ heatmaps of the stacked model of LR, RF, and U-Net trained using no spatial features, using latitude and longitude values or using PE.} The NCEP-CFSv2 ensemble is used. See section \ref{sec:disc_spatial_data} for details.
}
  }\label{fig:lat_lon_test_stacked}
\end{figure}

As an illustration, \cref{table:compare_pe_lat} and \Cref{fig:lat_lon_test_stacked} demonstrate the performance of a stacked model using LR, RF, and U-Net trained using positional encoding, using latitude and longitude values and using no features representing the spatial information (no PE and no latitude or longitude values). Other inputs to the LR, RF, and U-Net models are ensemble member forecasts, lagged target variable, climate variables, and SSTs (\new{except} in the case of temperature forecasting). 
The results suggest that using PE enhances the predictive skill of our models, compared to using just the lat/lon values or no location information, especially for the temperature forecasting task. Using no information about locations hurts the performance of precipitation regression. Thus, our models can account for spatial dependencies using input features, and PE is more beneficial than the raw latitude and longitude information. These findings on PE effectiveness are consistent with prior findings in ML. For example, \cite{wu2021rethinking} investigate the efficacy of PE in the context of the visual transformer model used for image classification and object detection. We show a more detailed analysis with results for the LR, RF, and U-Net in Appendix \ref{app:lat_lon_test}.

\begin{table}[!ht]
\caption{\textbf{Test performance comparison of the stacked model of LR, RF, and U-Net trained using no spatial features, using latitude and longitude values, or using PE. } Utilizing spatial representations, including PE, latitude, and longitude values, helps advance the predictive skill. Furthermore, using positional encoding is more beneficial than using raw latitude and longitude values. The \textbf{best} results are in bold. MSE is reported in squared mm for precipitation and in squared $ ^\circ C $ for temperature.}
\label{table:compare_pe_lat}
\centering
\small
\begin{tabular}{c c  c  c  c  c  c  c  c} 
 \hline
Target & Features & \makecell{Mean \\ $R^2$ ($\uparrow$)} & \makecell{Mean \\ Sq Err ($\downarrow$)}\\
 \hline
& All + no location info &   -0.05 & 2.13$\pm$0.03 \\  
\multirow{-2}{*}{Precip}
 & All + lat/lon values  & -0.01 & 2.21$\pm$0.04 \\ 
 & All + PE & \bf 0.02 & \textbf{2.07}$\pm$0.03 \\ 
 
\hline
 & All + no location info &  0.12 &3.35$\pm$0.02  \\  
\multirow{-2}{*}{Tmp}
 & All + lat/lon values  &  0.12 & 3.33$\pm$0.02 \\ 
 & All + PE &  \bf 0.18 & \textbf{3.11}$\pm$ 0.02\\ 
\hline
\end{tabular}
\end{table}

\subsection{Variable importance}\label{sec:variable_emportance}
One consideration when implementing ML for SSF is that ML models can incorporate side information (such as spatial information, lagged temperature and precipitation values, and climate variables). In this section, we explore the importance of the various components of side information. 
We see that including the observational climate variables improves the performance of the random forest and the U-Net when doing precipitation regression. Furthermore, including positional encoding of the locations improves the performance of the U-Net, while the principle components of the sea surface temperature do not make a notable difference in the case of temperature prediction. 

More specifically, \cref{table:precip_regression_ncep_ftrs} summarizes grouped feature importance of precipitation regression using the NCEP-CFSv2 ensemble. We observe that models, in particular random forest and U-Net,  trained on all available data achieve the best performance. In the case of linear regression, the SSTs features are neither very helpful nor actively harmful. Therefore, to be consistent, we use predictions of these models trained on all features as input to the stacking model.

\begin{table}[!ht]
\caption{\textbf{Grouped feature importance results on validation for precipitation regression task using NCEP-CFSv2 ensemble members.}  The results suggest that using additional observational information helps to improve the performance of learning-based models for this task. \ditto means a repetition of features that are used above. For example, in the U-Net part of the table, ``\ditto \& lags'' means that ensemble members, PE, and lags are used as features and ``\ditto \& SSTs'' means ensemble members, PE and lags, land features and SSTs are used as features. The \textbf{best} results are in bold. \newold{MSE is reported in squared mm.}}
\label{table:precip_regression_ncep_ftrs}
\centering
\small
\begin{tabular}{c c  c  c  c  c  c  c  c} 
 \hline
Model & Features & \makecell{Mean \\ $R^2$ ($\uparrow$)} & \makecell{Median\\ $R^2$ ($\uparrow$)} & \makecell{Mean\\ Sq Err ($\downarrow$)} & \makecell{Median \\ MSE ($\downarrow$)}  &  
\makecell{90th prctl \\ MSE ($\downarrow$)}\\
 \hline
\multirow{1}{*}{LR} 
& Ens members & -0.13 & -0.08 & 2.11 $\pm$ 0.03 & 1.53 & 
4.63 \\  
 & \ditto \&  lags &  -0.11 & -0.07 & 2.10 $\pm$ 0.03 & 1.50 & 
4.59 \\ 
& \ditto \& climate variables (no SSTs) &  \textbf{-0.09} & \textbf{-0.06} & \textbf{2.06} $\pm$ 0.03 & \bf 1.47 & 
\bf 4.52 \\ 
 & \ditto \& SSTs&  -0.10 & -0.07 & 2.08 $\pm$ 0.03 & 1.47 & 
4.61 \\ 
 \hline
\multirow{1}{*}{U-Net} 
&  Ens members with  PE & -0.13 & -0.05 & 2.01  $\pm$ 0.03 & 1.50  & 
4.31\\ 
 & \ditto \& lags  & -0.08 & -0.02 & 1.92 $\pm$ 0.03 & 1.42  & 
4.17 \\ 
& \ditto \& climate variables (no SSTs) &  -0.02 & 0.05 & 1.86 $\pm$ 0.03 & 1.37  & 
4.02 \\ 
 & \ditto \& SSTs& \textbf{0.00} & \textbf{0.05} & \textbf{1.83} $\pm$ 0.03 & \bf 1.34 & 
\bf 3.94 \\ 
\hline
\multirow{1}{*}{RF}
 & Ens members with PE & -0.15 & -0.04 & 2.02 $\pm$ 0.03 & 1.49 & 
4.34 \\  
 & \ditto \& lags & -0.10 & 0.00 & 1.96 $\pm$ 0.03 & 1.44 & 
4.21 \\ 
& \ditto \& climate variables (no SSTs) & -0.08 & 0.02 & 1.93 $\pm$ 0.03 & 1.39 & 
4.16\\ 
& \ditto \& SSTs & \textbf{-0.06} & \textbf{0.04} &  \textbf{1.89} $\pm$ 0.03 & \bf 1.36 & 
\bf 4.08 \\ 
\hline
\end{tabular}
\end{table}

\cref{table:tmp_regression_ncep_ftrs} 
summarizes grouped feature importance of temperature regression using the NCEP-CFSv2 ensemble. In this case, adding some types of side information may yield only very small improvements to predictive skill, and in some cases, the additional information may decrease predictive skill. On the one hand, this effect can be explained by different training set sizes for different models: as we outline in Section \ref{sec:spatial}, the training set size for RF is $n=TL$, while $n=T$ for U-Net. This effect also may be a sign of overfitting, as temperature forecasting presents a comparatively less complex challenge than precipitation forecasting. We also note that SSTs provide only marginal (if any) improvement in predictive skill, in part because Pacific SSTs are less helpful away from the western U.S. \citep{mamalakis2018new, seager2007model}. It could also be that information from the SSTs is already being well-captured by the output from the dynamical models, and thus, including observed SSTs does not provide much additional information. In order to be consistent, we use predictions of these models trained on all features except SSTs as input to the stacking model.

\begin{table}[ht!]
\caption{\textbf{Grouped feature importance results on validation for temperature regression task using NCEP-CFSv2 ensemble members.} The results demonstrate that using some additional information may yield only very small improvements in predictive skill, and in some cases, the side information may decrease predictive skill. \ditto means a repetition of features that are used above. For example, in the U-Net part of the table, ``\ditto \& lags'' means that ensemble members, PE, and lags are used as features and ``\ditto \& SSTs'' means ensemble members, PE and lags, land features and SSTs are used as features. The \textbf{best} results are in bold. \newold{MSE is reported in squared $ ^\circ C $.}}
\label{table:tmp_regression_ncep_ftrs}
\centering
\small
\begin{tabular}{ c c  c  c  c  c  c  c  c} 
 \hline
Model & Features & \makecell{Mean \\ $R^2$ ($\uparrow$)} & \makecell{Median\\ $R^2$ ($\uparrow$)}& \makecell{Mean\\ Sq Err ($\downarrow $)} & \makecell{Median \\ MSE ($\downarrow$)} &  
\makecell{90th prctl \\ MSE ($\downarrow$)}\\ \hline 
\multirow{1}{*}{LR}
& Ens members & 0.35 & 0.40 & 2.19 $\pm$ 0.02 & 2.00 & 
3.47 \\ 
& \ditto \& lags & \bf 0.37 & \bf 0.40 & \bf 2.12 $\pm$ 0.02 & \bf 1.94 & 
\bf 3.30 \\ 
& \ditto \& climate variables (no SSTs) & 0.36 & 0.39 & 2.14 $\pm$ 0.04 & 1.94 & 
3.40\\ 
& \ditto \& SSTs& 0.34 & 0.38 & 2.23 $\pm$ 0.02 & 1.99 & 
3.73 \\ 
 \hline 
\multirow{1}{*}{U-Net} 
&  Ens members with  PE & \bf 0.33 & \bf 0.41 & \bf 2.22 $\pm$ 0.04& \bf 2.02 & 
\bf 3.47\\ 
& \ditto \& lags  &  0.32 & 0.40 & 2.24 $\pm$ 0.02 & 2.02 & 
3.49 \\ 
& \ditto \& climate variables (no SSTs) &   0.31 & 0.41 & 2.26 $\pm$ 0.02 & 2.08 & 
3.48 \\ 
& \ditto \& SSTs& 0.28 & 0.38 & 2.47 $\pm$ 0.02 & 2.20 & 
3.95 \\ 
\hline 
\multirow{1}{*}{RF}
 & Ens members with PE & 0.11 & 0.37 & 2.85  $\pm$ 0.04 & 2.28 & 
4.87 \\  
 & \ditto \& lags &   0.30 & \ 0.36 & 2.35 $\pm$ 0.02 & 2.12 & 
3.70\\ 
 & \ditto \& climate variables (no SSTs) &  \bf 0.30 & \bf 0.36 & \bf 2.33 $\pm$ 0.02 & \bf 2.10 & 
\bf 3.65\\ 
 & \ditto \& SSTs & 0.28 & 0.34 & 2.42 $\pm$ 0.02 & 2.17 & 
3.83 \\ 
\hline
\end{tabular}
\end{table}

\section{Conclusions and future directions} \label{sec:conclusions}

This paper systematically explores the use of machine learning methods for subseasonal forecasting, highlighting several important factors: (1) the importance of using ensembles of physics-based forecasts (as opposed to only using the mean, as in common practice); (2) the potential for forecasting temperature and precipitation extremes using quantile regression; (3) the efficacy of different mechanisms, such as positional encoding and convolutional neural networks, for modeling spatial dependencies; (4) the importance of various features, such as sea surface temperature and lagged temperature and precipitation values, for predictive accuracy; (5) model stacking provides substantial benefits by leveraging the different utilization of spatial data among contributing models. The stacking model probably capitalizes on this diversity, fostering performance enhancement. Together, these results provide new insights into using ML for subseasonal weather forecasting in terms of the selection of features, models, and methods. 

Our results also suggest several important directions for future research. In terms of \textbf{features}, there are many climate forecasting ensembles computed by organizations such as NOAA and ECMWF. This paper focuses on ensembles in which ensemble members have a distinct ordering (in terms of lagged initial conditions used to generate them), but other ensembles correspond to initial conditions or parameters drawn independently from some distribution. Leveraging such ensemble forecasts and potentially jointly leveraging ensemble members from multiple distinct ensembles may further improve the predictive accuracy of our methods.  

In terms of \textbf{models}, new neural architecture models such as transformers \new{show} remarkable performance on \new{several} image analysis tasks \citep{dosovitskiy2020image, carion2020end, chen2021pre, khan2022transformers} and have potential in the context of forecasting climate temperature and precipitation maps. A careful study is needed, as past image analysis work using transformers generally uses large quantities of training data, exceeding what is available in SSF contexts. \new{Recent advancements in data-driven global weather forecasting models, such as Pangu-Weather \citep{bi2022pangu}, FourCastNet \citep{pathak2022fourcastnet}, and GenCast \citep{price2023gencast}, demonstrate the potential of ML techniques to enhance forecasting capabilities across various timescales. These models outperform traditional numerical weather prediction approaches, suggesting that similar data-driven methods may hold promise for improving SSF quality.}

In terms of \textbf{methods}, two outstanding challenges are particularly salient to the SSF community. The first is uncertainty quantification; that is, we wish not only to forecast temperature or precipitation but also to predict the likelihood of certain extreme events. Our work on quantile regression is an important step in this direction and statistical methods like conformalized quantile regression \citep{Romano2019ConformalizedQR}
may provide additional insights. Second, we see in \Cref{fig:tx_fl_preds}
that, at least in some geographic regions, the distribution of ensemble hindcast and forecast data may be quite different. Employing methods that are more robust to \textit{distribution drift} \citep{wiles2021fine, subbaswamy2021evaluating, zhu2021shift} is particularly important not only for handling forecast and hindcast data but also for accurate SSF in a changing climate.

\section*{Acknowledgments}
The authors gratefully acknowledge the support of the NSF (OAC-1934637, DMS-1930049, and DMS-2023109) and C3.ai.

%% file: appendix_part_final.tex

\newpage
\appendix 
\newpage
\section{Regression results for for NASA-GMAO Dataset}\label{sec:app_nasa_res}

\subsection{Regression}\label{sec:app_nasa_regr}

\paragraph{Precipitation regression using NASA-GMAO}\label{sec:precip_regr_nasa}

Precipitation regression results on the test data from NASA-GMAO are presented in Table \ref{table:precip_regression_nasa_test}. On this dataset, no learned method or method leveraging ensemble model forecasts significantly outperforms climatology. Note that the best $R^2$ value associated with the climatology is still negative; the low values for all methods indicate the difficulty of the forecasting problem.

\begin{table}[ht!]
\caption{\textbf{Test results for precipitation regression using NASA-GMAO dataset.} LR refers
to linear regression on all features including ensemble members, lagged data, land variables, and SSTs. Model stacking is performed on models that are learned on all features. Bold values indicate the best performance for each statistic. MSE is reported in squared mm.
}
\label{table:precip_regression_nasa_test}
\centering
\small
\begin{tabular}{c c  c  c  c  c  c  c  c } 
\hline
Model & Features & \makecell{Mean \\ $R^2$ ($\uparrow$)} & \makecell{Median\\ $R^2$ ($\uparrow$)}& \makecell{Mean\\ Sq Err ($\downarrow $)} & \makecell{Median \\ MSE ($\downarrow$)} & 
\makecell{90th prctl \\MSE ($\downarrow$)}\\
 \hline 
& Climatology & \bf -0.07 & \bf -0.02 & 2.14 $\pm$ 0.04 & \bf 1.51 & 
4.40\\ 
 & Ens mean & -0.11 & -0.06 & 2.13 $\pm$ 0.04 & 1.52 & 
4.31 \\ 
\multirow{-3}{*}{Baseline}
& Linear Regr & -0.18 & -0.14 & 2.25 $\pm$ 0.04 & 1.62 & 
4.68 \\
 \hline 
 
 \multirow{1}{*}{LR} 
& All features & -0.40 & -0.29 & 2.62 $\pm$ 0.05 & 1.93 & 
5.42 \\
\multirow{1}{*}{U-Net} 
& All features & -0.19 & -0.09 & 2.11 $\pm$ 0.03 & 1.56 & 
\bf 4.25 \\
 \multirow{1}{*}{RF}
& All features &  -0.18 & -0.11 & 2.17 $\pm$ 0.04 & 1.55 & 
4.44 \\ 
\multirow{1}{*}{Stacked} & LR, U-Net, RF, outputs &  -0.08 & -0.06 & \bf 2.09 $\pm$ 0.04 & 1.52 & 
4.27 \\ 
\hline
\end{tabular}
\end{table}

\begin{figure}[ht]
\centerline{\includegraphics[width=25pc]{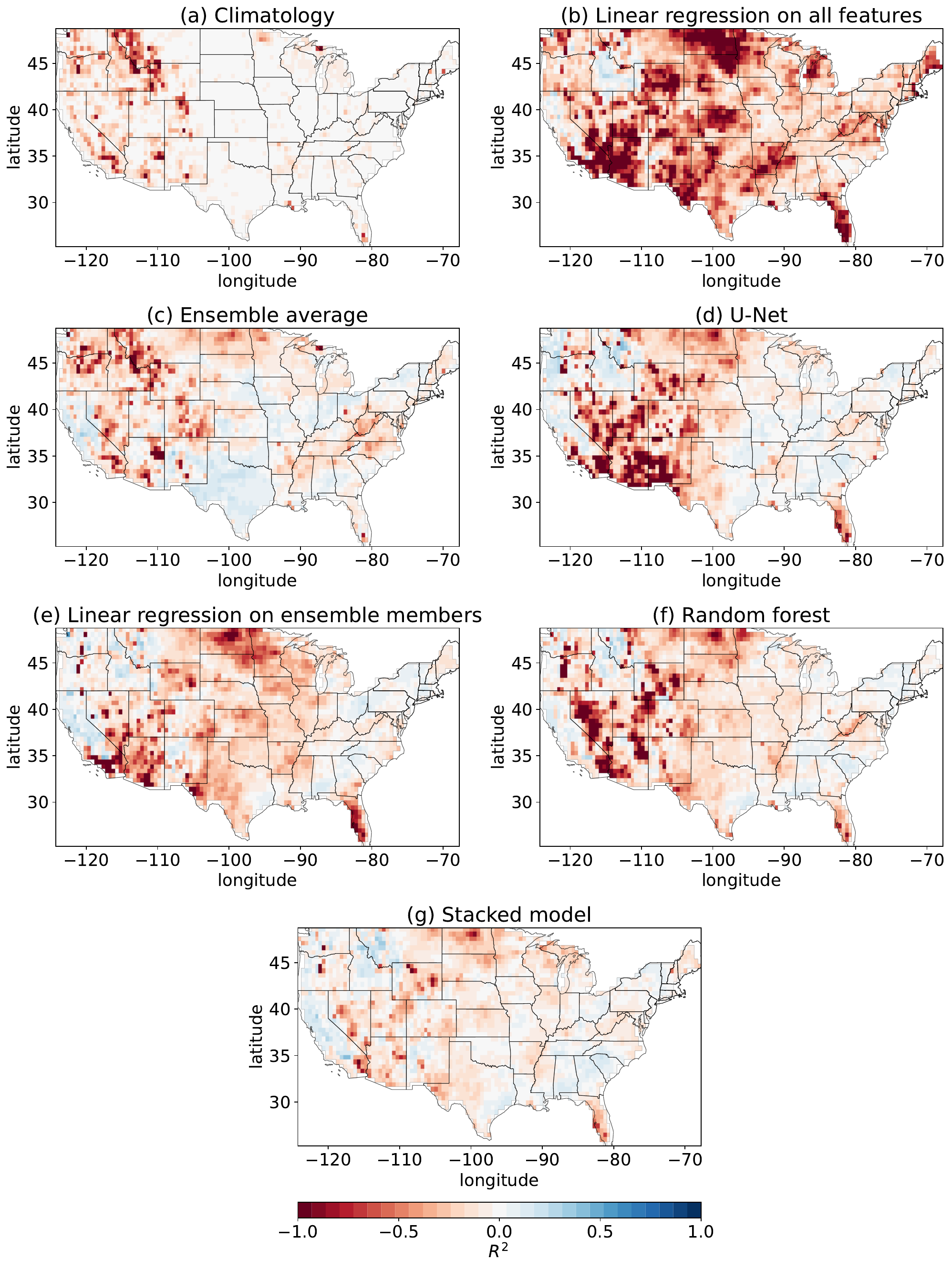}}
  \caption{\textbf{Test $R^2$ score heatmaps of baselines and learning-based methods for precipitation regression using the NASA-GMAO dataset}. Positive values (blue) indicate better performance. See Appendix \ref{sec:precip_regr_nasa} for details. 
  }\label{fig:precip_reg_nasa_test}
\end{figure}

Figure \ref{fig:precip_reg_nasa_test} illustrates the test performance of key methods on NASA-GMAO data with $R^2$ heatmaps over the U.S. Although the stacked model does not show the best performance in terms of mean $R^2$ score, it has more geographic regions with positive $R^2$ than any other method.

\paragraph{Regression of temperature}\label{sec:tmp_regr_nasa}
Temperature regression results using NASA-GMAO ensemble members are presented in Table \ref{table:tmp_regression_nasa_test}. The random forest and linear regression outperform all baselines in terms of both $R^2$ score and MSE. However, the U-Net model's performance is lower compared to other learned methods, which might be a sign of overfitting. Despite this performance drop of U-Net, the model stacking approach still demonstrates the best predictive skill. Note that the model stacking approach is applied to the models that are trained on all available features except SSTs  (similar to NCEP-CFSv2 data). 

\begin{table}[ht!]
\caption{\textbf{Test results for temperature regression using NASA-GMAO dataset.} LR refers
to linear regression on all features including ensemble members, lagged data, and land variables. Model stacking is performed on models that are learned on all features except SSTs. \newold{Bold values indicate the best performance for each statistic.} MSE is reported in squared $ ^\circ C $.
}
\label{table:tmp_regression_nasa_test}
\centering
\small
\begin{tabular}{c c  c  c  c  c  c  c  c } 
\hline 
Model & Features & \makecell{Mean \\ $R^2$ ($\uparrow$)} & \makecell{Median\\ $R^2$ ($\uparrow$)}& \makecell{Mean\\ Sq Err ($\downarrow $)} & \makecell{Median \\ MSE ($\downarrow$)} & 
\makecell{90th prctl \\ MSE ($\downarrow$)}\\  
 \hline 

& Climatology & -0.70 & -0.20 & 6.49 $\pm$ 0.11 & 5.06 & 
9.72\\ 
& Ens mean & -0.28 & 0.12 & 4.82 $\pm$ 0.10 & 3.43 &
7.82\\ 
\multirow{-3}{*}{Baseline} 
& Linear Regr & 0.12 & 0.14 & 3.32 $\pm$ 0.02 & 3.11 & 
4.70\\
 \hline 
\multirow{1}{*}{LR} 
& All features wo SSTs  & 0.17& 0.17 & 3.10$\pm$ 0.02 & 3.05 & 
4.26\\
\multirow{1}{*}{U-Net} 
& All features wo SSTs  & 0.06 & 0.12 & 3.40 $\pm$ 0.02 & 3.27 & 
4.52 \\
\multirow{1}{*}{RF}
& All features wo SSTs  & 0.20 & 0.22 & 3.03 $\pm$ 0.02 & 2.94 & 
4.25 \\ 
\multirow{1}{*}{Stacked} & LR, U-Net, RF, outputs &    \bf 0.21 & \bf 0.22 & \bf 2.94 $\pm$ 0.02& \bf 2.89 & 
\bf 3.97 \\
\hline
\end{tabular}
\end{table}

\begin{figure}[ht!]
 \centerline{\includegraphics[width=25pc]{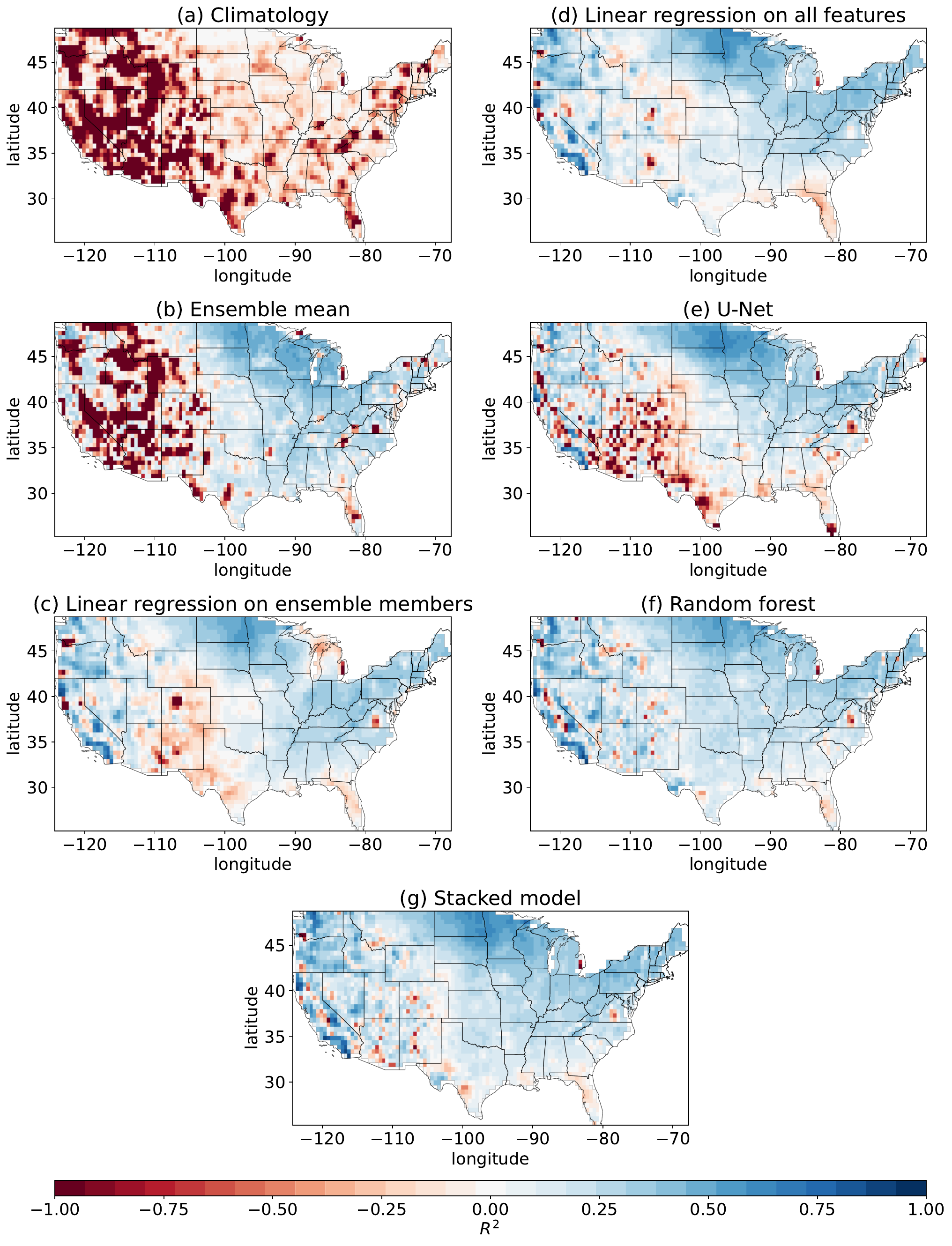}}
  \caption{\textbf{Test $R^2$ score heatmaps of baselines and learning-based methods for temperature regression using NASA-GMAO dataset}. Positive values (blue) indicate better performance. See Appendix \ref{sec:tmp_regr_nasa} for details.
  }\label{fig:tmp_reg_nasa_test}
\end{figure}

Figure \ref{fig:tmp_reg_nasa_test}  illustrates the test performance of key methods on NASA-GMAO data with $R^2$ heatmaps over the U.S. The stacked model shows the best performance across spatial locations. Similar to the NCEP-CFSv2 dataset, we notice that there are still regions where all models tend to exhibit a negative $R^2$ score.

\subsection{Quantile regression}\label{app:sec_q_regr}

\paragraph{Quantile regression of precipitation using NCEP-CFSv2 ensemble}\label{app:par_q_regr_precip_ncep}

\begin{figure}[ht]
\centerline{\includegraphics[width=25pc]{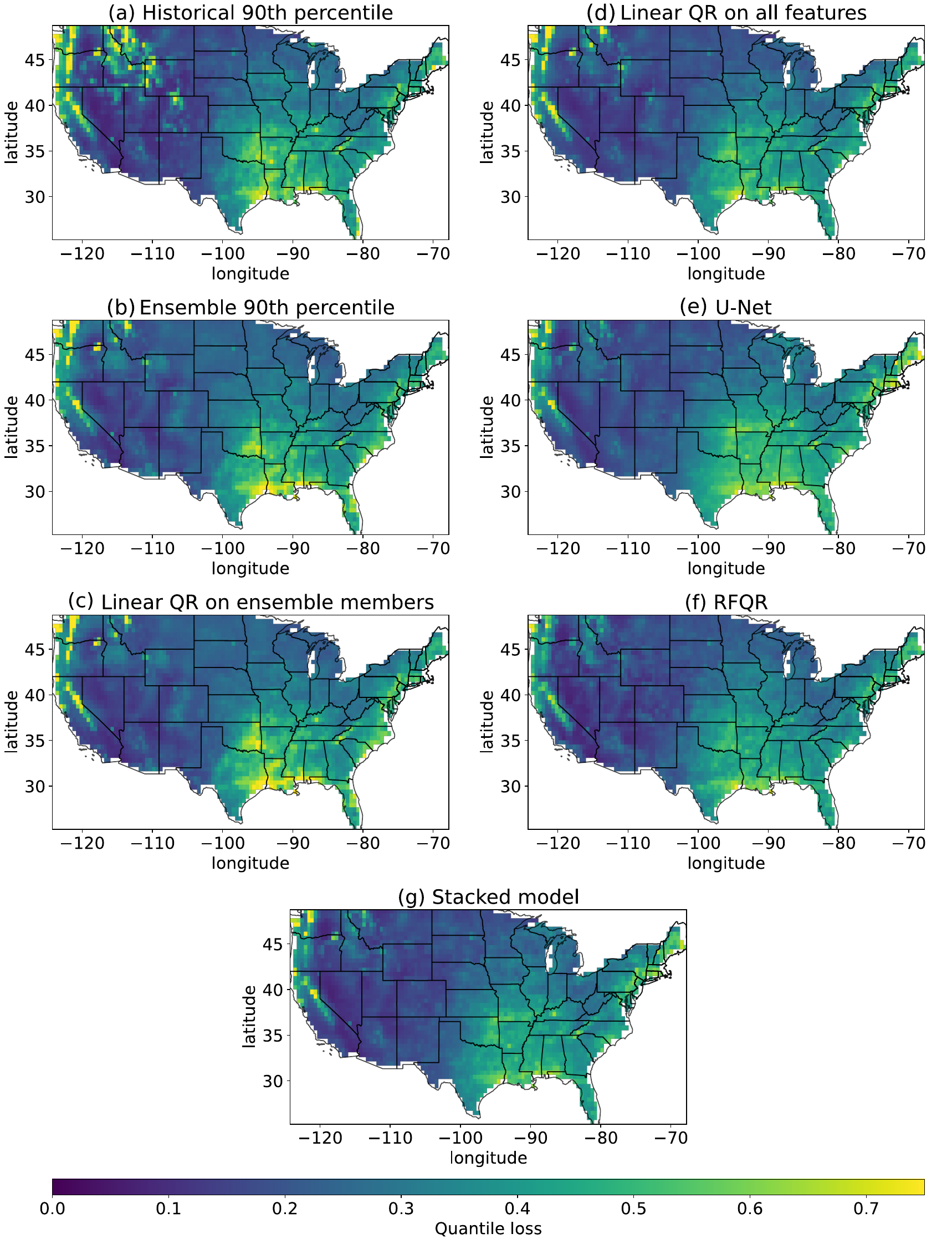}}
  \caption{\textbf{Test quantile loss heatmaps of baselines and learning-based methods for precipitation quantile regression using NCEP-CFSv2 dataset}. Blue regions indicate smaller quantile loss. See Section \ref{sec:qregr_ncep_precip} for details.
  }\label{fig:precip_qtr_loss_ncep_test}
\end{figure}

In \Cref{fig:precip_qtr_loss_ncep_test}, we show heatmaps of quantile loss using all locations in the U.S., where blue means smaller quantile loss and yellow means larger quantile loss. 
We observe that the learning-based models outperform the baselines, especially in Washington, California, Idaho, and near the Gulf of Mexico.

\paragraph{Quantile regression of precipitation using NASA-GMAO ensemble} \label{app:par_q_regr_precip}
Table \ref{table:precip_qtr_nasa_test} summarizes results for precipitation quantile regression using the NASA-GMAO ensemble. 
The models are the same as the models applied to the NCEP-CFSv2 ensemble.
Our best model shows similar performance to that of the baselines. One possible reason is that the NASA-GMAO ensemble shows worse performance than the NCEP-CFSv2 ensemble empirically. Furthermore, according to the designs of ensemble members from both climate models, the NASA-GMAO ensemble has fewer ensemble members and, therefore, has less coverage on the distribution of precipitation than the NCEP-CFSv2 ensemble, so our learned model has access to less information about the true distribution of precipitation.

\begin{table}[ht!]
\caption{\textbf{Test results for precipitation quantile regression using NASA-GMAO dataset, with target quantile = 0.9}. 
Model stacking is performed on models that are learned on all features. The \textbf{best} results are in bold. Quantile loss is reported in mm.}
\label{table:precip_qtr_nasa_test}
\centering
\small
\begin{tabular}{c c  c  c  c  c  c  c  c } 
\hline
Model & Features & \makecell{Mean \\ Qtr Loss ($\downarrow$)}    & \makecell{Median\\ Qtr Loss ($\downarrow$)}& \makecell{90th prctl \\Qtr Loss ($\downarrow$)} \\ %
 \hline 
& Historical 90-th percentile & 0.295 $\pm$ 0.003& 0.263  & 0.484\\ 

& Ens 90-th percentile & 0.378$\pm$ 0.005& 0.308  & 0.673\\ 
\multirow{-3}{*}{Baseline}
& Linear QR ens only & 0.336 $\pm$ 0.004& 0.286  & 0.531\\

\hline 
 
\multirow{1}{*}{Linear QR} 
& All features &  \bf 0.290 $\pm$ 0.003& \bf 0.253  &\bf 0.456 \\
 \multirow{1}{*}{U-Net} 
& All features &  0.310 $\pm$ 0.002& 0.278 &0.489 \\
 \multirow{1}{*}{RFQR}
& All features &  0.290 $\pm$ 0.002&  0.265  & 0.471 \\ 
 \multirow{1}{*}{Stacked} & U-Net, RFQR, LQR outputs &    0.296$\pm$ 0.002& 0.268 & 0.467 \\ 
\hline
\end{tabular}
\end{table}

\begin{figure}[ht]
  \centerline{\includegraphics[width=25pc]{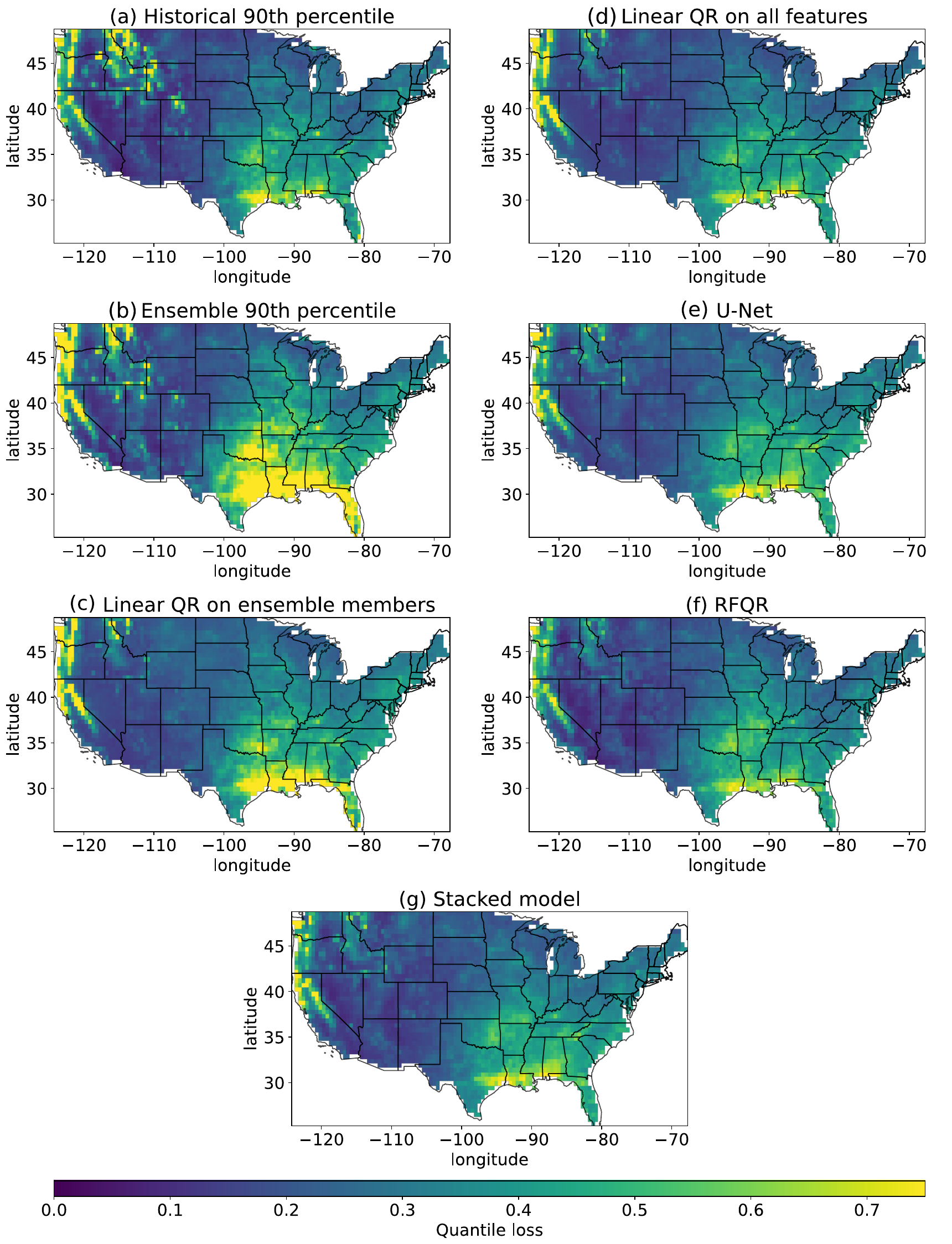}}
  \caption{\textbf{Test quantile loss heatmaps of baselines and learning-based methods for precipitation quantile regression using NASA-GMAO dataset}. Blue regions indicate smaller quantile loss. \newold{See Appendix \ref{app:par_q_regr_precip} for details.} 
  }\label{fig:precip_qtr_loss_nasa_test}
\end{figure}

\paragraph{Quantile regression of temperature}\label{app:par_q_regr_tmp}

\begin{figure}[ht]
 \centerline{\includegraphics[width=25pc]{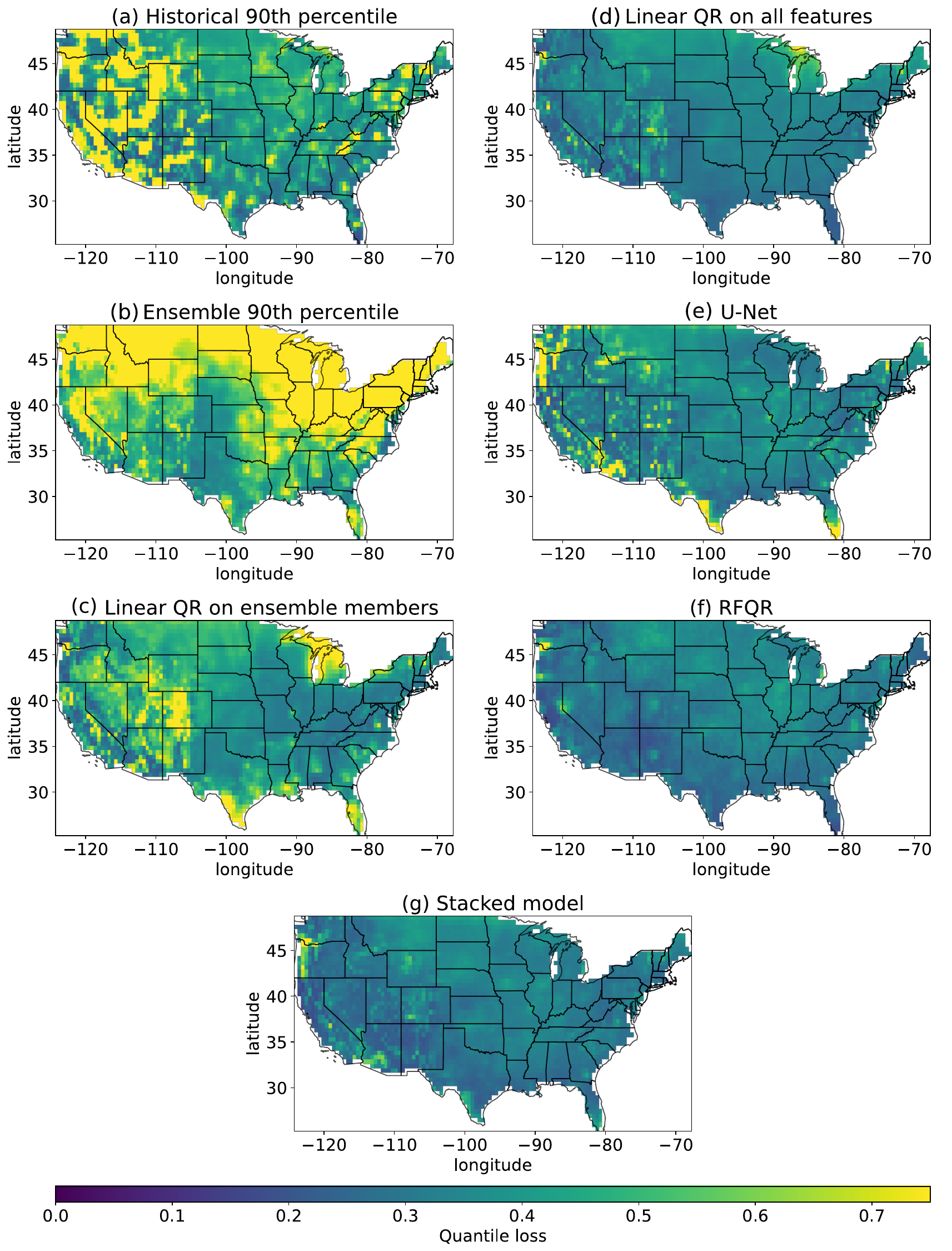}}
  \caption{\textbf{Test quantile loss heatmaps of baselines and learning-based methods for temperature quantile regression using NASA-GMAO dataset}. Blue regions indicate smaller quantile loss. \newold{See Appendix \ref{app:par_q_regr_tmp} for details.} }\label{fig:tmp_qtr_loss_nasa_test}
\end{figure}

Table \ref{table:tmp_qtr_nasa_test} and \Cref{fig:tmp_qtr_loss_nasa_test} summarize results for temperature quantile regression using the NASA-GMAO ensemble. All of our learned models are able to outperform all baselines.

\begin{table}[ht!]
\caption{\textbf{Test results for temperature quantile regression using NASA-GMAO dataset, with target quantile = 0.9}. Model stacking is performed on models that are learned on all features except for SSTs. The \textbf{best} results are in bold. Quantile loss is reported in $ ^\circ C $.}
\label{table:tmp_qtr_nasa_test}
\centering
\small
\begin{tabular}{c c  c  c  c  c  c  c  c } 
\hline
Model & Features & \makecell{Mean \\ Qtr Loss ($\downarrow$)} & \makecell{Median\\ Qtr Loss ($\downarrow$)} & \makecell{90th prctl \\Qtr Loss ($\downarrow$)} \\ %
 \hline
& Historical 90-th percentile & 0.596 $\pm$ 0.010& 0.438 & 0.988\\ 
 & Ens 90-th percentile & 0.812 $\pm$ 0.009& 0.646  & 1.493\\ 
\multirow{-3}{*}{Baseline}
& Linear QR ens only & 0.445 $\pm$ 0.003& 0.411  & 0.618\\
\hline
 
\multirow{1}{*}{Linear QR} 
& All features wo SSTs  &  0.341 $\pm$ 0.001& 0.333  &0.419 \\
 \multirow{1}{*}{U-Net} 
& All features wo SSTs &  0.375 $\pm$ 0.003& 0.347  &0.477 \\
 \multirow{1}{*}{RFQR}
& All features wo SSTs & 0.318 $\pm$ 0.002 & 0.316  & 0.376 \\ 
 \multirow{1}{*}{Stacked} & U-Net, RFQR, LQR outputs & \textbf{0.315} $\pm$ 0.002& \bf 0.310  & \bf 0.374 \\ 
\hline
\end{tabular}
\end{table}

\section{Tercile Classification}\label{appendix:tercile}

In this section, we present results for the tercile classification task for both climate variables and both datasets.

\subsection{Tercile classification of precipitation}\label{app:terc_precip}

In this case, the proposed learning-based methods are directly trained on the classification task. Predictions of baselines, such as the ensemble mean, are split into three classes according to the 33rd and 66th percentile values. Note that random forest and U-Net are trained for classification using all available features. We do not notice a significant difference in the performance of logistic regression on the validation if the inputs are ensemble members only or ensemble members with side information. So, we use logistic regression on ensemble members only. The model stacking is applied to the logistic regression, U-Net, and random forest outputs.

Table \ref{table:precip_tercile_test} summarizes results for NCEP-CFSv2 and NASA-GMAO datasets on the test data. For this task, the learning-based methods achieve the best performance in terms of accuracy for both datasets. In the case of NCEP-CFSv2 data, U-Net achieves the highest accuracy score, and the performance of the stacked model is comparable with it. For NASA-GMAO data, the stacked model shows the best performance.

\begin{table}[ht!]
\caption{\textbf{Test results for tercile classification of precipitation on different datasets.} Accuracy in \% is reported. Note that for this task, our models are trained for classification directly while baselines perform regression, and a threshold for predicted values is applied. For stacking, logistic regression, U-Net and RF outputs are used.}
\label{table:precip_tercile_test}
\centering
\small
\begin{tabular}{ c c  c  c  c  c  c  c  c} 
 \hline
 Data & Model & \makecell{Mean  \\ accuracy ($\uparrow$)} & \makecell{Median \\ accuracy ($\uparrow$)} \\  
 \hline 
 & Ens mean &  38.00 $\pm$0.16& 37.61  \\
  & Logistic Regr & 41.22 $\pm$0.14& 40.17  \\  
 \multirow{-3}{4em}{NCEP-CFSv2} &  U-Net &  \textbf{43.88}  $\pm$0.12& \textbf{42.74} \\ 
 &  RF & 42.38 $\pm$0.13& 41.88  \\ 
&  Stacked & 43.81 $\pm$0.13& \textbf{42.74}   \\ 
\hline
 & Ens mean &  38.64 $\pm$0.14& 37.65 \\  
 &   Logistic regr & 41.51 $\pm$0.16& 40.00  \\
\multirow{-3}{4em}{NASA-GMAO} &  U-Net & 40.53 $\pm$0.11& 40.00  \\ 
 & RF &   40.79 $\pm$0.14& 40.00  \\ 
 &  Stacked & \textbf{42.08} $\pm$0.14& \textbf{41.18}  \\ 
\hline 
\end{tabular}
\end{table}

\begin{figure}[ht!]
\centerline{\includegraphics[width=27pc]{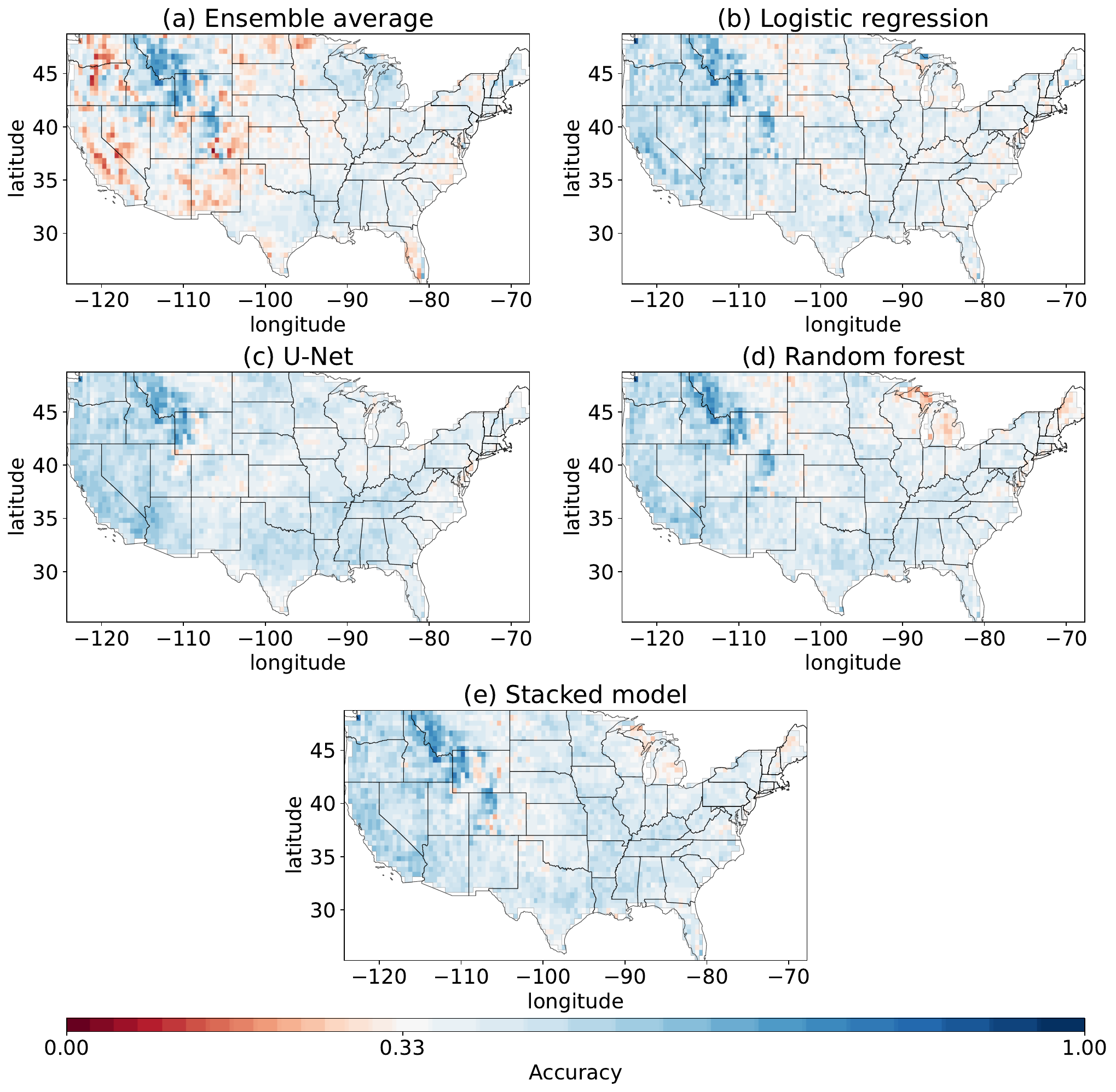}}
  \caption{\textbf{Test accuracy heatmaps of baselines and learning-based methods for tercile classification of precipitation using NCEP-CFSv2 dataset.} The accuracy colorbar is recentered to be white at $\frac{1}{3}$, what corresponds to a random guess score. Blue pixels indicate better performance, while red pixels correspond to performance that is worse than a random guess. \newold{See Appendix \ref{app:terc_precip} for details.} 
  }\label{fig:tercile_precip_ncep}
\end{figure}

The accuracy heatmaps over U. S. land are presented in the Figure \ref{fig:tercile_precip_ncep} for NCEP-CFSv2 dataset. The plots corresponding to the learning-based methods show the best results, especially at the West Coast, Colorado and North America. 

\begin{figure}[ht!]
\centerline{\includegraphics[width=27pc]{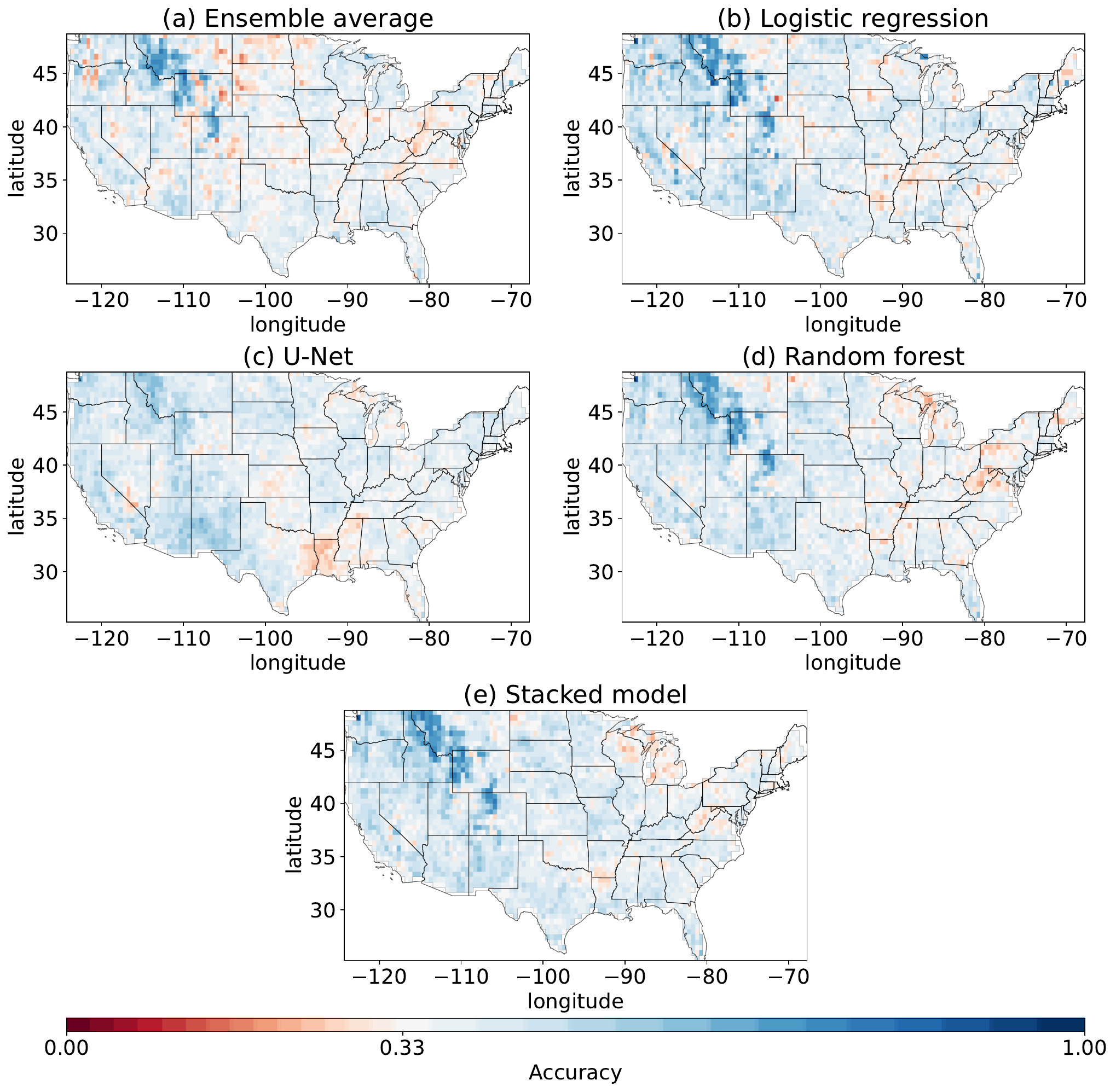}}
  \caption{\textbf{Test accuracy heatmaps of baselines and learning-based methods for tercile classification of precipitation using NASA-GMAO dataset.} The accuracy colorbar is recentered to be white at $\frac{1}{3}$, what corresponds to a random guess score. Blue pixels indicate better performance, while red pixels correspond to performance that is worse than a random guess. See Appendix \ref{app:terc_precip} for details. 
  }\label{fig:tercile_precip_nasa}
\end{figure}

The accuracy heatmaps over the U. S. land are presented in Figure \ref{fig:tercile_precip_nasa} for the NASA-GMAO dataset. The plots corresponding to the learning-based methods show the best results, the ensemble mean's figure has the most red regions.

\subsection{Tercile classification of temperature}\label{app:terc_temp}

The next task is tercile classification of 2-meter temperature. In this case, the threshold is applied to the regression predictions of all methods, meaning there is no direct training for a classification. Table \ref{table:tmp_tercile_test} summarizes results for NCEP-CFSv2 and NASA-GMAO datasets on the test data. For this task, the learning-based methods achieve the best performance in terms of accuracy, stacked model using NCEP-CFSv2 data and linear regression using NASA-GMAO data and all additional features (except SSTs). In general, all learning-based models significantly outperform the ensemble mean.

\begin{table}[ht!]
\caption{\textbf{Test results for tercile classification of temperature on different datasets.} Accuracy in \% is reported. Note that for this task, our models are trained for regression and the threshold for predicted values is applied.}
\label{table:tmp_tercile_test}
\centering
\small
\begin{tabular}{ c c  c  c  c  c  c  c  c} 
 \hline
 Data & Model & \makecell{Mean  \\ accuracy ($\uparrow$)} & \makecell{Median \\ accuracy ($\uparrow$)}   \\  
 \hline 
 & Ens mean &  44.84 $\pm$0.36& 42.74  \\  
 &  Linear Regr &  57.10 $\pm$0.25& 54.69  \\
\multirow{-3}{4em}{NCEP-CFSv2} & LR &  57.34 $\pm$0.25& \bf 54.71\\
 & U-Net &  53.80 $\pm$0.28& 50.43 \\ 
 & RF &   58.07 $\pm$0.27& 54.70 \\ 
 & Stacked &  \textbf{58.12} $\pm$0.20& \bf 54.71  \\
\hline

& Ens mean & 52.23 $\pm$0.25& 49.41\\  
 &  Linear Regr & 57.75 $\pm$0.25&  54.11\\
\multirow{-3}{4em}{NASA-GMAO}  &  LR & \textbf{58.97} $\pm$0.25& \bf 55.29 \\ 
 & U-Net &  55.64 $\pm$0.27& 51.76\\ 
&  RF &  58.78 $\pm$0.26& \bf 55.29 \\ 
 & Stacked & 58.72 $\pm$0.25& 54.12 \\ 
\hline 
\end{tabular}
\end{table}

\begin{figure}[ht!]
\centerline{\includegraphics[width=27pc]{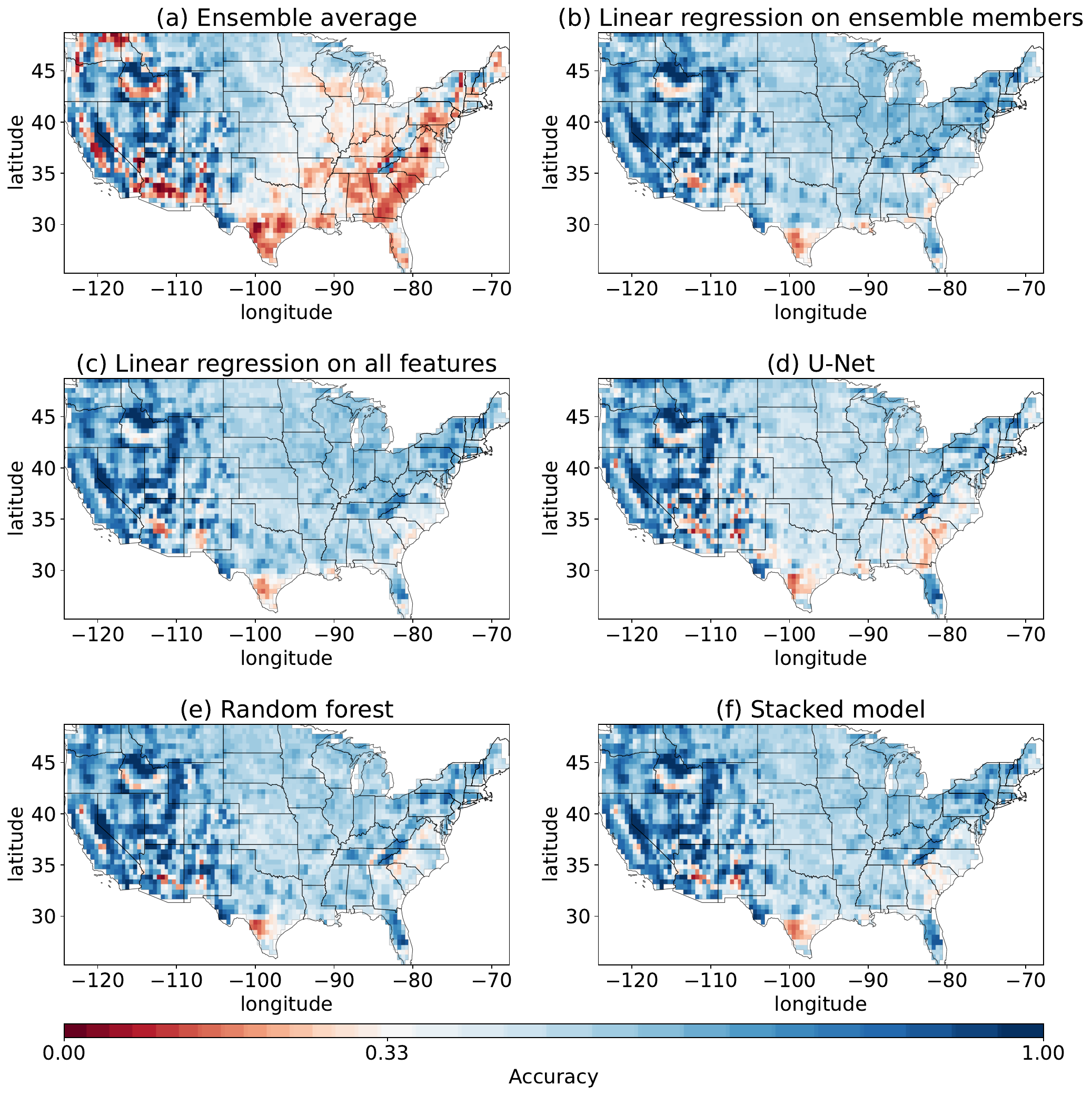}}
  \caption{\textbf{Test accuracy heatmaps of baselines and learning-based methods for tercile classification of temperature using NCEP-CFSv2 dataset.} The accuracy colorbar is recentered to be white at $\frac{1}{3}$, what corresponds to a random guess score. Blue pixels indicate better performance, while red pixels correspond to performance that is worse than a random guess.  \newold{See Appendix \ref{app:terc_temp} for details.}  
  }\label{fig:tercile_tmp_ncep}
\end{figure}

Figure \ref{fig:tercile_tmp_ncep} shows accuracy heatmaps over the U.S. for different methods using NCEP-CFSv2 data. The stacked model shows the best performance across spatial locations. For example, the ensemble mean does not show great performance in the Southeast and Middle Atlantic regions, while learning-based methods demonstrate much stronger predictive skills in these areas. However, there are still some areas, such as Texas or South West region, with red pixels for all methods.

\begin{figure}[ht!]
\centerline{\includegraphics[width=27pc]{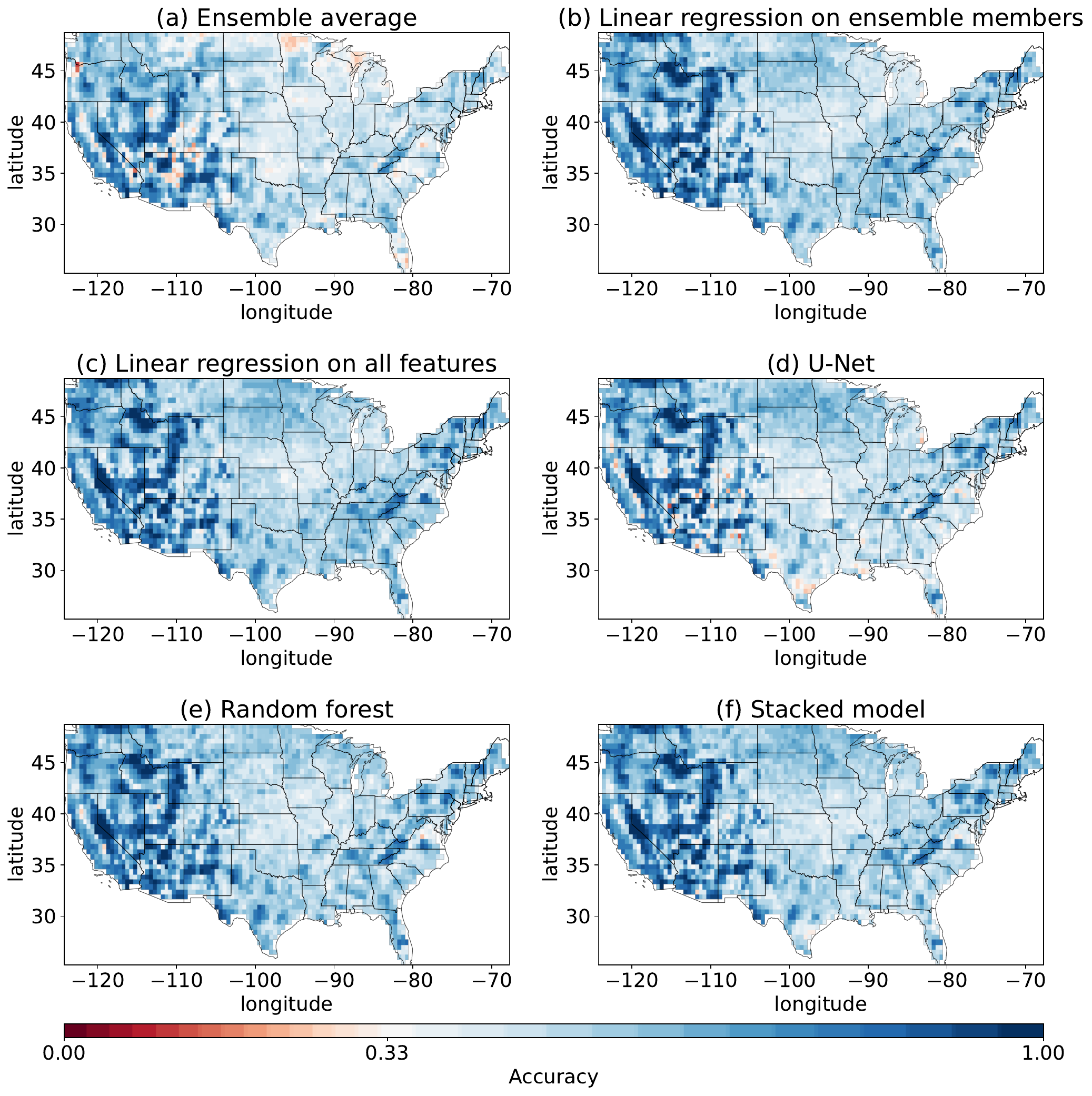}}
  \caption{\textbf{Test accuracy heatmaps of baselines and learning-based methods for tercile classification of temperature using NASA-GMAO dataset.} The accuracy colorbar is recentered to be white at $\frac{1}{3}$, what corresponds to a random guess score. Blue pixels indicate better performance, while red pixels correspond to performance that is worse than a random guess. \newold{See Appendix \ref{app:terc_temp} for details.}
  }\label{fig:tercile_tmp_nasa}
\end{figure}

Figure \ref{fig:tercile_tmp_nasa} shows accuracy heatmaps over the U.S. for different methods using NASA-GMAO data. In this case, linear regression on all features achieves the best scores. Other learning-based methods outperform the ensemble mean too, especially in the West and in Minnesota. 

\section{Extended Discussions}\label{app:ext_disc}

In this section, we present more detailed results for Section \ref{section:discussion}. We also provide additional experiments on the temperature forecasting analysis, and experiments on training set sizes and bootstrap.

\subsection{PE v.s. latitude/longitude values or no location information}\label{app:lat_lon_test}

In this section, we elaborate on our experiment with different uses of location information. Similar to Section \ref{sec:disc_spatial_data}, we train and test our models with three different settings: using no location information, using latitude/longitude values, or using positional encodings. For the stacked model, we first train the LR, RF, and U-Net using these different settings before training the stacked model using the corresponding LR, RF, and U-Net outputs. Table \ref{table:precip_lat_lon_test_all} summarizes test performance for precipitation regression with these three settings of using location information. For linear regression, we observe that having latitude/longitude values or adding PE features does not improve its performance. One interesting result is that adding PE for the LR degraded its performance, which may be due to the fact that the PE features are non-linear transformations of latitude and longitude values, which is hard to fit with a linear model. 
\par
\begin{figure}[!hbt]

\centerline{\includegraphics[width=27pc]{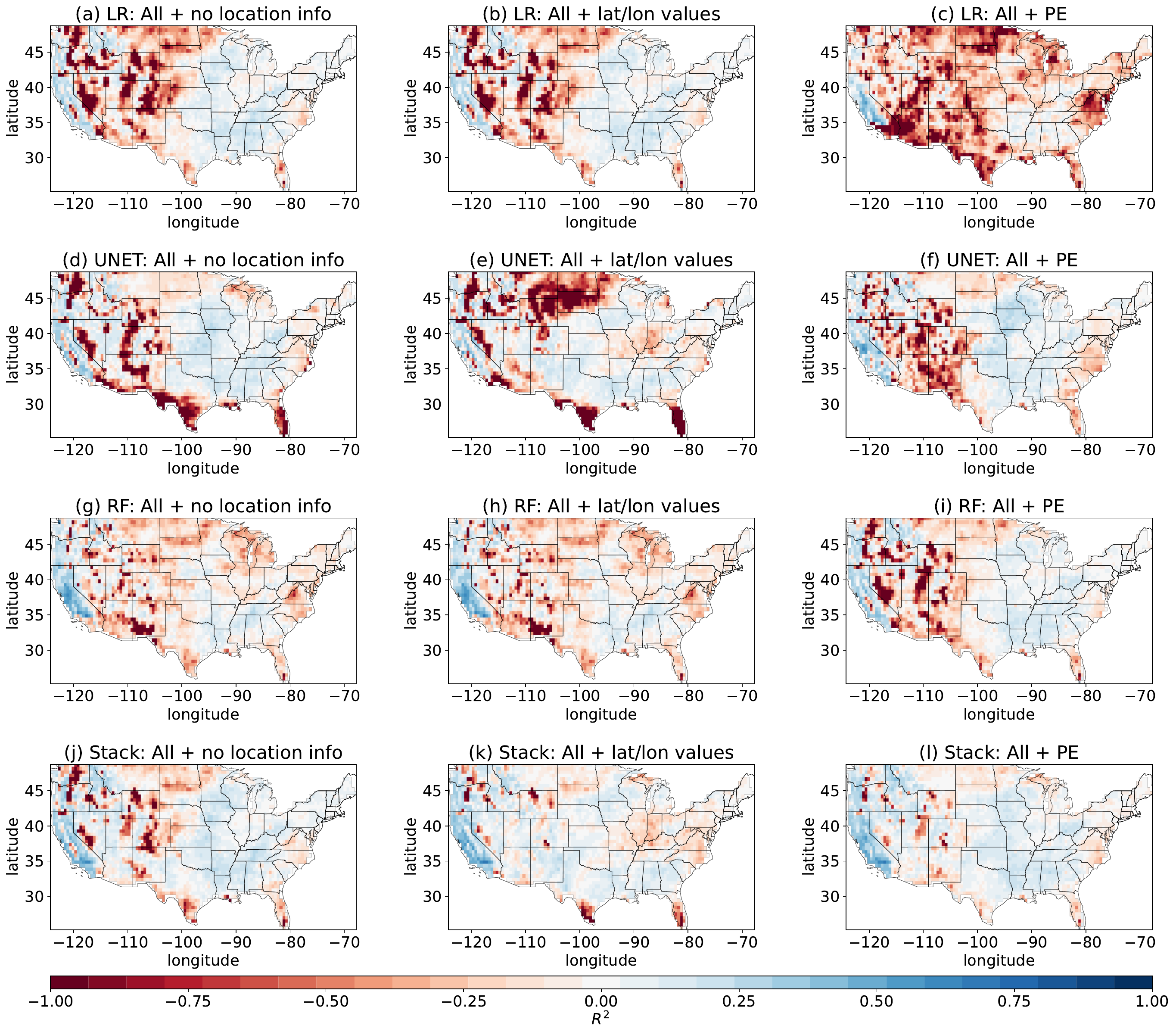}}
  \caption{\newold{\textbf{Precipitation regression test $R^2$ heatmaps of LR, U-Net, RF and stacked model trained using no spatial features, using latitude and longitude values or using PE.} The NCEP-CFSv2 ensemble is used. See Appendix \ref{app:lat_lon_test} for more details.
}
  }\label{fig:lat_lon_test_precip_ncep}
\end{figure}

 For the RF, U-Net, and the stacked model, we observe that adding PE improves their performance with significance, i.e. having at least one standard error smaller MSE. For the U-Net, using latitude/longitude values yields worse overall performance compared to using no location information. 
Figure \ref{fig:lat_lon_test_precip_ncep} shows the test $R^2$ heatmaps for LR, U-Net, RF, and stacked model under these three settings for precipitation regression. We observe that adding PE features not only improves performance for the U-Net and RF, the stacked model's performance also improves from having better predictions from the U-Net and RF.

\begin{table}[!ht]
\caption{\textbf{Precipitation regression test performance comparison of LR, U-Net, RF and stacked model trained using no spatial features, using latitude and longitude values or using PE. } The \textbf{best} results are in bold.}
\label{table:precip_lat_lon_test_all}

\centering
\small
\begin{tabular}{c c  c  c  c  c  c  c  c} 
 \hline
Model & Features & \makecell{Mean \\ $R^2$ ($\uparrow$)} & \makecell{Mean \\ Sq Err ($\downarrow$)}\\
 \hline
& All + no location info &  \bf -0.11 & \textbf{2.29}$\pm$0.04 \\  
\multirow{-2}{*}{LR}
 & All + lat/lon values  & \bf -0.11 & \textbf{2.29}$\pm$0.04 \\ 
 & All + PE & -0.33 & 2.71$\pm$0.05\\ 
 
\hline
 & All + no location info &  -0.16 &2.31$\pm$0.04  \\  
\multirow{-2}{*}{U-Net}
 & All + lat/lon values  &  -0.28 & 2.53$\pm$0.05 \\ 
 & All + PE &  \bf -0.10 & \textbf{2.18}$\pm$ 0.03\\ 

 \hline
 & All + no location info &  -0.18 &2.23$\pm$0.04  \\  
\multirow{-2}{*}{RF}
 & All + lat/lon values  &  -0.16 & 2.21$\pm$0.04 \\ 
 & All + PE &  \bf -0.11 & \textbf{2.17}$\pm$ 0.05\\ 

 \hline
 & All + no location info &  -0.05 &2.13$\pm$0.03  \\  
\multirow{-2}{*}{Stacked}
 & All + lat/lon values  &  -0.01 & 2.21$\pm$0.04 \\ 
 & All + PE &  \bf 0.02 & \textbf{2.07}$\pm$ 0.03\\ 
\hline
\end{tabular}
\end{table}

\par
\begin{figure}[!hbt]

\centerline{\includegraphics[width=27pc]{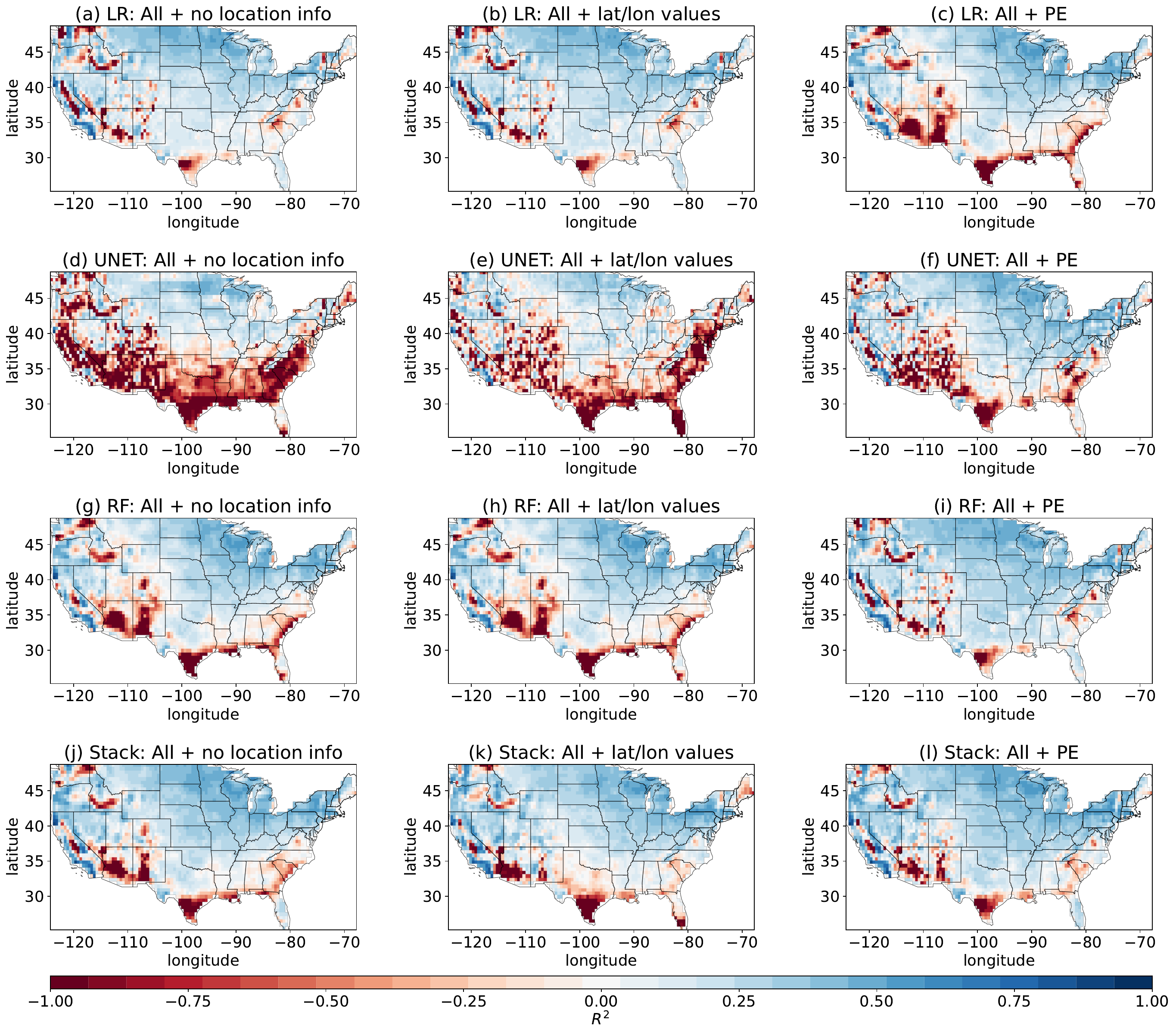}}
  \caption{\newold{\textbf{Temperature regression test $R^2$ heatmaps of LR, U-Net, RF and stacked model trained using no spatial features, using latitude and longitude values or using PE.} The NCEP-CFSv2 ensemble is used. See Appendix \ref{app:lat_lon_test} for more details.
}
  }\label{fig:lat_lon_test_tmp_ncep}
\end{figure}

 Table \ref{table:tmp_lat_lon_test_all} shows the test performances of LR, U-Net, RF and stacked model on temperature regression. Similar to precipitation regression, adding latitude/longitude values or adding PE does not help the LR, but we observe significant performance improvement when adding positional encoding features for the U-Net, RF, and the stacked model. Figure \ref{fig:lat_lon_test_tmp_ncep} then shows the test $R^2$ heatmaps for temperature regression. We can see that adding PE to the U-Net, RF, and stacked model improves forecast performance, especially in regions like Arizona, New Mexico, and Texas.
\begin{table}[!ht]
\caption{\textbf{Temperature regression test performance comparison of LR, U-Net, RF and stacked model trained using no spatial features, using latitude and longitude values or using PE. } The \textbf{best} results are in bold. }
\label{table:tmp_lat_lon_test_all}

\centering
\small
\begin{tabular}{c c  c  c  c  c  c  c  c} 
 \hline
Model & Features & \makecell{Mean \\ $R^2$ ($\uparrow$)} & \makecell{Mean \\ Sq Err ($\downarrow$)}\\
 \hline
& All + no location info &  \bf 0.05 & \textbf{3.57}$\pm$0.03 \\  
\multirow{-2}{*}{LR}
 & All + lat/lon values  & \bf 0.05 & \textbf{3.57}$\pm$0.03 \\ 
 & All + PE & \bf 0.05 & 3.57$\pm$0.03\\ 
 
\hline
 & All + no location info &  -0.35 &4.81$\pm$0.04  \\  
\multirow{-2}{*}{U-Net}
 & All + lat/lon values  &  -0.21 & 4.47$\pm$0.03 \\ 
 & All + PE &  \bf 0.01 & \textbf{3.65}$\pm$ 0.02\\ 

 \hline
 & All + no location info & 0.11 &3.37$\pm$0.02  \\  
\multirow{-2}{*}{RF}
 & All + lat/lon values  &  0.14 & 3.28$\pm$0.02 \\ 
 & All + PE &  \bf 0.16& \textbf{3.17}$\pm$ 0.02\\ 

 \hline
 & All + no location info &  0.12 &3.35$\pm$0.02  \\  
\multirow{-2}{*}{Stacked}
 & All + lat/lon values  &  0.12 & 3.33$\pm$0.02 \\ 
 & All + PE &  \bf 0.18 & \textbf{3.11}$\pm$ 0.02\\ 
\hline
\end{tabular}
\end{table}

\subsection{Bootstrap experiments}\label{sec:unet_bootstrap}
\par
\begin{figure}[!hbt]

\centerline{\includegraphics[width=19pc]{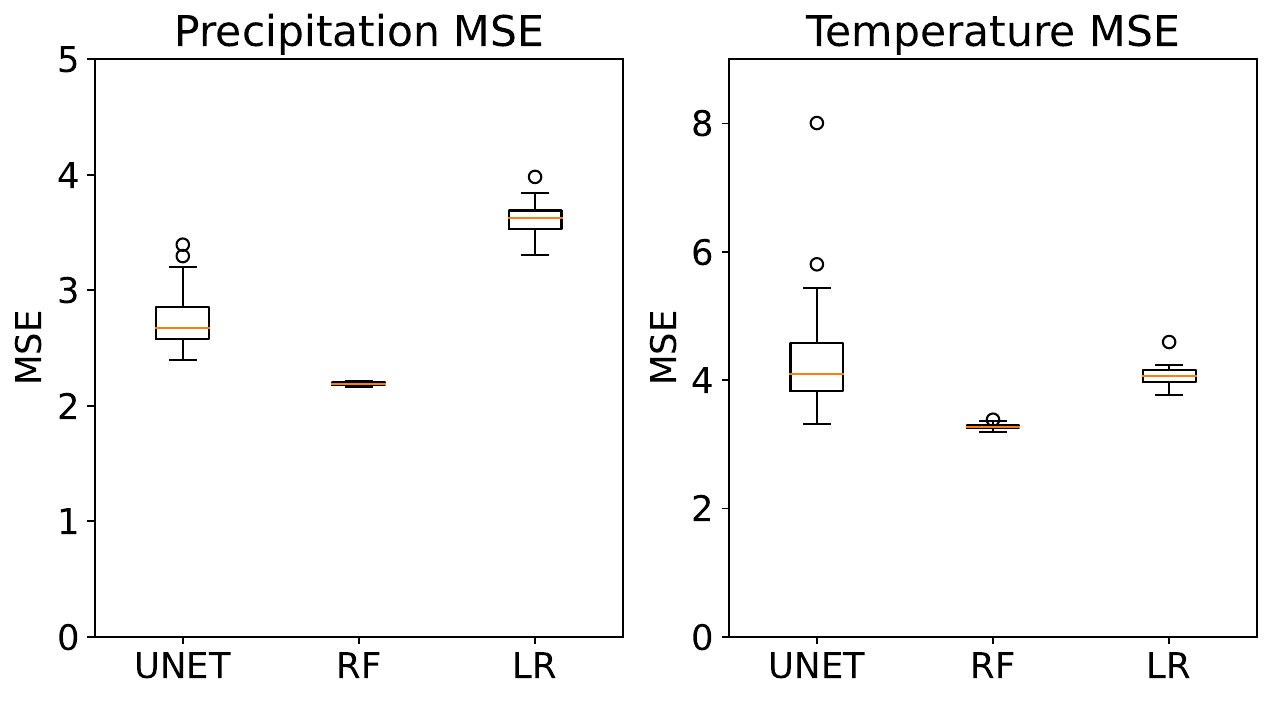}}
  \caption{\newold{\textbf{Box plots of MSEs for the U-Net, LR and RF trained on 50 sets of different bootstrap samples, each with size 200.} The NCEP-CFSv2 ensemble is used. See Appendix \ref{sec:unet_bootstrap} for more details.
}
  }\label{fig:bootstrap}
\end{figure}

 To evaluate the stability of our machine learning models with small sample sizes, we perform the following bootstrap experiments: We take bootstrap samples of size 200 from our training set and retrain our U-Net, RF, and LR. Then we evaluate these models on the test set. We repeat this process 50 times and show the results in \cref{fig:bootstrap}. We observe from the plots that the U-Net performs consistently better than the LR in precipitation regression but not for temperature regression. This result is consistent with what we showed in \cref{table:precip_regression_ncep_test} and \cref{table:tmp_regression_ncep_test}. 
\par We also observe that the U-Net is more sensitive to different bootstrap samples than the RF and LR, which is not surprising since for the U-Net, the bootstrap samples correspond to 200 different spatial maps for training. In contrast, for the RF and LR, the bootstrap samples correspond to $200 \times 3274$ training samples.

\subsection{Precipitation forecast example} \label{sec:ens_members_analysis_app}

\begin{figure}[ht!]
\centerline{\includegraphics[width=25pc]{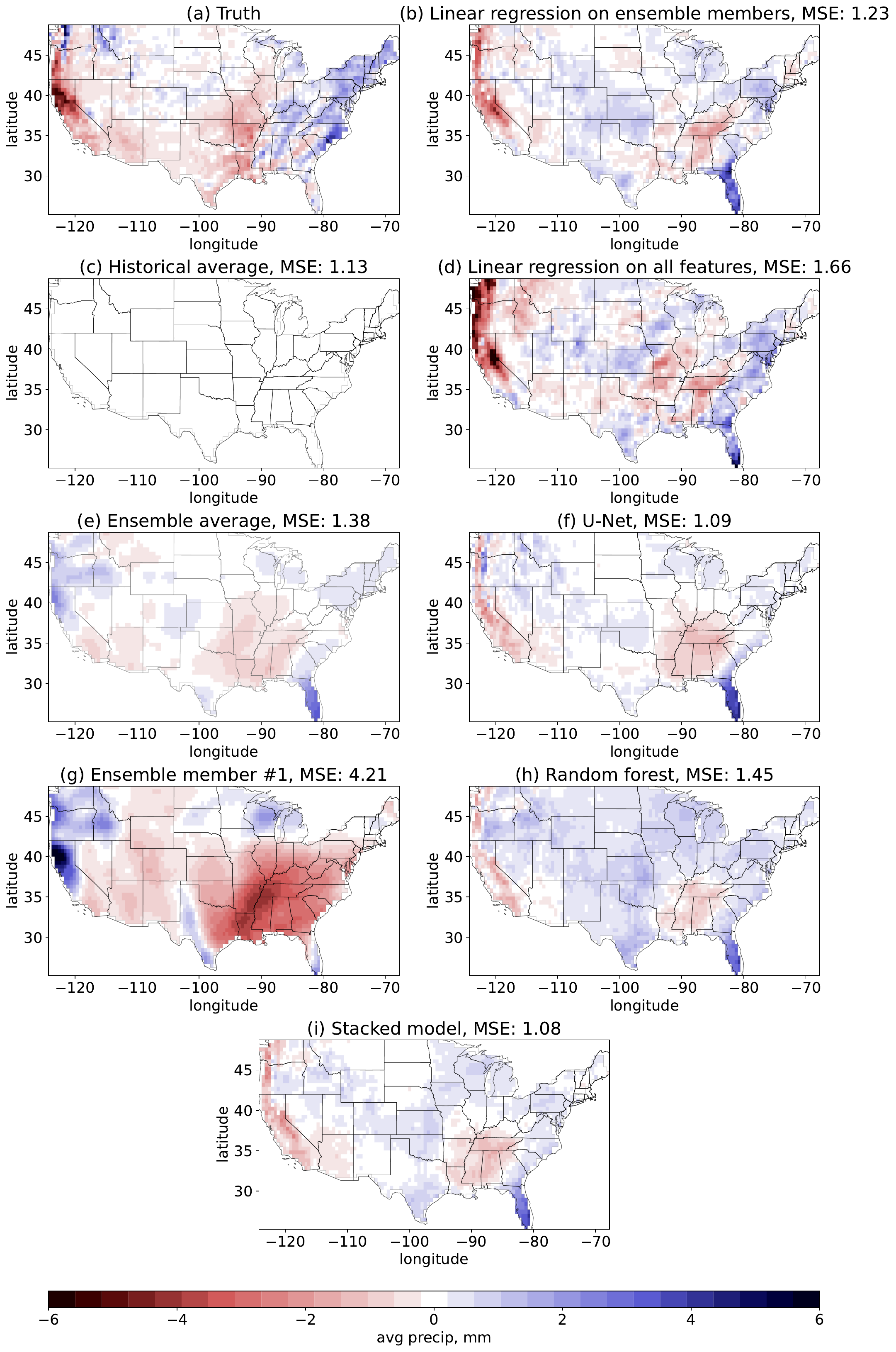}}
  \caption{\textbf{An illustration of precipitation predictions $\newold{\hat y_{t, l}^{\text{anomaly}}}$ ( \newold{in mm}) of different methods for February 2016 (in test period).}
  (a) True precipitation. 
  (b) LR on ensemble members.
  (c) Climatology.
  (d) LR on all features.
  (e) Ensemble mean.
  (f) U-Net on all features. 
  (g) Example single ensemble member. 
  (h) Random forest on all features. 
  (i) Stacked model. 
  See Appendix \ref{sec:ens_members_analysis_app} for details.
  }\label{fig:compare_predictions}
\end{figure}

While climate simulations and ensemble forecasts are designed to provide useful predictions of temperature and precipitation based on carefully developed physical models, we see that machine learning applied to those ensembles can yield a significantly higher predictive skill for a range of SSF tasks. \Cref{fig:compare_predictions} illustrates key differences between different predictive models for predicting monthly precipitation with a lead time of 14 days.
Individual ensemble members are predictions with high levels of spatial smoothness and more extreme values. Linear regression, the random forest, the U-Net, and the stacked model produce higher spatial frequencies. The linear regression result, which uses a different model trained for each spatial location separately, has the least spatial smoothness of all \newold{methods}; this is especially visible in the southeast and potentially does not reflect realistic spatial structure. 
The learning-based models more accurately predict localized regions of high and low precipitation compared to the ensemble mean.

\begin{figure}[ht]
\centerline{\includegraphics[width=25pc]{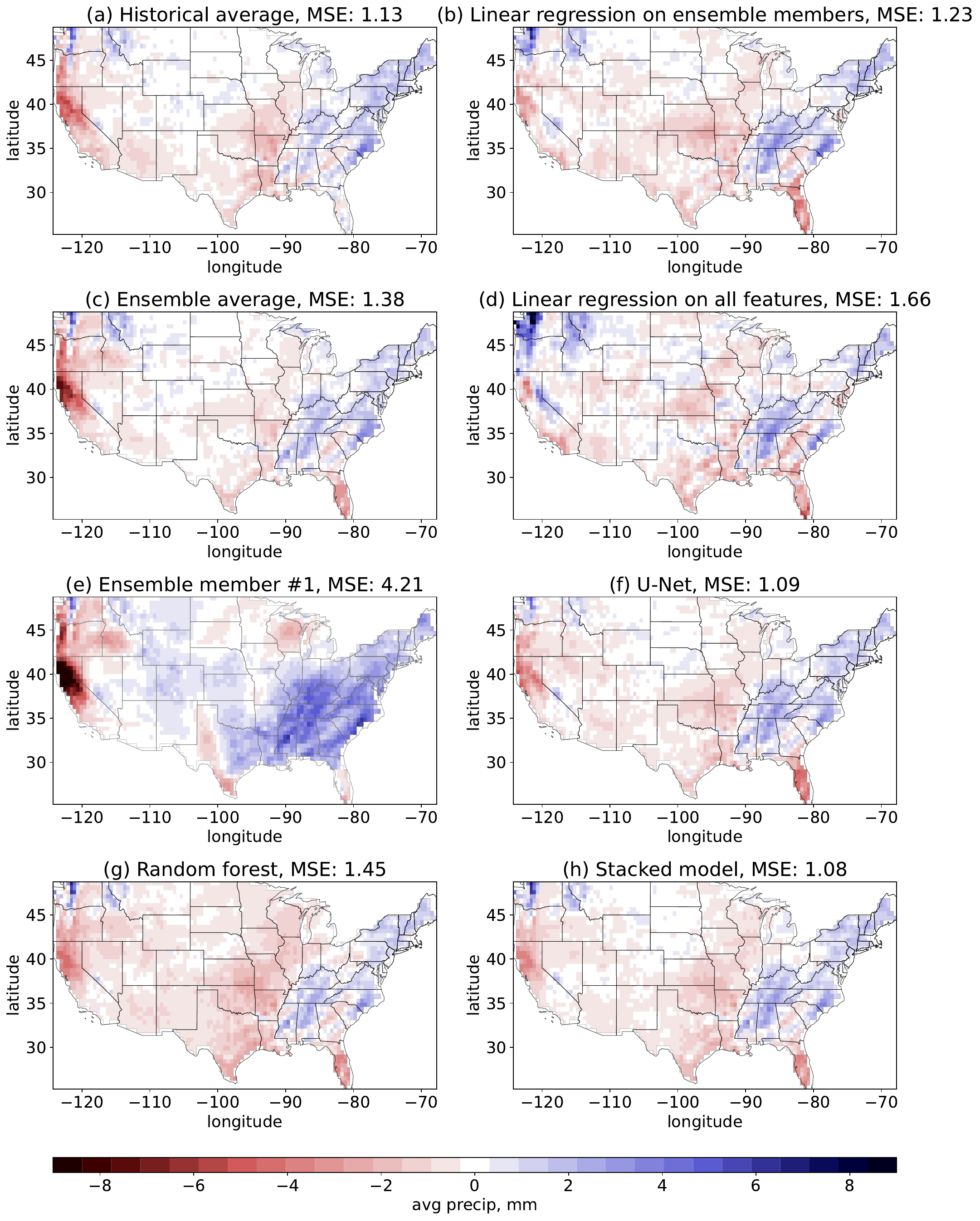}}
  \caption{\textbf{An illustration of differences $y_{t, l}^{\text{anomaly}} - \hat y_{t, l}^{\text{anomaly}}$ in precipitation predictions \newold{in mm} of different methods for February 2016 (in test period).} Red pixels indicate areas where a forecasting method predicts higher precipitation levels compared to the ground truth, blue pixels indicate an underestimation of the precipitation, and white pixels correspond to a precise prediction. See Appendix \ref{sec:ens_members_analysis_app} for details.
  }\label{fig:compare_predictions_diff}
\end{figure}

\Cref{fig:compare_predictions_diff} demonstrates differences between the ground truth and different model predictions. In this figure, the color white is associated with the smallest errors, while red pixels indicate overestimating precipitation and blue pixels indicate underestimating precipitation. 
The individual ensemble member in \Cref{fig:compare_predictions_diff}(e) exhibits dark red regions across the West, while the ensemble mean in \Cref{fig:compare_predictions_diff}(e) shows better performance in this area. The colors are more muted for the stacked model in \Cref{fig:compare_predictions_diff}(h). The climatology in \Cref{fig:compare_predictions_diff}(a) has the most neutral areas. However, its MSE is slightly higher than the stacked model's MSE. In general, all methods, including linear regression (b, d), U-Net (f), and random forest (g), tend to underpredict precipitation in the Southeast, Mid-Atlantic, and North Atlantic and predict higher precipitation levels in the West. 

\subsection{Temperature forecasting analysis}\label{sec:tmp_regions_analysis}

\begin{figure}[!ht]
     \centering
     
         \includegraphics[width=19pc]{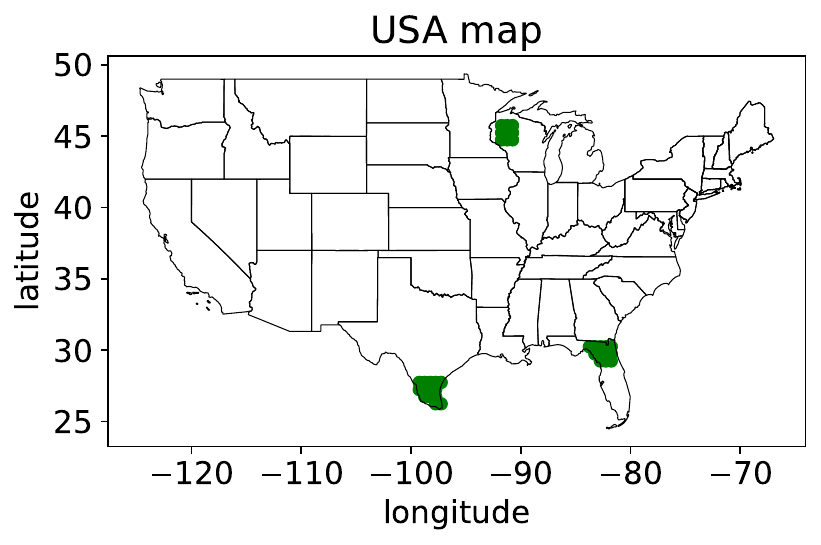}
        \caption[]{\textbf{Regions where the temperature forecast is analyzed.} \newold{See Appendix \ref{sec:tmp_regions_analysis} for details.} 
        }
        \label{fig:regions_tmp_analysis}
\end{figure}

\Cref{fig:tmp_reg_ncep_test} shows regions in Texas and Florida where the ensemble mean and linear regression performance is poor, while a random forest achieves far superior performance. We conduct an analysis of forecasts of the ensemble mean, linear regression, and random forests in these regions together with a region in Wisconsin where all methods show good performance. \Cref{fig:regions_tmp_analysis} indicates these regions and \cref{table:regions_performance}
summarizes the performance of different methods in these regions: the ensemble mean prediction quality dramatically drops between the validation and test periods in Texas and Florida, which is not the case for the random forest. 

\begin{table}[ht!]
\caption{\textbf{Train, validation, and test performance of different methods in Texas, Florida, and Wisconsin regions.} The task is temperature regression; NCEP-CFSv2 dataset is used. The performance of the ensemble mean and linear regression in the test period significantly decreases in Texas and Florida while the random forest is able to demonstrate reasonable results. All methods perform well in Wisconsin.}
\label{table:regions_performance}
\centering
\small
\begin{tabular}{c c  c  c  c  c  c  c  c } 
\hline

 \makecell{Region \\ location} & Model & \makecell{Train \\ mean $R^2$ ($\uparrow$)} & \makecell{Validation \\ mean $R^2$ ($\uparrow$)} & \makecell{Test \\ mean $R^2$ ($\uparrow$)} \\ %
 \hline 
& Ens mean & 0.19 & 0.36 & -1.55 \\ 
\multirow{-2}{*}{Texas}
 &  LR & 0.53 & 0.49 & -1.29 \\ 
&  RF & 0.97 & 0.32 & -0.33 \\ 
\hline 
&  Ens mean & 0.11 & 0.34 & -0.87 \\ 
\multirow{-2}{*}{Florida}
 &  LR & 0.47 & 0.58 & -0.56 \\ 
 &  RF & 0.97 & 0.36 & 0.11 \\
\hline
&  Ens mean & 0.30 & 0.36 & 0.39\\
\multirow{-2}{*}{Wisconsin}
 &  LR &  0.53 & 0.57 & 0.51\\ 
 &  RF &  1.00 & 0.47 & 0.47\\
\hline
\end{tabular}
\end{table}

\begin{figure}[ht]
 \centerline{\includegraphics[width=27pc]{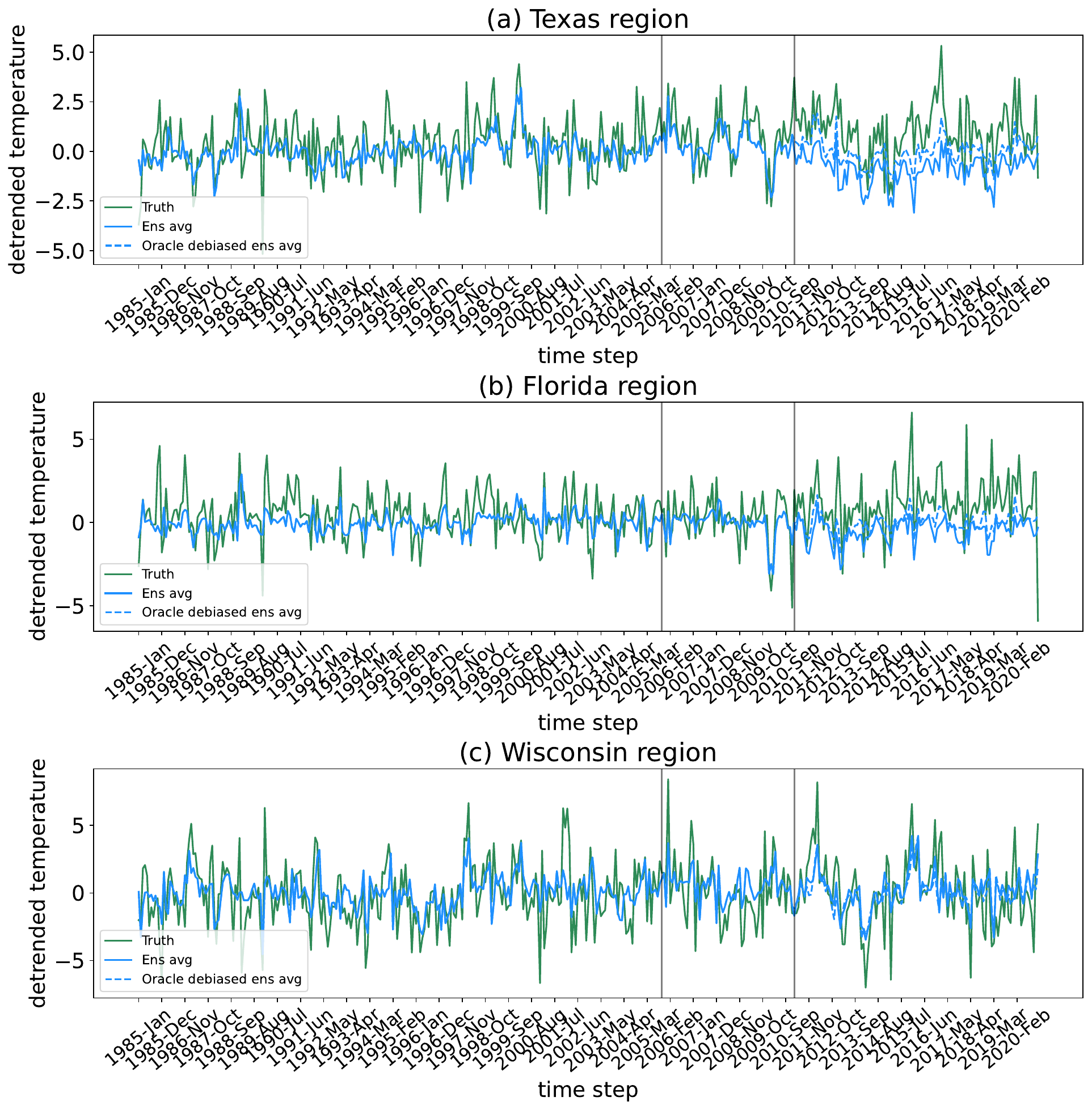}}
  \caption{\textbf{Temperature predictions \newold{in $ ^\circ C $} of different methods at Texas, Florida, and Wisconsin regions.} Black lines correspond to train/val and val/test splits; train and validation correspond to the hindcast regime of the ensemble, while test corresponds to the forecast regime. \newold{See Appendix \ref{sec:tmp_regions_analysis} for details.}  }\label{fig:tx_fl_preds}
\end{figure}

Why does RF perform so much better than simpler methods in some regions? One possibility is that the RF is a nonlinear model capable of more complex predictions. However, if that were the only cause of the discrepancy in performance, then we would expect that the RF would be better not only during the test period, but during the validation period as well. \cref{table:regions_performance} does not support this argument; it shows that the ensemble mean and linear regression have comparable, if not superior, performance to the random forest during the validation period.  A second hypothesis is that the distribution of temperature is different during the test period than during the training and validation periods. This hypothesis is plausible for two reasons: (1) climate change, and (2) the training and validation data use hindcast ensembles while the test data uses forecast ensembles. To investigate this hypothesis, in \Cref{fig:tx_fl_preds} we plot the true temperature and ensemble mean in the training, validation, and test periods for the three geographic regions. The discrepancy between the true temperatures and ensemble means in the test period is generally greater than during the training and validation periods in Texas and Florida (though not in Wisconsin, a region where validation and test performance are comparable for all methods). This lends support to the hypothesis that hindcast and forecast ensembles exhibit distribution drift, and the superior performance of the RF during the test period may be due to a greater robustness to that distribution drift.

The hindcast and forecast ensembles may have different predictive accuracies because the hindcast ensembles have been debiased to fit past observations -- a procedure not possible for forecast data.  To explore the potential impact of debiasing, \Cref{fig:tx_fl_preds}  shows the ``oracle debiased ensemble mean", which is computed by using the test data to estimate the forecast ensemble bias and subtracting it from the ensemble mean. This procedure, \textit{which would not be possible in practice and is used only to probe distribution drift  ensemble bias}, yields smaller discrepancies between the true data and the (oracle debiased) ensemble mean 
than the discrepancies between the true data and the original (biased) ensemble mean.
Specifically, the oracle ensemble member achieves -0.20 mean $R^2$ score (TX) and -0.28 mean $R^2$ score (FL) vs.\ -1.55 $R^2$ (TX) and -0.87 $R^2$ (FL) of the original forecast ensemble mean.
\newold{The errors during the test period are generally larger than during the train and validation period, even after debiasing the ensemble members using future data. This effect may be attributed both to (a) the nonstationarity of the climate (note that there are more extreme values during the test period than during the training and validation periods, particularly in Texas and Florida) and (b) the fact that in the train and validation periods, we use hindcast ensemble members, whereas in the test period, we use forecast ensemble members.}

\section{Machine learning architectures}\label{app:architecture}
\subsection{U-Net details}
\begin{figure}[!hbt]
    \centering
    \includegraphics[width=27pc]{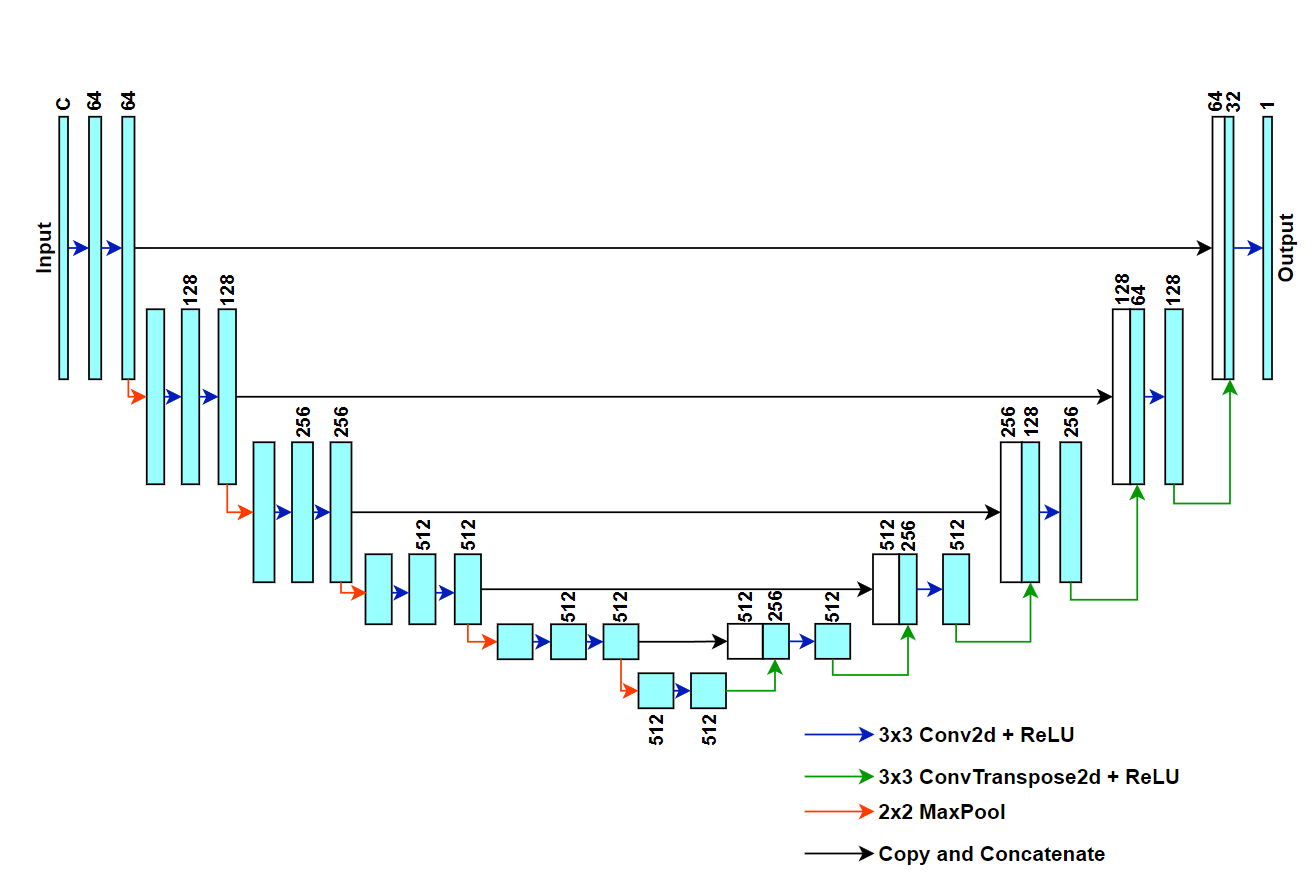}
    \caption{U-Net architecture with input channels = $C$. $C$ is the number of input channels, which, in our case, equals the number of ensemble members plus all climate data. }
    \label{fig:UNET}
\end{figure}

The U-Net has residual connections from layers in the encoder part to the decoder part in a paired way so that it forms a U-shape. Figure
\ref{fig:UNET} shows the architecture of the U-Net. The U-Net is a powerful deep convolutional network that is widely used in image processing tasks such as image segmentation \citep{Ronneberger2015UNetCN, hao2020brief} or style transfer \citep{gatys2016image, jing2019neural}.

Our U-Net differs from the original U-Net by modifying the first 2D convolutional layer after input. Since our input channels can be different when we choose a different subset of features or different ensemble (NCEP-CFSv2 or NASA-GMAO), this 2D convolutional layer is used to transform our input with $C$ channels into a latent representation with 64 channels. The number of channels $C$ depends on which ensemble we are using and what task we are performing. For example, for precipitation tasks using the NCEP-CFSv2 ensemble, the input channels include $24$ ensemble members, $5$ lagged observations, $4$ other observational variables, $8$ principal components of SSTs, and 24 positional encodings, resulting in $65$ channels in total. For temperature tasks using the NASA-GMAO ensemble, there are only $11$ ensemble members, and we don't include SSTs information, hence there are only $44$ channels in total. The other following layers use the same configurations with the standard U-Net \cite{Ronneberger2015UNetCN}. 

We also perform careful hyperparameter tuning for the U-NET. In particular, we run a 10-fold cross-validation on our training set, and use grid search for tuning learning rate, batch size, number of epochs, and weight decay. Since we use different loss functions for different forecast tasks and different numbers of input channels for NCEP-CFSv2 and GMAO-GMAO ensemble, we run hyperparameter tuning with the same cross-validation scheme separately for these tasks. For instance, for precipitation regression, we choose from 100, 120, 150, 170, 200, and 250 epochs; batch size may be equal to 8, 16, 32; learning rate values are chosen from 0.0001, 0.001, 0.01; weight decay can be 0, 0.001, 0.0001. In case of NCEP-CFSv2 precipitation regression, the optimal parameters are 170 epochs, batch size 16, learning rate 0.0001, and weight decay 0.0001. For temperature regression using the same data, the best parameters are 100 epochs, batch size 16, learning rate 0.001, and weight decay 0.001. For tercile classification of precipitation, the best parameters are 80 epochs (we chose from 60, 70, 80, 90, and 100 epochs during classification), batch size 8, learning rate 0.001, and weight decay 0.0001.

\subsection{Random Forest Quantile Regressor details}
\label{RFQR}
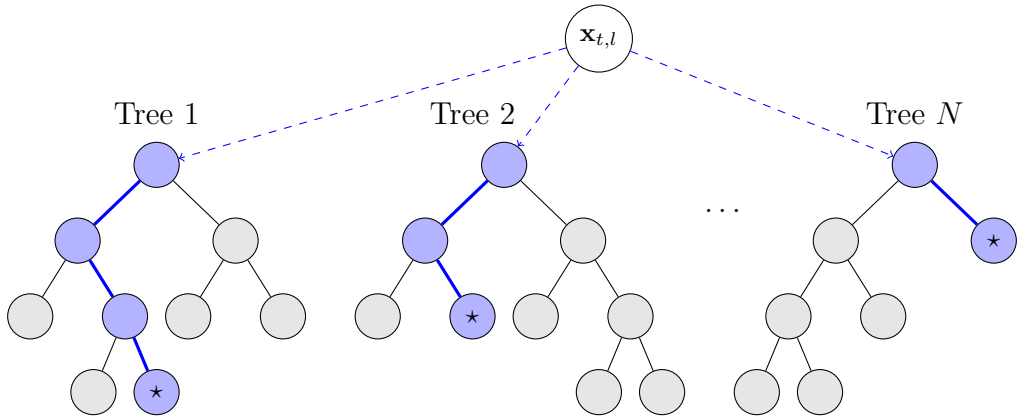
\begin{figure}
    \centering
\begin{tikzpicture}[level distance=1.2cm,
level 1/.style={sibling distance=2.5cm},
level 2/.style={sibling distance=1.5cm},
level 3/.style={sibling distance=1cm}, 
every edge/.style={draw},
scale=0.84]
\tikzstyle{input} = [circle, draw, minimum size=0.6cm]
\tikzstyle{node} = [circle, draw, fill=black!10, minimum size=0.6cm]
\tikzstyle{decision} = [circle, draw, fill=black!10, minimum size = 0.6cm]
\tikzstyle{activedecision} = [circle, draw, fill=blue!30, minimum size = 0.6cm]
\tikzstyle{leaf} = [circle, draw, fill=black!10, minimum size = 0.6cm]
\tikzstyle{output} = [circle, draw, fill=blue!30, minimum size = 0.6cm]
\tikzstyle{tree_label} = [above, font=\large]

\node at (0,0) [activedecision] (L14) {} 
    child {node [activedecision] (L13){}
        child {node [leaf]  { }[thin]}
        child {node [activedecision] (L12){}
            child {node [leaf] { }[thin]}
            child {node [output] (L11) { $\star$ }}
        } 
    } 
    child {node [decision] {}
        child {node [leaf] { }}
        child {node [leaf] { }}
    };
\node at (0, 0.5) [tree_label] {Tree 1};
\node at (5.5,0) [activedecision] (L23){}
    child {node [activedecision] (L22){}
        child {node [leaf] { }}
        child {node [output] (L21) { $\star$ }
        }
    }
    child {node [decision] {}
        child {node [leaf] { }}
        child {node [decision] {}
            child {node [leaf] { }
            }child {node [leaf] { }}
        }
    };
\node at (5.0, 0.5) [tree_label] {Tree 2};
\node at (12,0) [activedecision] (L33) {}
    child {node [decision] {}
        child {node [decision] {}
            child {node [leaf] { }}
            child {node [leaf] { }}
        }
        child {node [leaf] { }}
    }
    child {node [output] (L31) { $\star$ }
    edge from parent[very thick, blue]};
\node at (12, 0.5) [tree_label] {Tree $N$};

\node at (9, -0.9) [tree_label] {$\dots$};

\node at (7,2) [input] (X) {$\bx_{t,l}$};

\draw[->,dashed,blue] (X) -- (L14);
\draw[->,dashed,blue] (X) -- (L23);
\draw[->,dashed,blue] (X) -- (L33);

\begin{scope}[very thick, blue]
  \draw (L14) --(L13) -- (L12) -- (L11);
  \draw (L23) -- (L22) -- (L21);
\end{scope}

\end{tikzpicture}

\caption{\newold{Illustration of a random forest, which serves as a visual aid for our discussion. \textbf{Quantile regression forests} outputs the empirical $\alpha$ quantiles of the collection of all responses in all the leaves associated with $\bx_{t,l}$ (marked with a star for each tree).}}\label{fig:RFQR}
\end{figure}

\newold{We show a figure representation of the RFQR in \cref{fig:RFQR}. The RFQR is essentially trained as a regular random forest, but it makes a quantile estimate by taking the sample quantile of the responses in all leaves associated with a new input.}

\subsection{Stacking model details} \label{sec:stacking_arch}

The stacking model is a simple one-layer neural network with 100 hidden neurons and a sigmoid activation function for regression and softmax for classification. We use an implementation from Scikit-learn library \citep{scikit-learn}. We choose 100 neurons based on the stacking model performance on the validation data (we also try 50, 75, 100 and 120 neurons). The stacking model demonstrates stable performance in general, but with 100 neurons it usually achieves the best results. We use the ``lbfgs" optimizer from quasi-Newton methods for the regression tasks, and the SGD optimizer for classification tasks.

\section{Preprocessing Details} \label{app:preprocessing}

Random forest and U-Net require different input formats. For U-Net, all input variables have natural image representation except SSTs and information about location. For example, ensemble predictions can be represented as a tensor of shape $(K, W, H)$, where $K$ corresponds to the number of ensemble members (or number of channels of an image), and $W$ and $H$ are width and height of the corresponding image. In our case, $W=64$ and $H=128$. For the U-Net model, we handle the missing land variables over the sea regions by the nearest neighbor interpolation of available values.

\paragraph{Sea surface temperatures}
There are more than 100,000 SSTs locations available. We extract the top eight principal components. Principal component analysis fits on the train part and then is applied to the rest of the data. In the case of U-Net, we deal with PCs of SSTs by adding additional input channels that are constant across space, with each channel corresponding to one of PCs. Random forest can use PCs from SSTs directly with no special preprocessing.

\paragraph{Normalization}
We apply channel-wise min-max normalization to the input features at each location based on the training part of the dataset in the case of U-Net. As for normalization of the true values, min-max normalization is applied for precipitation, and standardization is applied for temperature. This choice affects the final layer of the U-Net model, too: for the precipitation regression task, a sigmoid activation is used, and no activation function is applied for temperature regression. For the stacking model, we apply min-max normalization to both input and target values.

\newpage

%% file: main.bbl
\begin{thebibliography}{78}
\providecommand{\natexlab}[1]{#1}
\providecommand{\url}[1]{\texttt{#1}}
\expandafter\ifx\csname urlstyle\endcsname\relax
  \providecommand{\doi}[1]{doi: #1}\else
  \providecommand{\doi}{doi: \begingroup \urlstyle{rm}\Url}\fi

\bibitem[Ban et~al.(2016)Ban, Bitz, Brown, Chassignet, Dutton, Hallberg,
  Kamrath, Kleist, Lermusiaux, Lin, Myers, Pullen, Sandgathe, Shafer, Waliser,
  and Zhang]{nas_subseasonal}
R.~J Ban, C.~M. Bitz, A.~Brown, E.~Chassignet, J.~A. Dutton, R.~Hallberg,
  A.~Kamrath, D.~Kleist, P.~F.~J. Lermusiaux, H.~Lin, L.~Myers, J.~Pullen,
  S.~Sandgathe, M.~Shafer, D.~Waliser, and C.~Zhang.
\newblock \emph{Next Generation Earth System Prediction: Strategies for
  Subseasonal to Seasonal Forecasts.}
\newblock The National Academies Press, 2016.

\bibitem[Lorenc(1986)]{lorenc1986analysis}
Andrew~C Lorenc.
\newblock Analysis methods for numerical weather prediction.
\newblock \emph{Quarterly Journal of the Royal Meteorological Society},
  112\penalty0 (474):\penalty0 1177--1194, 1986.

\bibitem[{National Academies of Sciences}(2016)]{boar16}
{National Academies of Sciences}.
\newblock \emph{Next generation earth system prediction: strategies for
  subseasonal to seasonal forecasts}.
\newblock National Academies Press, 2016.

\bibitem[{National Research Council}(2010)]{nati10}
{National Research Council}.
\newblock \emph{Assessment of intraseasonal to interannual climate prediction
  and predictability}.
\newblock National Academies Press, 2010.

\bibitem[Simmons and Hollingsworth(2002)]{simmholl2002}
A.~J. Simmons and A.~Hollingsworth.
\newblock Some aspects of the improvement in skill of numerical weather
  prediction.
\newblock \emph{Quarterly Journal of the Royal Meteorological Society},
  128\penalty0 (580):\penalty0 647--677, 2002.

\bibitem[Barnston et~al.(2012)Barnston, Tippett, L'Heureux, Li, and
  DeWitt]{barnston2012skill}
Anthony~G Barnston, Michael~K Tippett, Michelle~L L'Heureux, Shuhua Li, and
  David~G DeWitt.
\newblock Skill of real-time seasonal enso model predictions during 2002--11:
  Is our capability increasing?
\newblock \emph{Bulletin of the American Meteorological Society}, 93\penalty0
  (5):\penalty0 631--651, 2012.

\bibitem[Brunet et~al.(2010)Brunet, Shapiro, Hoskins, Moncrieff, Dole, Kiladis,
  Kirtman, Lorenc, Mills, Morss, et~al.]{brunet2010collaboration}
Gilbert Brunet, Melvyn Shapiro, Brian Hoskins, Mitch Moncrieff, Randall Dole,
  George~N Kiladis, Ben Kirtman, Andrew Lorenc, Brian Mills, Rebecca Morss,
  et~al.
\newblock Collaboration of the weather and climate communities to advance
  subseasonal-to-seasonal prediction.
\newblock \emph{Bulletin of the American Meteorological Society}, 91\penalty0
  (10):\penalty0 1397--1406, 2010.

\bibitem[White et~al.(2022)White, Domeisen, Acharya, Adefisan, Anderson, Aura,
  Balogun, Bertram, Bluhm, Brayshaw, et~al.]{white2022advances}
Christopher~J White, Daniela~IV Domeisen, Nachiketa Acharya, Elijah~A Adefisan,
  Michael~L Anderson, Stella Aura, Ahmed~A Balogun, Douglas Bertram, Sonia
  Bluhm, David~J Brayshaw, et~al.
\newblock Advances in the application and utility of subseasonal-to-seasonal
  predictions.
\newblock \emph{Bulletin of the American Meteorological Society}, 103\penalty0
  (6):\penalty0 E1448--E1472, 2022.

\bibitem[Mouatadid et~al.(2023)Mouatadid, Orenstein, Flaspohler, Cohen,
  Oprescu, Fraenkel, and Mackey]{mouatadid2023adaptive}
Soukayna Mouatadid, Paulo Orenstein, Genevieve Flaspohler, Judah Cohen, Miruna
  Oprescu, Ernest Fraenkel, and Lester Mackey.
\newblock Adaptive bias correction for improved subseasonal forecasting.
\newblock \emph{Nature Communications}, 14\penalty0 (1):\penalty0 3482, 2023.

\bibitem[Haupt et~al.(2021)Haupt, Chapman, Adams, Kirkwood, Hosking, Robinson,
  Lerch, and Subramanian]{haupt2021towards}
Sue~Ellen Haupt, William Chapman, Samantha~V Adams, Charlie Kirkwood, J~Scott
  Hosking, Niall~H Robinson, Sebastian Lerch, and Aneesh~C Subramanian.
\newblock {Towards implementing artificial intelligence post-processing in
  weather and climate: proposed actions from the Oxford 2019 workshop}.
\newblock \emph{Philosophical Transactions of the Royal Society A:
  Mathematical, Physical and Engineering Sciences}, 379\penalty0
  (2194):\penalty0 20200091, 2021.

\bibitem[Vannitsem et~al.(2021)Vannitsem, Bremnes, Demaeyer, Evans, Flowerdew,
  Hemri, Lerch, Roberts, Theis, Atencia, et~al.]{vannitsem2021statistical}
S~Vannitsem, JB~Bremnes, J~Demaeyer, GR~Evans, J~Flowerdew, S~Hemri, S~Lerch,
  N~Roberts, S~Theis, A~Atencia, et~al.
\newblock Statistical postprocessing for weather forecasts: Review, challenges,
  and avenues in a big data world.
\newblock \emph{Bulletin of the American Meteorological Society}, 102\penalty0
  (3):\penalty0 E681–E699, 2021.

\bibitem[Hwang et~al.(2018)Hwang, Orenstein, Pfeiffer, Cohen, and
  Mackey]{rodeo}
Jessica Hwang, Paulo Orenstein, Karl Pfeiffer, Judah Cohen, and Lester Mackey.
\newblock Improving subseasonal forecasting in the western us with machine
  learning, 2018.

\bibitem[Hwang et~al.(2019)Hwang, Orenstein, Cohen, Pfeiffer, and
  Mackey]{hwang2019improving}
Jessica Hwang, Paulo Orenstein, Judah Cohen, Karl Pfeiffer, and Lester Mackey.
\newblock Improving subseasonal forecasting in the western us with machine
  learning.
\newblock In \emph{Proceedings of the 25th ACM SIGKDD International Conference
  on Knowledge Discovery \& Data Mining}, pages 2325--2335, 2019.

\bibitem[Kirtman et~al.(2014)Kirtman, Min, Infanti, Kinter, Paolino, Zhang, Van
  Den~Dool, Saha, Mendez, Becker, et~al.]{kirtman2014north}
Ben~P Kirtman, Dughong Min, Johnna~M Infanti, James~L Kinter, Daniel~A Paolino,
  Qin Zhang, Huug Van Den~Dool, Suranjana Saha, Malaquias~Pena Mendez, Emily
  Becker, et~al.
\newblock {The North American multimodel ensemble: phase-1
  seasonal-to-interannual prediction; phase-2 toward developing intraseasonal
  prediction}.
\newblock \emph{Bulletin of the American Meteorological Society}, 95\penalty0
  (4):\penalty0 585--601, 2014.

\bibitem[Saha et~al.(2014)Saha, Moorthi, Wu, Wang, Nadiga, Tripp, Behringer,
  Hou, Chuang, Iredell, et~al.]{saha2014ncep}
Suranjana Saha, Shrinivas Moorthi, Xingren Wu, Jiande Wang, Sudhir Nadiga,
  Patrick Tripp, David Behringer, Yu-Tai Hou, Hui-ya Chuang, Mark Iredell,
  et~al.
\newblock {The NCEP climate forecast system version 2}.
\newblock \emph{Journal of climate}, 27\penalty0 (6):\penalty0 2185--2208,
  2014.

\bibitem[Nakada et~al.(2018)Nakada, Kovach, Marshak, and Molod]{nasagmao}
Kazumi Nakada, Robin~M Kovach, Jelena Marshak, and Andrea Molod.
\newblock {Global modeling and assimilation office - NASA}, Apr 2018.
\newblock URL \url{https://gmao.gsfc.nasa.gov/pubs/docs/Nakada1033.pdf}.

\bibitem[Nebeker(1995)]{nebeker1995calculating}
Frederik Nebeker.
\newblock \emph{Calculating the weather: Meteorology in the 20th century}.
\newblock Elsevier, 1995.

\bibitem[Chen et~al.(2023)Chen, Han, Wang, Zhao, Yang, and
  Yang]{chen2023machine}
Liuyi Chen, Bocheng Han, Xuesong Wang, Jiazhen Zhao, Wenke Yang, and Zhengyi
  Yang.
\newblock Machine learning methods in weather and climate applications: A
  survey.
\newblock \emph{Applied Sciences}, 13\penalty0 (21):\penalty0 12019, 2023.

\bibitem[Nagaraj and Kumar(2023)]{nagaraj2023univariate}
R~Nagaraj and Lakshmi~Sutha Kumar.
\newblock Univariate deep learning models for prediction of daily average
  temperature and relative humidity: The case study of chennai, india.
\newblock \emph{Journal of Earth System Science}, 132\penalty0 (3):\penalty0
  100, 2023.

\bibitem[Frnda et~al.(2022)Frnda, Durica, Rozhon, Vojtekova, Nedoma, and
  Martinek]{frnda2022ecmwf}
Jaroslav Frnda, Marek Durica, Jan Rozhon, Maria Vojtekova, Jan Nedoma, and
  Radek Martinek.
\newblock {ECMWF short-term prediction accuracy improvement by deep learning}.
\newblock \emph{Scientific Reports}, 12\penalty0 (1):\penalty0 7898, 2022.

\bibitem[Herman and Schumacher(2018)]{herman2018dendrology}
Gregory~R Herman and Russ~S Schumacher.
\newblock {“Dendrology” in numerical weather prediction: What random
  forests and logistic regression tell us about forecasting extreme
  precipitation}.
\newblock \emph{Monthly Weather Review}, 146\penalty0 (6):\penalty0 1785--1812,
  2018.

\bibitem[Ghaderi et~al.(2017)Ghaderi, Sanandaji, and Ghaderi]{ghaderi2017deep}
Amir Ghaderi, Borhan~M Sanandaji, and Faezeh Ghaderi.
\newblock Deep forecast: Deep learning-based spatio-temporal forecasting, 2017.

\bibitem[Grover et~al.(2015)Grover, Kapoor, and Horvitz]{grover2015deep}
Aditya Grover, Ashish Kapoor, and Eric Horvitz.
\newblock A deep hybrid model for weather forecasting.
\newblock In \emph{Proceedings of the 21th ACM SIGKDD international conference
  on knowledge discovery and data mining}, pages 379--386, 2015.

\bibitem[Radhika and Shashi(2009)]{radhika2009atmospheric}
Y~Radhika and M~Shashi.
\newblock Atmospheric temperature prediction using support vector machines.
\newblock \emph{International journal of computer theory and engineering},
  1\penalty0 (1):\penalty0 55, 2009.

\bibitem[Cof{\i}no et~al.(2002)Cof{\i}no, Cano, Sordo, and
  Gutierrez]{cofino2002bayesian}
Antonio~S Cof{\i}no, Rafael Cano, Carmen Sordo, and Jose~M Gutierrez.
\newblock Bayesian networks for probabilistic weather prediction.
\newblock In \emph{15th Eureopean Conference on Artificial Intelligence
  (ECAI)}, 2002.

\bibitem[Lam et~al.(2023)Lam, Sanchez-Gonzalez, Willson, Wirnsberger,
  Fortunato, Alet, Ravuri, Ewalds, Eaton-Rosen, Hu, et~al.]{lam2023learning}
Remi Lam, Alvaro Sanchez-Gonzalez, Matthew Willson, Peter Wirnsberger, Meire
  Fortunato, Ferran Alet, Suman Ravuri, Timo Ewalds, Zach Eaton-Rosen, Weihua
  Hu, et~al.
\newblock Learning skillful medium-range global weather forecasting.
\newblock \emph{Science}, page eadi2336, 2023.

\bibitem[Yang et~al.(2023)Yang, Ling, Li, and Luo]{yang2023improving}
Shuxian Yang, Fenghua Ling, Yue Li, and Jing-Jia Luo.
\newblock Improving seasonal prediction of summer precipitation in the
  middle--lower reaches of the yangtze river using a tu-net deep learning
  approach.
\newblock \emph{Artificial Intelligence for the Earth Systems}, 2\penalty0
  (2):\penalty0 220078, 2023.

\bibitem[Hewage et~al.(2021)Hewage, Trovati, Pereira, and
  Behera]{hewage2021deep}
Pradeep Hewage, Marcello Trovati, Ella Pereira, and Ardhendu Behera.
\newblock Deep learning-based effective fine-grained weather forecasting model.
\newblock \emph{Pattern Analysis and Applications}, 24\penalty0 (1):\penalty0
  343--366, 2021.

\bibitem[Cohen et~al.(2019)Cohen, Coumou, Hwang, Mackey, Orenstein, Totz, and
  Tziperman]{cohen2019s2s}
Judah Cohen, Dim Coumou, Jessica Hwang, Lester Mackey, Paulo Orenstein, Sonja
  Totz, and Eli Tziperman.
\newblock S2s reboot: An argument for greater inclusion of machine learning in
  subseasonal to seasonal forecasts.
\newblock \emph{Wiley Interdisciplinary Reviews: Climate Change}, 10\penalty0
  (2):\penalty0 e00567, 2019.

\bibitem[Totz et~al.(2017)Totz, Tziperman, Coumou, Pfeiffer, and
  Cohen]{totz2017winter}
Sonja Totz, Eli Tziperman, Dim Coumou, Karl Pfeiffer, and Judah Cohen.
\newblock {Winter precipitation forecast in the European and Mediterranean
  regions using cluster analysis}.
\newblock \emph{Geophysical Research Letters}, 44\penalty0 (24):\penalty0
  12--418, 2017.

\bibitem[Iglesias et~al.(2015)Iglesias, Kale, and Liu]{iglesias2015examination}
Gilberto Iglesias, David~C Kale, and Yan Liu.
\newblock An examination of deep learning for extreme climate pattern analysis.
\newblock In \emph{The 5th International Workshop on Climate Informatics},
  2015.

\bibitem[Badr et~al.(2014)Badr, Zaitchik, and Guikema]{badr2014application}
Hamada~S Badr, Benjamin~F Zaitchik, and Seth~D Guikema.
\newblock Application of statistical models to the prediction of seasonal
  rainfall anomalies over the sahel.
\newblock \emph{Journal of Applied meteorology and climatology}, 53\penalty0
  (3):\penalty0 614--636, 2014.

\bibitem[Vitart et~al.(2012)Vitart, Robertson, and
  Anderson]{vitart2012subseasonal}
Fr{\'e}d{\'e}ric Vitart, Andrew~W Robertson, and David~LT Anderson.
\newblock Subseasonal to seasonal prediction project: Bridging the gap between
  weather and climate.
\newblock \emph{Bulletin of the World Meteorological Organization}, 61\penalty0
  (2):\penalty0 23, 2012.

\bibitem[Min et~al.(2020)Min, Ham, Yoo, and Han]{min2020recent}
Young-Mi Min, Suryun Ham, Jin-Ho Yoo, and Su-Hee Han.
\newblock Recent progress and future prospects of subseasonal and seasonal
  climate predictions.
\newblock \emph{Bulletin of the American Meteorological Society}, 101\penalty0
  (5):\penalty0 E640--E644, 2020.

\bibitem[He et~al.(2020)He, Li, DelSole, Ravikumar, and Banerjee]{he2020sub}
Sijie He, Xinyan Li, Timothy DelSole, Pradeep Ravikumar, and Arindam Banerjee.
\newblock Sub-seasonal climate forecasting via machine learning: Challenges,
  analysis, and advances, 2020.

\bibitem[Srinivasan et~al.(2021)Srinivasan, Khim, Banerjee, and
  Ravikumar]{srinivasan2021subseasonal}
Vishwak Srinivasan, Justin Khim, Arindam Banerjee, and Pradeep Ravikumar.
\newblock {Subseasonal climate prediction in the Western US using bayesian
  spatial models}.
\newblock In \emph{Uncertainty in artificial intelligence}, pages 961--970.
  PMLR, 2021.

\bibitem[He et~al.(2021)He, Li, Trenary, Cash, DelSole, and
  Banerjee]{he2021learning}
Sijie He, Xinyan Li, Laurie Trenary, Benjamin~A Cash, Timothy DelSole, and
  Arindam Banerjee.
\newblock Learning and dynamical models for sub-seasonal climate forecasting:
  Comparison and collaboration, 2021.

\bibitem[Gr{\"{o}}nquist et~al.(2020)Gr{\"{o}}nquist, Yao, Ben{-}Nun, Dryden,
  Dueben, Li, and Hoefler]{GrEnsShort}
Peter Gr{\"{o}}nquist, Chengyuan Yao, Tal Ben{-}Nun, Nikoli Dryden, Peter
  Dueben, Shigang Li, and Torsten Hoefler.
\newblock Deep learning for post-processing ensemble weather forecasts.
\newblock \emph{CoRR}, abs/2005.08748, 2020.
\newblock URL \url{https://arxiv.org/abs/2005.08748}.

\bibitem[Loken et~al.(2022)Loken, Clark, and McGovern]{LokenRF}
Eric~D. Loken, Adam~J. Clark, and Amy McGovern.
\newblock Comparing and interpreting differently designed random forests for
  next-day severe weather hazard prediction.
\newblock \emph{Weather and Forecasting}, 37\penalty0 (6):\penalty0 871 -- 899,
  2022.
\newblock \doi{https://doi.org/10.1175/WAF-D-21-0138.1}.
\newblock URL
  \url{https://journals.ametsoc.org/view/journals/wefo/37/6/WAF-D-21-0138.1.xml}.

\bibitem[NOAA(2022)]{noaa_climate}
NOAA.
\newblock {NOAA National Centers for Environmental information, Climate at a
  Glance: National Time Series}.
\newblock \url{https://www.ncdc.noaa.gov/cag/}, 2022.

\bibitem[Fan and Van~den Dool(2008)]{fan2008global}
Yun Fan and Huug Van~den Dool.
\newblock A global monthly land surface air temperature analysis for
  1948--present.
\newblock \emph{Journal of Geophysical Research: Atmospheres}, 113\penalty0
  (D1), 2008.

\bibitem[Xie et~al.(2010)Xie, Chen, and Shi]{xie2010cpc}
P~Xie, M~Chen, and W~Shi.
\newblock Cpc global unified gauge-based analysis of daily precipitation.
\newblock In \emph{Preprints, 24th Conf. on Hydrology, Atlanta, GA, Amer.
  Metero. Soc}, volume~2, 2010.

\bibitem[Reynolds et~al.(2007)Reynolds, Smith, Liu, Chelton, Casey, and
  Schlax]{reynolds2007daily}
Richard~W Reynolds, Thomas~M Smith, Chunying Liu, Dudley~B Chelton, Kenneth~S
  Casey, and Michael~G Schlax.
\newblock Daily high-resolution-blended analyses for sea surface temperature.
\newblock \emph{Journal of climate}, 20\penalty0 (22):\penalty0 5473--5496,
  2007.

\bibitem[Kalnay et~al.(1996)Kalnay, Kanamitsu, Kistler, Collins, Deaven,
  Gandin, Iredell, Saha, White, Woollen, et~al.]{kalnay1996ncep}
Eugenia Kalnay, Masao Kanamitsu, Robert Kistler, William Collins, Dennis
  Deaven, Lev Gandin, Mark Iredell, Suranjana Saha, Glenn White, John Woollen,
  et~al.
\newblock {The NCEP/NCAR 40-year reanalysis project}.
\newblock \emph{Bulletin of the American meteorological Society}, 77\penalty0
  (3):\penalty0 437--472, 1996.

\bibitem[Meinshausen(2006)]{RFQR:2006}
Nicolai Meinshausen.
\newblock Quantile regression forests.
\newblock \emph{Journal of Machine Learning Research}, 7\penalty0
  (35):\penalty0 983--999, 2006.
\newblock URL \url{http://jmlr.org/papers/v7/meinshausen06a.html}.

\bibitem[Hastie et~al.(2009)Hastie, Tibshirani, Friedman, and
  Friedman]{hastie2009elements}
Trevor Hastie, Robert Tibshirani, Jerome~H Friedman, and Jerome~H Friedman.
\newblock \emph{The elements of statistical learning: data mining, inference,
  and prediction}, volume~2.
\newblock Springer, 2009.

\bibitem[Biau and Scornet(2016)]{biau2016random}
G{\'e}rard Biau and Erwan Scornet.
\newblock A random forest guided tour.
\newblock \emph{Test}, 25:\penalty0 197--227, 2016.

\bibitem[Pedregosa et~al.(2011)Pedregosa, Varoquaux, Gramfort, Michel, Thirion,
  Grisel, Blondel, Prettenhofer, Weiss, Dubourg, Vanderplas, Passos,
  Cournapeau, Brucher, Perrot, and Duchesnay]{scikit-learn}
F.~Pedregosa, G.~Varoquaux, A.~Gramfort, V.~Michel, B.~Thirion, O.~Grisel,
  M.~Blondel, P.~Prettenhofer, R.~Weiss, V.~Dubourg, J.~Vanderplas, A.~Passos,
  D.~Cournapeau, M.~Brucher, M.~Perrot, and E.~Duchesnay.
\newblock Scikit-learn: Machine learning in {P}ython.
\newblock \emph{Journal of Machine Learning Research}, 12:\penalty0 2825--2830,
  2011.

\bibitem[Ronneberger et~al.(2015)Ronneberger, Fischer, and
  Brox]{Ronneberger2015UNetCN}
Olaf Ronneberger, Philipp Fischer, and Thomas Brox.
\newblock {U-Net: Convolutional Networks for Biomedical Image Segmentation}.
\newblock In \emph{MICCAI}, 2015.

\bibitem[Yakubovskiy(2020)]{Yakubovskiy:2019}
Pavel Yakubovskiy.
\newblock Segmentation models pytorch.
\newblock \url{https://github.com/qubvel/segmentation_models.pytorch}, 2020.

\bibitem[Kingma and Ba(2014)]{kingma2014adam}
Diederik~P Kingma and Jimmy Ba.
\newblock Adam: A method for stochastic optimization, 2014.

\bibitem[Pavlyshenko(2018)]{pavlyshenko2018using}
Bohdan Pavlyshenko.
\newblock Using stacking approaches for machine learning models.
\newblock In \emph{2018 IEEE Second International Conference on Data Stream
  Mining \& Processing (DSMP)}, pages 255--258. IEEE, 2018.

\bibitem[Vaswani et~al.(2017)Vaswani, Shazeer, Parmar, Uszkoreit, Jones, Gomez,
  Kaiser, and Polosukhin]{vaswani2017attention}
Ashish Vaswani, Noam Shazeer, Niki Parmar, Jakob Uszkoreit, Llion Jones,
  Aidan~N Gomez, {\L}ukasz Kaiser, and Illia Polosukhin.
\newblock Attention is all you need.
\newblock \emph{Advances in neural information processing systems}, 30, 2017.

\bibitem[Devlin et~al.(2018)Devlin, Chang, Lee, and Toutanova]{devlin2018bert}
Jacob Devlin, Ming-Wei Chang, Kenton Lee, and Kristina Toutanova.
\newblock Bert: Pre-training of deep bidirectional transformers for language
  understanding, 2018.

\bibitem[Petroni et~al.(2019)Petroni, Rockt{\"a}schel, Lewis, Bakhtin, Wu,
  Miller, and Riedel]{petroni2019language}
Fabio Petroni, Tim Rockt{\"a}schel, Patrick Lewis, Anton Bakhtin, Yuxiang Wu,
  Alexander~H Miller, and Sebastian Riedel.
\newblock Language models as knowledge bases?, 2019.

\bibitem[Narayanan et~al.(2016)Narayanan, Chandramohan, Chen, Liu, and
  Saminathan]{narayanan2016subgraph2vec}
Annamalai Narayanan, Mahinthan Chandramohan, Lihui Chen, Yang Liu, and
  Santhoshkumar Saminathan.
\newblock subgraph2vec: Learning distributed representations of rooted
  sub-graphs from large graphs, 2016.

\bibitem[Gamboa(2017)]{gamboa2017deep}
John Cristian~Borges Gamboa.
\newblock Deep learning for time-series analysis, 2017.

\bibitem[Gehring et~al.(2017)Gehring, Auli, Grangier, Yarats, and
  Dauphin]{gehring2017convolutional}
Jonas Gehring, Michael Auli, David Grangier, Denis Yarats, and Yann~N Dauphin.
\newblock Convolutional sequence to sequence learning.
\newblock In \emph{International conference on machine learning}, pages
  1243--1252. PMLR, 2017.

\bibitem[Khan et~al.(2022)Khan, Naseer, Hayat, Zamir, Khan, and
  Shah]{khan2022transformers}
Salman Khan, Muzammal Naseer, Munawar Hayat, Syed~Waqas Zamir, Fahad~Shahbaz
  Khan, and Mubarak Shah.
\newblock Transformers in vision: A survey.
\newblock \emph{ACM computing surveys (CSUR)}, 54\penalty0 (10s):\penalty0
  1--41, 2022.

\bibitem[Knapp et~al.(2011)Knapp, Ansari, Bain, Bourassa, Dickinson, Funk,
  Helms, Hennon, Holmes, Huffman, et~al.]{knapp2011globally}
Kenneth~R Knapp, Steve Ansari, Caroline~L Bain, Mark~A Bourassa, Michael~J
  Dickinson, Chris Funk, Chip~N Helms, Christopher~C Hennon, Christopher~D
  Holmes, George~J Huffman, et~al.
\newblock Globally gridded satellite observations for climate studies.
\newblock \emph{Bulletin of the American Meteorological Society}, 92\penalty0
  (7):\penalty0 893--907, 2011.

\bibitem[DelSole and Tippett(2014)]{delsole2014comparing}
Timothy DelSole and Michael~K Tippett.
\newblock Comparing forecast skill.
\newblock \emph{Monthly Weather Review}, 142\penalty0 (12):\penalty0
  4658--4678, 2014.

\bibitem[Cash et~al.(2019)Cash, Manganello, and Kinter]{cash2019evaluation}
Benjamin~A Cash, Julia~V Manganello, and James~L Kinter.
\newblock {Evaluation of NMME temperature and precipitation bias and forecast
  skill for South Asia}.
\newblock \emph{Climate dynamics}, 53:\penalty0 7363--7380, 2019.

\bibitem[Wu et~al.(2021)Wu, Peng, Chen, Fu, and Chao]{wu2021rethinking}
Kan Wu, Houwen Peng, Minghao Chen, Jianlong Fu, and Hongyang Chao.
\newblock Rethinking and improving relative position encoding for vision
  transformer.
\newblock In \emph{Proceedings of the IEEE/CVF International Conference on
  Computer Vision}, pages 10033--10041, 2021.

\bibitem[Mamalakis et~al.(2018)Mamalakis, Yu, Randerson, AghaKouchak, and
  Foufoula-Georgiou]{mamalakis2018new}
Antonios Mamalakis, Jin-Yi Yu, James~T Randerson, Amir AghaKouchak, and Efi
  Foufoula-Georgiou.
\newblock A new interhemispheric teleconnection increases predictability of
  winter precipitation in southwestern us.
\newblock \emph{Nature communications}, 9\penalty0 (1):\penalty0 2332, 2018.

\bibitem[Seager et~al.(2007)Seager, Ting, Held, Kushnir, Lu, Vecchi, Huang,
  Harnik, Leetmaa, Lau, et~al.]{seager2007model}
Richard Seager, Mingfang Ting, Isaac Held, Yochanan Kushnir, Jian Lu, Gabriel
  Vecchi, Huei-Ping Huang, Nili Harnik, Ants Leetmaa, Ngar-Cheung Lau, et~al.
\newblock Model projections of an imminent transition to a more arid climate in
  southwestern north america.
\newblock \emph{Science}, 316\penalty0 (5828):\penalty0 1181--1184, 2007.

\bibitem[Dosovitskiy et~al.(2020)Dosovitskiy, Beyer, Kolesnikov, Weissenborn,
  Zhai, Unterthiner, Dehghani, Minderer, Heigold, Gelly,
  et~al.]{dosovitskiy2020image}
Alexey Dosovitskiy, Lucas Beyer, Alexander Kolesnikov, Dirk Weissenborn,
  Xiaohua Zhai, Thomas Unterthiner, Mostafa Dehghani, Matthias Minderer, Georg
  Heigold, Sylvain Gelly, et~al.
\newblock An image is worth 16x16 words: Transformers for image recognition at
  scale, 2020.

\bibitem[Carion et~al.(2020)Carion, Massa, Synnaeve, Usunier, Kirillov, and
  Zagoruyko]{carion2020end}
Nicolas Carion, Francisco Massa, Gabriel Synnaeve, Nicolas Usunier, Alexander
  Kirillov, and Sergey Zagoruyko.
\newblock End-to-end object detection with transformers.
\newblock In \emph{European conference on computer vision}, pages 213--229.
  Springer, 2020.

\bibitem[Chen et~al.(2021)Chen, Wang, Guo, Xu, Deng, Liu, Ma, Xu, Xu, and
  Gao]{chen2021pre}
Hanting Chen, Yunhe Wang, Tianyu Guo, Chang Xu, Yiping Deng, Zhenhua Liu, Siwei
  Ma, Chunjing Xu, Chao Xu, and Wen Gao.
\newblock Pre-trained image processing transformer.
\newblock In \emph{Proceedings of the IEEE/CVF Conference on Computer Vision
  and Pattern Recognition}, pages 12299--12310, 2021.

\bibitem[Bi et~al.(2022)Bi, Xie, Zhang, Chen, Gu, and Tian]{bi2022pangu}
Kaifeng Bi, Lingxi Xie, Hengheng Zhang, Xin Chen, Xiaotao Gu, and Qi~Tian.
\newblock {Pangu-Weather: A 3d high-resolution model for fast and accurate
  global weather forecast}.
\newblock \emph{arXiv preprint arXiv:2211.02556}, 2022.

\bibitem[Pathak et~al.(2022)Pathak, Subramanian, Harrington, Raja,
  Chattopadhyay, Mardani, Kurth, Hall, Li, Azizzadenesheli,
  et~al.]{pathak2022fourcastnet}
Jaideep Pathak, Shashank Subramanian, Peter Harrington, Sanjeev Raja, Ashesh
  Chattopadhyay, Morteza Mardani, Thorsten Kurth, David Hall, Zongyi Li, Kamyar
  Azizzadenesheli, et~al.
\newblock {FourCastNet: A global data-driven high-resolution weather model
  using adaptive Fourier neural operators}.
\newblock \emph{arXiv preprint arXiv:2202.11214}, 2022.

\bibitem[Price et~al.(2023)Price, Sanchez-Gonzalez, Alet, Ewalds, El-Kadi,
  Stott, Mohamed, Battaglia, Lam, and Willson]{price2023gencast}
Ilan Price, Alvaro Sanchez-Gonzalez, Ferran Alet, Timo Ewalds, Andrew El-Kadi,
  Jacklynn Stott, Shakir Mohamed, Peter Battaglia, Remi Lam, and Matthew
  Willson.
\newblock {GenCast: Diffusion-based ensemble forecasting for medium-range
  weather}.
\newblock \emph{arXiv preprint arXiv:2312.15796}, 2023.

\bibitem[Romano et~al.(2019)Romano, Patterson, and
  Cand{\`e}s]{Romano2019ConformalizedQR}
Yaniv Romano, Evan Patterson, and Emmanuel~J. Cand{\`e}s.
\newblock Conformalized quantile regression.
\newblock In \emph{NeurIPS}, 2019.

\bibitem[Wiles et~al.(2021)Wiles, Gowal, Stimberg, Alvise-Rebuffi, Ktena,
  Cemgil, et~al.]{wiles2021fine}
Olivia Wiles, Sven Gowal, Florian Stimberg, Sylvestre Alvise-Rebuffi, Ira
  Ktena, Taylan Cemgil, et~al.
\newblock A fine-grained analysis on distribution shift, 2021.

\bibitem[Subbaswamy et~al.(2021)Subbaswamy, Adams, and
  Saria]{subbaswamy2021evaluating}
Adarsh Subbaswamy, Roy Adams, and Suchi Saria.
\newblock Evaluating model robustness and stability to dataset shift.
\newblock In \emph{International Conference on Artificial Intelligence and
  Statistics}, pages 2611--2619. PMLR, 2021.

\bibitem[Zhu et~al.(2021)Zhu, Ponomareva, Han, and Perozzi]{zhu2021shift}
Qi~Zhu, Natalia Ponomareva, Jiawei Han, and Bryan Perozzi.
\newblock Shift-robust gnns: Overcoming the limitations of localized graph
  training data.
\newblock \emph{Advances in Neural Information Processing Systems},
  34:\penalty0 27965--27977, 2021.

\bibitem[Hao et~al.(2020)Hao, Zhou, and Guo]{hao2020brief}
Shijie Hao, Yuan Zhou, and Yanrong Guo.
\newblock A brief survey on semantic segmentation with deep learning.
\newblock \emph{Neurocomputing}, 406:\penalty0 302--321, 2020.

\bibitem[Gatys et~al.(2016)Gatys, Ecker, and Bethge]{gatys2016image}
Leon~A Gatys, Alexander~S Ecker, and Matthias Bethge.
\newblock Image style transfer using convolutional neural networks.
\newblock In \emph{Proceedings of the IEEE conference on computer vision and
  pattern recognition}, pages 2414--2423, 2016.

\bibitem[Jing et~al.(2019)Jing, Yang, Feng, Ye, Yu, and Song]{jing2019neural}
Yongcheng Jing, Yezhou Yang, Zunlei Feng, Jingwen Ye, Yizhou Yu, and Mingli
  Song.
\newblock Neural style transfer: A review.
\newblock \emph{IEEE transactions on visualization and computer graphics},
  26\penalty0 (11):\penalty0 3365--3385, 2019.

\end{thebibliography}
